\documentclass[letterpaper, 10 pt, journal]{IEEEtran}  
\IEEEoverridecommandlockouts                              

\PassOptionsToPackage{hyphens}{url}
\usepackage[colorlinks=true, linkcolor=red!80!black, citecolor = green!50!black, urlcolor = black,filecolor=black, 
pagebackref=false,hypertexnames=false, plainpages=false, pdfpagelabels ]{hyperref}
\usepackage[square,numbers,sort&compress]{natbib}
\usepackage{graphics}
\usepackage{epsfig}
\usepackage{mathptmx}
\usepackage{times}
\usepackage{amsmath}
\usepackage{amssymb}
\usepackage{tabularx}
\usepackage{multicol}
\usepackage{multirow}
\usepackage{graphicx}
\usepackage{float}
\usepackage{wrapfig}
\usepackage{lipsum}
\usepackage{float}
\restylefloat{figure}
\graphicspath{{./im/}}
\usepackage[font=small]{caption}
\usepackage[font=small]{subcaption}
\usepackage[draft]{pgf}
\usepackage{mathtools}
\usepackage{amsmath}
\usepackage{color}
\usepackage{sidecap}
\usepackage{subcaption}
\usepackage{colortbl}
\usepackage[edges]{forest}
\usepackage{tikz}
\usepackage{makecell}
\definecolor{lightgray}{gray}{0.9}
\usepackage{listings}
\usepackage{enumitem}
\usepackage{comment}
\usepackage{makecell}
\usepackage[]{mdframed}

\definecolor{darkgreen}{rgb}{0.0, 0.4, 0.0}
\definecolor{orange}{rgb}{1.0, 0.49, 0.0}
\definecolor{purple}{rgb}{0.54, 0.17, 0.89}
\definecolor{ForestGreen}{RGB}{34,139,34}
\definecolor{ModernBlue}{RGB}{0,153,153}
\definecolor{LightModernBlue}{RGB}{0,204,204}
\definecolor{DarkPink}{RGB}{204,0,102}

\newcolumntype{L}[1]{>{\raggedright\let\newline\\\arraybackslash\hspace{0pt}}m{#1}}
\newcolumntype{C}[1]{>{\centering\let\newline\\\arraybackslash\hspace{0pt}}m{#1}}
\newcolumntype{R}[1]{>{\raggedleft\let\newline\\\arraybackslash\hspace{0pt}}m{#1}}

\lstdefinestyle{mystyle}{
    frame=single,
    backgroundcolor=\color{backcolour},   
    commentstyle=\bfseries\color{codegreen},
    keywordstyle=\bfseries\color{magenta},
    numberstyle=\tiny\color{codegray},
    stringstyle=\bfseries\color{codepurple},
    basicstyle=\ttfamily\footnotesize,
    breakatwhitespace=false,         
    breaklines=true,                 
    captionpos=b,                    
    keepspaces=true,                 
    numbersep=5pt,                  
    showspaces=false,                
    showstringspaces=false,
    showtabs=false,                  
    tabsize=2
}

\title{\LARGE \bf
Principles and Guidelines\\ for Evaluating Social Robot Navigation Algorithms
}

\author{
{
Anthony~Francis\textsuperscript{1,\ddag,*},
Claudia~P\'{e}rez-D’Arpino{\textsuperscript{2,*}},
Chengshu~Li{\textsuperscript{3}},
Fei~Xia{\textsuperscript{4}},
\\
Alexandre~Alahi{\textsuperscript{5}},
Rachid Alami{\textsuperscript{15}},
Aniket~Bera{\textsuperscript{6}},
Abhijat~Biswas{\textsuperscript{7}},
Joydeep~Biswas{\textsuperscript{8}},
Rohan~Chandra{\textsuperscript{8}},
Hao-Tien~Lewis~Chiang{\textsuperscript{4}},
Michael~Everett{\textsuperscript{10}},
Sehoon~Ha{\textsuperscript{11}},
Justin~Hart{\textsuperscript{8}},
Jonathan P. How{\textsuperscript{9}},
Haresh~Karnan{\textsuperscript{8}},
Tsang-Wei~Edward~Lee{\textsuperscript{4}},
Luis~J.~Manso{\textsuperscript{12}},
Reuth~Mirksy{\textsuperscript{13}},
S\"oren~Pirk{\textsuperscript{14}},
Phani~Teja~Singamaneni{\textsuperscript{15}},
Peter~Stone{\textsuperscript{8,16}},
Ada~V.~Taylor{\textsuperscript{7}},
Peter~Trautman{\textsuperscript{17}},
Nathan~Tsoi{\textsuperscript{18}},
Marynel~V\'azquez{\textsuperscript{18}},\\
Xuesu~Xiao{\textsuperscript{19}},
Peng~Xu{\textsuperscript{4}},
Naoki~Yokoyama{\textsuperscript{11}},\\
Alexander~Toshev{\textsuperscript{20,\dag}},
Roberto~Mart\'{i}n-Mart\'{i}n{\textsuperscript{8,\dag}}
}
\\~\\
{{\textsuperscript{1}}Logical~Robotics}
{{\textsuperscript{2}}NVIDIA}
{{\textsuperscript{3}}Stanford}
{{\textsuperscript{4}}Google}
{{\textsuperscript{5}}EPFL}
{{\textsuperscript{6}}Purdue}
{{\textsuperscript{7}}CMU}
{{\textsuperscript{8}}UT Austin}
{{\textsuperscript{9}}MIT}
\\
{{\textsuperscript{10}}Northeastern}
{{\textsuperscript{11}}Georgia~Tech}
{{\textsuperscript{12}}Aston}
{{\textsuperscript{13}}Bar~Ilan}
{{\textsuperscript{14}}Adobe}
{{\textsuperscript{15}}LAAS-CNRS,~Université~de~Toulouse}
\\
{{\textsuperscript{16}}Sony AI}
{{\textsuperscript{17}}Honda}
{{\textsuperscript{18}}Yale}
{{\textsuperscript{19}}GMU}
{{\textsuperscript{20}}Apple}
\thanks{\textsuperscript{*} These authors contributed equally.}
\thanks{\textsuperscript{\dag} Equal advising contribution.}
\thanks{\textsuperscript{\ddag} Anthony Francis's contributions began while at Google.}
}

\begin{document}

\maketitle


\begin{abstract}

A major challenge to deploying robots widely is navigation in human-populated environments, commonly referred to as \textit{social robot navigation}. While the field of social navigation has advanced tremendously in recent years, the fair evaluation of algorithms that tackle social navigation remains hard because it involves not just robotic agents moving in static environments but also dynamic human agents and their perceptions of the appropriateness of robot behavior. 
In contrast, clear, repeatable, and accessible benchmarks have accelerated progress in fields like computer vision, natural language processing 
and traditional robot navigation
by enabling researchers to fairly compare algorithms, revealing limitations of existing solutions and illuminating promising new directions.  We believe the same approach can benefit social navigation. 
In this paper, we pave the road towards common, widely accessible, and repeatable benchmarking criteria to evaluate social robot navigation.
Our contributions include (a) a definition of a socially navigating robot as one that respects the principles of safety, comfort, legibility, politeness, social competency, agent understanding, proactivity, and responsiveness to context, (b) guidelines for the use of metrics, development of scenarios, benchmarks, datasets, and simulators to evaluate social navigation, and (c) a design of a social navigation metrics framework to make it easier to compare results from different simulators, robots and datasets.

\end{abstract}


\begin{figure*}[t]
    \includegraphics[width=\textwidth]{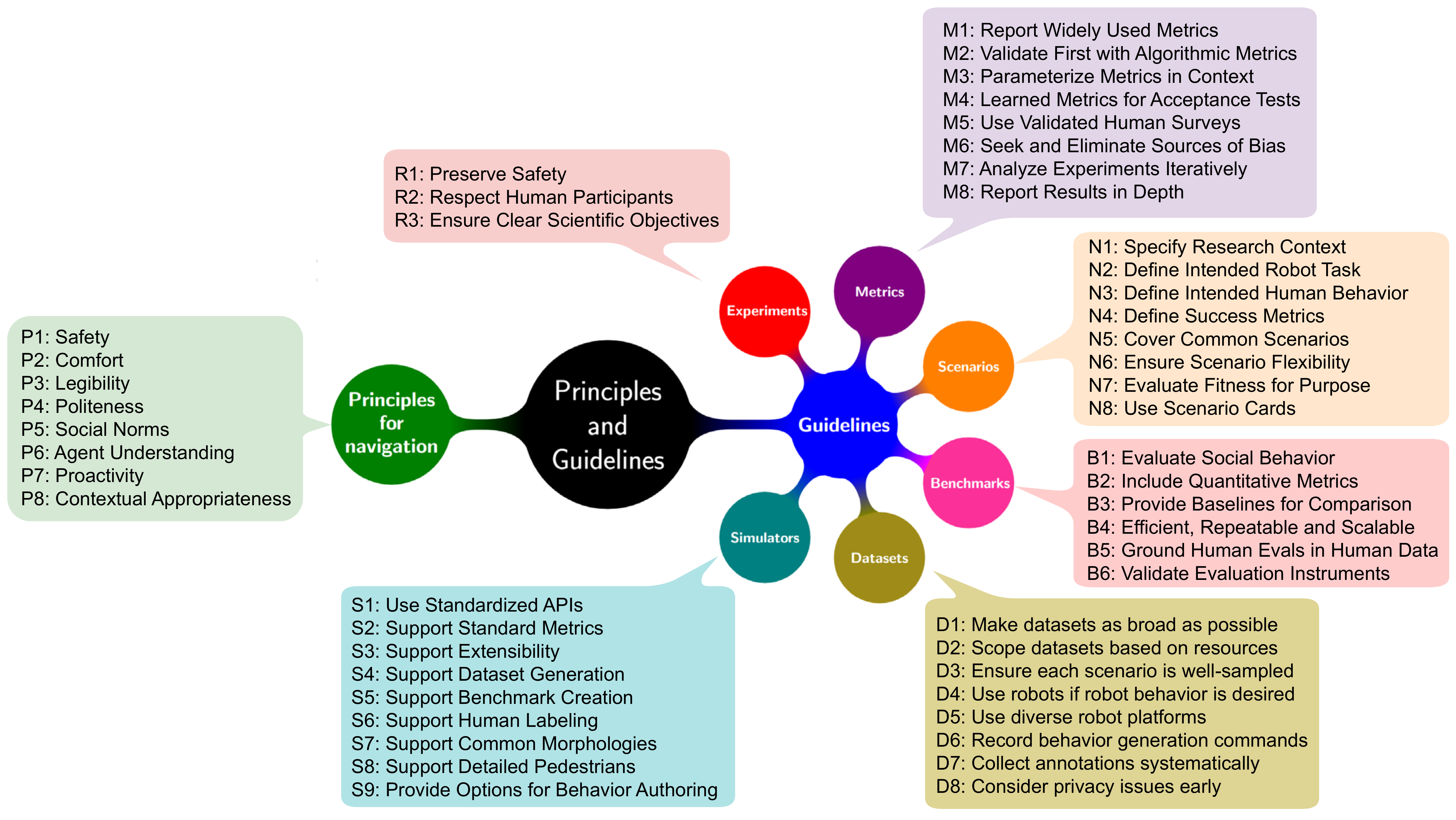}
    \caption{We identify eight broad principles of social robot navigation - including safety, comfort, legibility, politeness, social competency, agent understanding, proactivity, and contextual appropriateness - which motivate specific guidelines for experiments, metrics, scenarios, benchmarks, datasets, and simulators. Principles and guidelines are labeled with two-letter codes, with P for principles, R for real-world issues, M for metrics, N for scenarios, B for benchmarks, D for datasets, and S for simulators.}
    \label{fig:guidelines}
\end{figure*}

\section{Introduction}
\label{s:intro}

The study of social robot navigation has a long history, but a crisp definition of what makes navigation ``social''
remains elusive.
Researchers on social robot navigation often have a personal sense of what it is and use that intuition to guide their research into how to make robots move better around people, but the field does not yet have a consensus on a definition of social navigation or how to achieve it. 
Indeed, at the Social Navigation Symposium\footnote{\url{https://sites.google.com/view/socialnavigationsymposium/home}}, a diverse spectrum of researchers presented a variety of views on what robotic social navigation is and their approaches to solving it, including a range of definitions, variants, problems and subproblems.

Ideas presented at the the Symposium included a variety of methods to evaluate social navigation performance, involving different experimental setups, evaluation metrics, robot simulators, social datasets, and deployment environments.
As the researchers continued their discussions following the symposium, a taxonomy of aspects of social navigation began to emerge, which helped clarify the social robot navigation problem and converged to a set of general recommendations on how to evaluate solutions in ways that were more comparable.

This paper summarizes our work to define the social robot navigation problem, identify a taxonomy of its important aspects, create guidelines for its evaluation, and define a common API to make evaluations more comparable. After a review of related work in Section \ref{s:relatedwork}, 
Section \ref{s:definition} proposes a definition of social navigation and a strategy for achieving it by respecting other agents' goals through principles of safety, comfort, legibility, politeness, social competency, agent understanding, proactivity, and responsiveness to context. Section \ref{s:questions} reviews the different scientific questions asked by social navigation researchers, and Section \ref{s:taxonomy} outlines our taxonomy for analyzing social navigation benchmarks, datasets and simulators. Section \ref{s:metrics} then discusses the metrics that have been developed for measuring social navigation, including subjective human evaluation metrics, analytic metrics that can be directly computed, and research efforts designed to create learned metrics. Section \ref{s:scenarios} discusses the typical scenarios used in social navigation, and Section \ref{s:benchmarks} describes benchmarks built on these scenarios, while Section \ref{s:datasets} reviews datasets collected on social navigation. Section \ref{s:simulation} reviews simulators and presents our work to create a unified interface across simulators. 

Figures~\ref{fig:guidelines}~and~\ref{fig:principles} illustrate the principles and guidelines we present for the development and evaluation of social navigation. Principles are high-level goals that social navigation methods should try to achieve, as illustrated in Figure~\ref{fig:principles}. Guidelines are concrete, actionable recommendations that practitioners of social navigation research may consider when creating and testing their solutions, as summarized in Figure~\ref{fig:guidelines} and unpacked in the rest of the document.
We hope our work helps researchers create fair and comparable evaluations of their social navigation solutions, which will shed light on the field's current state and point the way to the challenges ahead.

\section{Related Work}
\label{s:relatedwork}

The field of social robot navigation is vast and we will not attempt to summarize it; instead, we refer to many recent surveys on social navigation  \cite{kruse2013human,rios2015proxemics,chik2016review,charalampous2017recent,cheng2018autonomous,gao2021evaluation,mavrogiannis2021core,mirsky2021prevention,moller2021survey,wang2022metrics}. 
Among these, \cite{gao2021evaluation} focuses on evaluating social robot navigation algorithms, reviewing 177 recent papers to gather evaluation methods, scenarios, datasets, and metrics, using their findings to discuss shortcomings of existing research and to make recommendations for future research directions.  Another recent survey by Mavrogiannis et al.~\cite{mavrogiannis2021core} focuses on the core challenges of social navigation with respect to navigation algorithms, human behavior models, and evaluation. Our work builds on the works of \cite{gao2021evaluation} and \cite{mavrogiannis2021core} and similar surveys to map the field. We contribute a crisp definition of social robot navigation based on discussions held at the 2022 Social Navigation Symposium, an overview of methodologies for research, and a taxonomy of the field which we use to examine existing metrics, scenarios, benchmarks, datasets, and simulators, and shared principles to make social navigation evaluations comparable across the community.

\cite{wang2022metrics} proposes new metrics evaluating the principles defined in \cite{kruse2013human}, comfort, naturalness, and sociability. We expand the principles in \cite{kruse2013human} to a broader set including safety, comfort, legibility, politeness, social norms, agent understanding, proactivity, and contextual appropriateness, and proposes a lifecycle of social navigation with recommendations for metrics, scenarios, benchmarks, datasets and simulators, along with guidelines for metric usage and recommended metrics.

Beyond social navigation, clear, repeatable, and accessible benchmarks have accelerated progress in fields like computer vision~\cite{russakovsky2015imagenet} and natural language processing~\cite{bowman-etal-2015-large,rajpurkar2016squad,wang2019superglue} 
enabling researchers to compare algorithms, revealing limitations of existing solutions, and illuminating promising new directions. Our effort builds on benchmark challenges in traditional robot navigation~\cite{anderson2018evaluation,savva2019habitat,48697interactivegibsonRAL,deitke2020robothor, xiao2022motion, deitke2022retrospectives}, social navigation benchmarks and challenges~\cite{kastner2022arena,inria2020safe,nair7dynabarn,Everett18_IROS,perez2023hunavsim,li2021igibson, shen2021igibson,biswas2022socnavbench,seanavbench} and social navigation scenario development~\cite{pirk2022protocol,xiao2022learning,cuan2022gesture2path}. We review social navigation scenarios and benchmarks in Sections \ref{s:scenarios} and \ref{s:benchmarks}. We contribute guidelines for scenario development, a review of scenarios in the literature, a social navigation scenario card, as well as guidelines for social navigation benchmarking and dataset development.

Simulators are a key component in social navigation, though many simulators exist with diverse APIs which are largely not compatible.  \cite{kastner2022arena}, discussed in more detail in Section \ref{s:benchmarks} is a benchmark which provides an API for easily generating new worlds and tasks for two different simulators. This paper proposes guidelines for simulator development and usage, as well as a common API design to unify simulator outputs to facilitate common evaluations using shared metrics.

\section{Towards a Definition of Social Navigation}
\label{s:definition}

Social navigation refers to a range of behaviors from simple navigation around dynamic obstacles, to complying with complex social norms, up to navigating with communicative intent. 
As such, it risks becoming a “suitcase word”, defined by Marvin Minsky~\cite{minsky2007emotion} as words that carry other concepts inside them, like memory, emotions, or consciousness; these terms must be unpacked to fully understand their meanings.

In this section, we unpack the term “social robot navigation”. First, we examine social robotics and what it means; then we focus on the social navigation problem itself. Then we examine the subproblems of social navigation and how context can affect what behaviors are considered social, formulating these subproblems as principles to guide research.

\subsection{What is a Social Robot?}\label{sec:social-robot}

Intuitively, we expect social robots to be able to recognize social cues, norms, and expectations, to have the understanding to interpret them correctly, and to have the capabilities to respond appropriately. This raises the question of what ``social'' is, and what kinds of sensing, interpretation and capabilities social robots need to effectively navigate social interactions.

In their review of Human-Robot Interaction (HRI) for social robotics, Kanda and Ishiguro~\cite{kanda2017human} argue that in addition to navigation (moving robots from place to place) and manipulation (changing objects in the environment) capable robots must also leverage social interactions, i.e., be able to interplay with humans or other robots to perform tasks. Further, they distinguish robots that simply encounter humans from those which have specifically designed socially interactive features, such as voices, expressive faces or the ability to gesture.

But simply having socially interactive features in a robot does not mean that the quality of its interactions would be acceptable to humans or efficient for other robots; additional principles are needed to apply these features in a positive way. Developing solutions that create high-quality social interactions autonomously is difficult; many social interactions that Kanda and Ishiguro studied were beyond the technology of the time and required a human to teleoperate the robot. What distinguishes ``social" robotics from pure interactivity?

To define ``social'' more precisely, we examined the terms social and antisocial for humans. Social sometimes means participating in society, i.e., participating in an interacting group whose individuals modify  their behavior to accommodate the needs of others while achieving their own. But social has a second meaning: a social individual has outstanding skills to work with others, based on an understanding of their feelings and needs and adapting to them. Antisocial individuals, in contrast, fail to follow the customs of society or live without consideration for others.
Inspired by these terms when applied to humans, we generalize this notion to other agents, and offer this definition of social robot navigation:

\begin{quote}
\begin{mdframed}
A socially navigating robot acts and interacts with humans or other robots, achieving its navigation goals while modifying its behavior so the experience of agents around the robot is not degraded or is even enhanced.
\end{mdframed}
\end{quote}
This social quality may be reflected through overt behavior changes, such as respecting social norms, or through understanding other agents’ needs, feelings, and capabilities.

\begin{figure*}[t]
    \includegraphics[width=\textwidth]{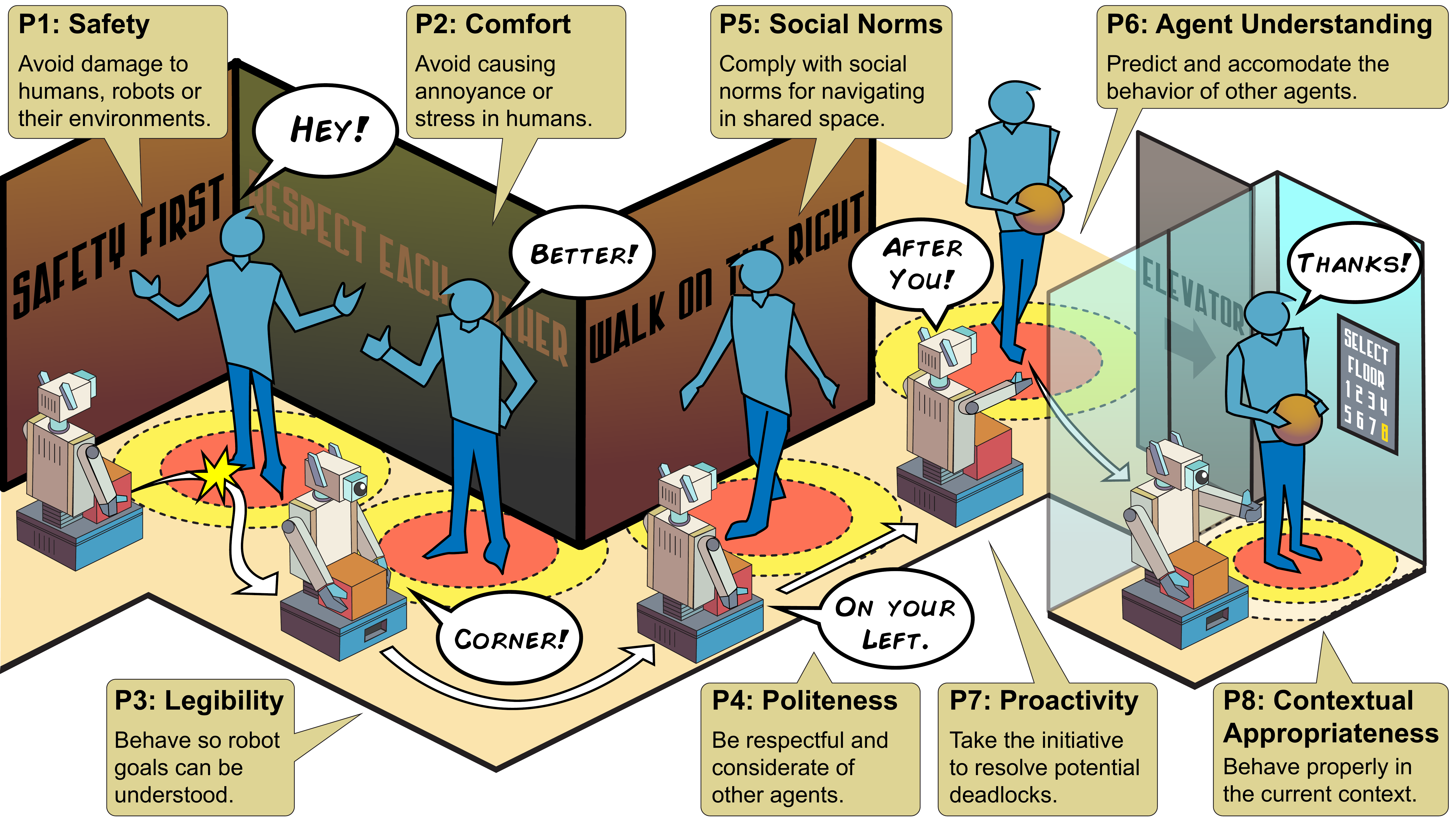}
    \caption{We define a socially navigating robot as one that interacts with humans and other robots in a way that achieves its navigation goals while enabling other agents to achieve theirs. To make this objective achievable, we propose eight principles for social robot navigation: safety, comfort, legibility, politeness, social competency, agent understanding, proactivity, and contextual appropriateness.}
    \label{fig:principles}
\end{figure*}

\subsection{Principles of Social Navigation }
\label{s:principles}

It is often difficult for an agent to know exactly what other agents, especially humans, need to achieve, or what they feel and like, and social norms that could guide us are often not verbalized. To operationalize these concerns, we identified \textit{principles of social navigation} that can be used to evaluate the quality of social behavior, including \textit{(1) safety, (2) comfort, (3) legibility, (4) politeness, (5) social competency, (6) understanding other agents, (7) proactivity, and (8) responding appropriately to context}, as illustrated in Figure~\ref{fig:principles}.

\begin{figure*}
\centering
\includegraphics[width=0.95\textwidth]{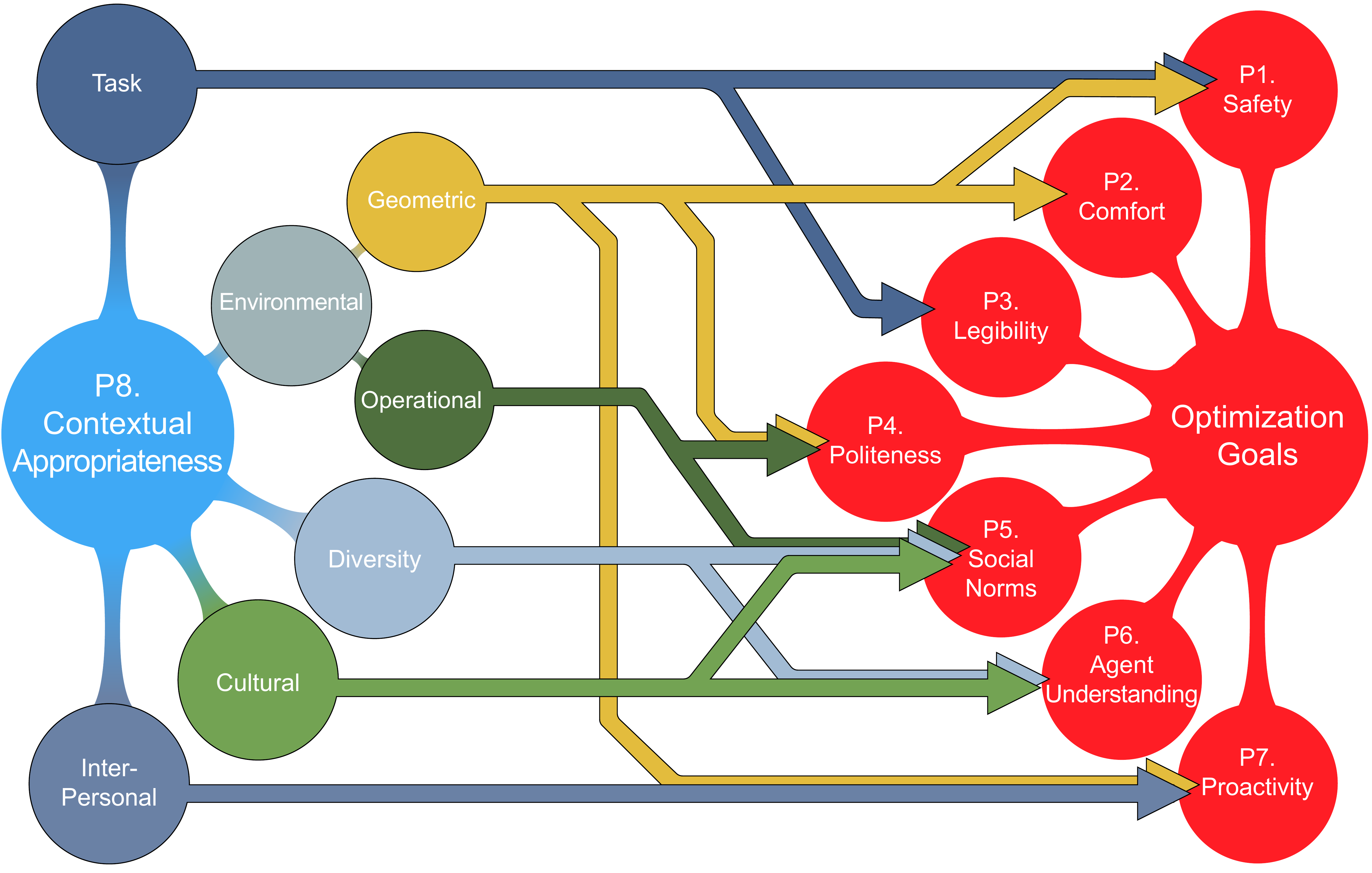}
\caption{Contextual factors of social navigation. While the first seven principles represent factors to optimize, the eighth principle, contextual appropriateness, calls out that the weighting of these factors can be affected by many features, including cultural, diversity, environmental, task and interpersonal context. Lines in the diagram are representative of common interactions but are not exclusive.}
\label{fig:context}
\end{figure*}

 Seen from the lens of optimization, the first seven principles of social navigation can be formalized as additional objectives that the robot needs to optimize for while still achieving its main objective, and the eighth principle, context, can be seen as weighting  which principles are most important at any given time, as shown in Figure~\ref{fig:context}. These principles are not completely orthogonal: improving legibility might improve safety and even comfort, whereas nonverbal politeness depends on understanding other agents' trajectories. In addition, what is considered appropriate or polite behavior depends on both the cultural context \cite{recchiuto2022diversity} and the robot's main objective: for example, a delivery robot arguably should maintain a greater distance from humans than one functioning as a guide.
 
 The principles mentioned above guide the development of metrics to evaluate social robots, discussed in more depth in Section \ref{s:metrics}. Properly studying these principles of social navigation directly impacts which metrics to measure \cite{gao2021evaluation}, what datasets to collect, how to build simulators, and how to structure benchmarks. 
In the following sections, we unpack these principles as often used in social robotics research:

\vspace{0.5em}
\textbf{Principle P1: Safety - Protect humans, other robots and their environments.}
A minimal requirement for robots and human sociality is not harming others in the course of business \cite{lasota2017survey,boddington2017epsrc}, as the robot fails to do in the first scenario in Figure~\ref{fig:principles} when it collides with a human's toe. Avoiding collisions with humans is important but is not the only safety concern \cite{mavrogiannis2021core, bera2017sociosense, narayanan2020proxemo}; robots can damage each other or their environments. While it might be acceptable for a factory robot to bump a guardrail defining the edge of its workspace, social robots should generally avoid damaging human environments, which often contain important objects that can be damaged or wall coverings whose visual appearance is important. Robots should also avoid damaging each other, or behaving in a way that induces humans or other robots to injure themselves.

\vspace{0.5em}
\textbf{Principle P2: Comfort - Do not create annoyance or stress.}
Humans should also feel comfortable around robots, defined in \cite{kruse2013human} as \textit{the absence of annoyance and stress for humans in interaction with robots}. Many features contribute to comfort, including maintaining human-robot distance, not cutting humans off, and naturalness of motion. Unacceptable robot speed, navigation jitter and unexpected head movements are factors that degrade humans' perception of comfort. Additionally, social robots should arguably not behave in a way that triggers the safety layers of other robots. \cite{kruse2013human} further argue that annoyance can be triggered by a failure to respect proxemics, the virtual personal space around a human that other humans instinctively respect~\cite{hall1966hidden}. Figure~\ref{fig:principles} illustrates proxemics with the ``intimate'' distance of $0.45m$ shown in red and the ``personal'' distance of $1.2m$; after initially violating a human's personal distance, the robot is shown attempting to stay in ``personal'' or more distant ``social'' spaces, except as required by the geometric context. Proxemics is a rich and controversial field; for an in-depth survey see~\cite{rios2015proxemics}.

\vspace{0.5em}
\textbf{Principle P3: Legibility - Behave so goals can be understood.}
Legibility refers to the property of an agent’s behavior that makes it possible for other agents to infer their goals \cite{dragan2013legibility}. This includes not only the robot’s goal but also incidental interactions when performing other tasks, e.g., moving to the right or left when passing in a hallway. Dragan and colleagues \cite{dragan2013legibility} suggests that legibility involves relaxing constraints such as predictability of trajectories (in the sense of an agent's own predictable style) in favor of more clearly understood behaviors (in the sense of moving to make goals explicit). Legibility arguably can help other robots understand another social robot's goals. While \cite{dragan2013legibility} focused on changes to robot paths to make them legible, a robot capable of communicating could explicitly announce its intentions, the way restaurant staff are trained to call ``corner'' when entering a blind corner, as the robot does in the middle of Figure~\ref{fig:principles}.

\vspace{0.5em}
\textbf{Principle P4: Politeness - Be respectful and considerate.}
Politeness refers to behavior that is respectful and considerate of people. There are at least two dimensions: physical politeness (how robots navigate around people, such as not cutting people off) and communicative politeness (gestures or verbal signals, such as saying ``excuse me'', or ``on your left'', as the robot does in Figure~\ref{fig:principles} when a narrow hallway forces it to transgress on a human's personal space). 
Politeness can have a strong effect on people’s perception of robots \cite{inbar2019politeness, randhavane2019pedestrian}. Social robots should also be considerate of each other, so they do not prevent other robots from accomplishing their tasks.

\vspace{0.5em}
\textbf{Principle P5: Social Competency - Comply with social norms.} 
Robots should comply with social, political, and legal norms for sharing space. Many social competencies are matters of following conventions rather than optimizing performance~\cite{cheung2018identifying,randhavane2019pedestrian,chen2017sociallyawareDRLmit_iros17}. For example, in the absence of norms there is no optimization preference for driving on the left or right, but identifying and following the local norms helps prevent conflicts~\cite{mirsky2021prevention} in the the third hallway interaction in Figure~\ref{fig:principles}. Some social competencies, like turn-taking, can emerge naturally \cite{kose2008emergent}, whereas others must be specifically engineered (or simulated with wizard-of-oz studies \cite{kanda2017human} if existing policies are not up to the task). Social norms may apply to more than just humans; conventions of behaviors may make it easier for robots to interact.

\vspace{0.5em}
\textbf{Principle P6: Understanding Other Agents - Predict and accommodate the behavior of other agents.} A significant portion of the discussion during the symposium revolved around the need to understand other agents in order to be able to correctly fulfill the principles listed above and be considered a capable social agent. Understanding, accommodating, and even facilitating other agents’ goals, activities, and motions was agreed to be a key element of social behavior. Accommodating the goals and comfort of other agents requires understanding of what they are perceiving, doing, and trying to accomplish. For example, to pass between two agents politely, it is important to understand whether they are in conversation \cite{petrak2021you}. 
Understanding when the \textit{interaction potential} - the likelihood of robots entering human personal space - should be minimized \cite{trautman2010unfreezing, bera2018socially} or maximized \cite{mead2017autonomous} depends on the task. Further, understanding how agents move can reduce the potential for \textit{conflicts} (short-term encounters in which humans and robots would collide without intervention \cite{mirsky2021prevention, chandra2023socialmapf}), as in the right side of Figure~\ref{fig:principles}, where a robot recognizes the human's path will cross theirs and stops to prevent a conflict.

\vspace{0.5em}
\textbf{Principle P7: Proactivity -  Taking the Initiative to Prevent and Resolve Issues} Simply understanding other agents is not enough, however: in some circumstances, being social involves taking the initiative. Perhaps the canonical example is several cars meeting at a four-way intersection: if all drivers act conservatively, there can be considerable delay~\cite{suriyarachchi2022gameopt}. To resolve the conflict in a timely fashion, one driver must either take the initiative and go first, or proactively propose a solution by waving other drivers through the intersection \cite{chandra2022gameplan}. Studies of self-driving vehicles suggest that non-conservative (or ``aggressive'') behavior may even be desirable if that is expected by a population of drivers \cite{chandra2021using, chandra2022game}. While it is less likely that socially navigating robots will be mistaken for humans as frequently as autonomous vehicles are mistaken for human-driven ones, appropriately resolving potential issues, similar situations to the four-way intersection deadlock can arise, such as when two pedestrians dodge to the same side or contend for the same doorway. In these cases, a socially navigating robot that can take the initiative to avoid a human or proactively suggest a solution will arguably be more social than one that is not, as in the right side of Figure~\ref{fig:principles}, where the robot proactively proposes that the human enter the elevator first to prevent this kind of deadlock. Section~\ref{s:scenarios} discusses measuring proactivity with scenarios designed to explicitly elicit this phenomenon.

\vspace{0.5em}
\textbf{Principle P8: Contextual Appropriateness - Behave properly in the current context.}
Social navigation should be evaluated within the context that it is to be deployed. Context helps us understand the relative importance of the previous objectives, and is a complex construct in its own right, as shown in Figure~\ref{fig:context}. An example shared in the symposium was a \textsc{Crash Cart} robot in a hospital bringing an emergency drug to a doctor: politeness is less important than task success. Also, when navigating a narrow corridor, we may be “less polite” and get closer to other agents. We identified several forms of context, including cultural context, environmental context, diversity, task context, and interpersonal context, all of which can change which response is right in a given situation:  
\begin{itemize}
    \item \textbf{Cultural Context:} Different cultures have different social norms, as notably documented in \cite{hall1966hidden}; more recently \cite{bera2017sociosense, bera2016glmp} and \cite{recchiuto2022diversity} examined cultural norms in social robotics but concluded more work needs to be done.
    \item \textbf{Diversity Context:} Different individuals with different abilities or different background histories may need different accommodations \cite{recchiuto2022diversity}.
    \item \textbf{Environmental Context:} The environment may affect the social navigation problem \cite{nigam2015social}, and includes both geometric factors - the shape of the space - and operational factors - how that shape is to be used.
    \begin{itemize}
        \item \textbf{Geometric Context:} The geometry of the environment may affect the social navigation problem . For example, the more crowded the space is, the smaller the acceptable distance is between the robot and other people.
        \item \textbf{Operational Context:} The operational domain the robot is intended to work in in affects what behaviors are considered good: for example, a robot may drive slower in a daycare than an office, even if the two settings had identical geometric layouts.
    \end{itemize}
    \item \textbf{Task Context:} In turn, the task the robot operates in affects what behaviors are appropriate: for example, even in a single environment like a hospital, whether a robot is performing mail delivery or is a crash cart changes its weighting of politeness against speed.
    \item \textbf{Interpersonal Context:} While there are many different areas of context that are appropriate, interpersonal context (e.g. whether humans are independent pedestrians, are traveling in a group, or stopped and conversing is critical to know how to navigate among them)
\end{itemize}

As an illustration of context, the robot in Figure~\ref{fig:principles} is first shown violating a person's intimate space distance in red, then attempting to avoid proxemics violations going forward. However, the corridor around the bend is too narrow to prevent the robot from passing though a person's personal space distance in yellow, prompting the robot to politely call out its presence. Then, in the relatively small elevator, the standard interpersonal distances are no longer easy to achieve, and both the robot and human adjust their perceived proxemics radii based on the current context, shown as a contraction of the proxemics circles from their original size.

Social navigation should be evaluated within the context that it is intended to be deployed. While defining context in a sufficiently precise way for a robot to identify or respond to it is a challenging problem, at the least the intended context should be defined well enough in terms of cultural, diversity, environmental, operational, task and interpersonal context for other researchers to gauge the applicability of the ideas and findings conveyed by research.

\section{Research Methodologies of Social Navigation}
\label{s:questions}

Benchmarks require measures and an evaluation methodology for comparing social robot navigation systems. Because social navigation is concerned with methods for controlling mobile robots to operate effectively around people, many at the symposium argued social robot evaluation methodologies should involve the collection of human perceptions of robot behavior; however, others argued there are important scientific questions, such as an algorithm's response to dynamic obstacles, which can affect social principles like safety and comfort. Ultimately, the scientific questions a benchmark asks are important in deciding its scope and content. These scientific questions lead to varied types of studies which gather different types of data, which in turn guide the development of methods, creating a lifecycle of social navigation research.

\subsection{Research Questions of Social Navigation}
The overarching research question of social robot navigation is developing a scientific understanding of the problem sufficient to build computational models that enable robots to perform acceptably in human environments.
This involves understanding the factors that influence social navigation, developing models of those factors, and implementing algorithms that take those factors into account. To fairly evaluate how algorithms compare to each other, we need benchmarks that help us understand their differences and identify which ones are better.
Given the complexity of social navigation, different benchmarks often focus on different aspects of the problem and thus different, more specific scientific questions. Some of these questions arise from traditional robot navigation research and can arguably be evaluated using traditional methods, with adjustments for human participants:

\begin{itemize}
\item \textbf{How do methods compare with each other against baselines?} Some aspects of method evaluation involve quantitative metrics measurable in simulation, such as revealing problems in a robot's safety layer as it faces increasing obstacle densities. However, when human evaluations are required, these are typically conducted in the real world, though toolkits are now coming into use that enable labeling simulated trajectories as well~\cite{baghel2021toolkit}. 

\item \textbf{How do components of a method affect its overall performance?} These are generally conducted by turning method components off, often called ``ablation studies'' in analogy to ablation studies in neuroscience \cite{meyes2019ablation}. While in theory ablation studies could be conducted on-robot, in practice these studies are often only conducted in simulation, as real human participant time can be wasted on variants of the algorithm expected to perform poorly (or known to perform poorly in simulation).

\item \textbf{How do behaviors generalize to different environments?} Benchmarks can test methods under different conditions to evaluate this, a task that is easier (though less realistic) in simulation.
\end{itemize}

Success at task performance is often measurable quantitative, but determining whether a robot is satisfying the principles of social navigation is trickier. While the physical aspects of Principle P1, Safety, may arguably be measureable quantitiatively (at least in the sense that the lack of safety can be measured through damage and collisions), others like Principle P4, Politeness, often require human evaluation, and still others like Principle P2, Comfort, are often explicitly defined in terms of human reactions. 
Scientific questions involving these subjective aspects therefore generally require measuring human perceptions and reactions to robot behaviors, and are best investigated through HRI studies:

\begin{itemize}
\item \textbf{Human Ratings:} How do humans rate the socialness of social navigation methods, either intrinsically or in comparison to a baseline? For some researchers, human ratings of policy behavior in real contexts are the gold standard for policy performance, but for these ratings to be effective, studies must follow proper HRI protocols and use validated survey instruments \cite{hoffman2020primer}.

\item \textbf{Behavior Analysis:} How does human behavior change when exposed to different robot navigation policies? While ratings are explicit, behavior change is implicit or even unconscious. Studies should be conducted according to HRI guidelines to ensure conditions are appropriately blinded so participant and rater reactions are valid. 

\item \textbf{Issue Discovery:} Benchmarks can also be used to conduct exploratory analyses. For example, these analyses could find out the frequency of encounter types between humans and robots as well as the frequency of problems that affect a given policy. This can guide research in the direction of problems that occur in the wild. These studies must be conducted with a robot in a live deployment.
\end{itemize}

Many benchmarks focus on a subset of these questions because different researchers have different aims and different groups have different needs. As a result, social navigation evaluation methodologies have become fragmented and a comprehensive evaluation methodology does not yet exist. 

Because different lines of research have different needs, we do not aim to provide one evaluation protocol for all social navigation methods, but a methodology by which researchers can make principled decisions to guide their own evaluations. Such a protocol will allow researchers to compare social navigation methods along the dimension relevant to a specific subdomain. The direct comparison of methods, via principled evaluation protocols, will allow us to interrogate the strengths and weaknesses of methods, push knowledge discovery in the field, and increase method performance.

In Section \ref{s:benchmarks} we argue that because social navigation involves understanding of both how robots affect other agents and which methods are effective, most benchmarks will benefit from incorporating both HRI components that evaluate human reactions in the real world as well as ablation studies, even if those are constrained to simulation.

\subsection{Types of Social Navigation Studies}
\label{s:study-types}

To enable progress on both social navigation policy development and the community’s scientific understanding of social navigation, we advocate viewing in-the-wild studies and controlled scenarios as part of a lifecycle of study of social navigation phenomena. To define terms, we can distinguish several different major classes of social navigation studies:

\begin{enumerate}

\item \textbf{Field Studies:} Field studies involve pedestrians who are not confederates of the experimenters, such as a mall, campus or boardwalk. Such studies are often called ``in the wild" as they are conducted in uncontrolled environments. Field studies provide an opportunity to collect natural data about robot-human interactions outside the influence of experiments or instructions, but individual encounters are not directly reproducible. Very large-scale studies offer a proxy of reproducibility when rare events re-occur with enough statistical frequency to be analyzed; however, large-scale field studies are the most resource-intensive, complex and potentially dangerous to conduct.

\item \textbf{Robot Deployments:} Robot deployments are conducted in environments partially under the control of the experimenters, such as an office, a classroom building, or factory. In this case, robot deployments necessarily involve experimenters informing participants about the robots, which may change their responses compared to someone encountering a robot in the wild; furthermore, participants necessarily develop experience about the robots that can distort human-robot interactions. Symposium participants reported that users unfamiliar to robots were less accepting of errors than robot researchers, who in turn were less accepting than experienced ``robot wranglers" responsible for managing the deployment; these anecdotal reports mirror studies which found evidence that both general computer user skill level \cite{kanda2001psychological} and familiarity with particular robots \cite{kim2013effects,paetzel2020persistence,saunderson2021robots} could improve assessments of robot capabilities and behavior. Semi-controlled robot deployments are similar to, but less naturalistic than field studies, but because robot deployment environments are more controlled than true in-the-wild studies, larger scale is often more practicable by conducting experiments over a longer period of time. For example, \cite{biswas20161} collected 1,000 kilometers of indoor navigation on a college campus, and the system described in \cite{xiao2022learning} was part of a deployment at Google that collected over 3,000 kilometers of data.

\item \textbf{Laboratory Experiments:} Laboratory experiments are sometimes considered the gold standard in science, but have distorting effects on human behavior due to the controlled environment and experiment instructions. While A/B testing in field studies or robot deployments can compare some algorithms, laboratory experiments are often necessary to answer scientific questions about human reactions to changes in robot behavior or to evaluate algorithmic changes prior to larger-scale deployments. However, we also need to ensure that laboratory experiments have good \textit{ecological validity}, defined as the degree to which laboratory results generalize to the real world~\cite{orne1962social,schmuckler2001ecological,kihlstrom2021ecological}. For social navigation experiments, the ecological validity of an experiment in turn depends on whether the scenario it tests has been properly validated. We discuss a methodology for scenario design in Section~\ref{s:scenario-methodology}, but validating scientific instruments to determine whether they correctly evaluate the variables they are designed to test, often called \textit{construct validity}, can take several iterations of experiment and analysis \cite{morling2014research, litwin1995measure}.

\item \textbf{Social Navigation Scenarios:} Social navigation scenarios, such as \textsc{Frontal Approach}, \textsc{Pedestrian Overtaking}, and \textsc{Intersection}, can be viewed as a subset of laboratory experiments which test specific scenarios discovered through field studies or robot deployments, with well-defined configurations validated through theoretical analysis, pilot studies, or social navigation issue discovery in existing datasets. The social navigation community is collecting a growing set of scenarios to guide experiments, enable data collection for imitation learning, and serve as regression tests for behavior.

\item \textbf{Staged Social Interactions:}  Due to the excessive costs of field studies and the lack of rare, naturally-occuring human-robot encounters in robot deployments and laboratory experiments, researchers developed Staged Social Interactions to evaluate robot social navigation.
In staged social interactions, participants are recruited to act in a structured but free-form fashion; this can be an explicit set of
scripts (so-called ``Guided Crowd Scenarios'') or a less
structured activity such as a ``Robot Happy Hour'' where
participants are recruited to perform a social activity
around where robots are operating. These studies are less
controllable than social navigation scenarios, and their
``staged'' nature makes them closer to robot deployments
or laboratory experiments than true field studies. However, they can create higher-density free-form interactions than may otherwise be available.

\end{enumerate}

\begin{figure}
\centering
\includegraphics[width=0.95\columnwidth]{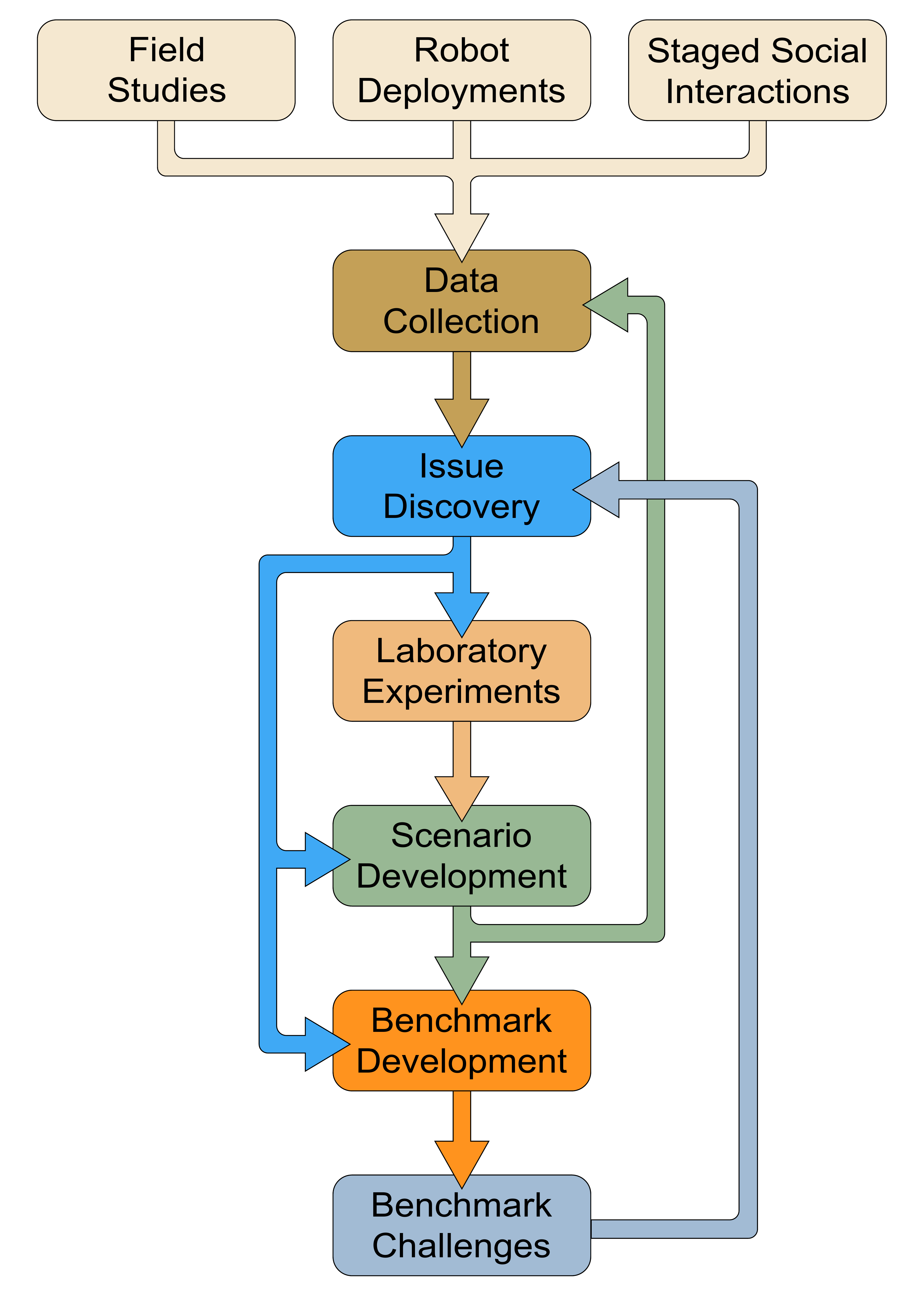}
\caption{Lifecycle of social navigation research. Field studies, robot deployments, and staged social interactions can be used to collect data, which helps identify issues and their prevalence. Issues discovered guide laboratory experiments and the development of social navigation scenarios, which in turn can inform data collection. Issue discovery also helps guide the development of benchmarks that test these issues, along with public benchmark challenges; attempts at solutions of these challenges can also help identify issues. }
\label{fig:lifecycle}
\end{figure}

\subsection{Lifecycle of Social Navigation Research}
\label{s:lifecycle}

Arguably, social navigation research should be driven by data collected from field studies or robot deployments, but these can be prohibitively expensive; conversely, validated social navigation scenarios enable analysis of known problems, but may not cover novel experimental conditions or detect problems that show up in the wild. Rather than focus on one or the other, it is more useful to think of the following lifecycle of social navigation benchmarking:

\begin{enumerate}
\item \textbf{Data Collection:} Field studies, robot deployments and staged interactions can be used for the first step of the scientific process: data collection. Ideally, these should be used for more than just A/B testing; they should be used to generate datasets that can be shared to extend the power of the  social navigation research community to collect data at scale. Data that can be collected includes but is not limited to robot and human behavior, surveys (e.g., subject's opinions on safety or comfort), or biometric data (e.g., heart rate, skin impedance).

\item \textbf{Issue Discovery:} The foundation of social robot navigation is humans interacting with robots. Issue discovery refers to mining human-robot encounter datasets for repeating problematic scenarios that can be reliably detected, enabling the statistical analysis of their frequency and properties. Ideally the focus should be on high-frequency issues (challenging scenarios that often occur, like frontal approach in a narrow hallway or the freezing robot problem) and high-risk issues (challenging scenarios where there is a high risk, like compromising a person's safety). Robot deployments in desired target environments are often the best way to collect this data, but large-scale field studies can serve as a proxy. 

\item \textbf{Laboratory Experiments:} Many scientific questions about human-robot interaction can be conducted even if large-scale datasets or issue statistics do not exist. Research groups not able to conduct large-scale studies or deployments can nevertheless formulate scientific questions and answer them. Where feasible, these experiments should use benchmarking procedures and metrics validated by the research community, such as those discussed in Section~\ref{s:metrics}. Ideally, these should use scenarios identified as frequent issues in the target domain.

\item \textbf{Scenario Development:} One outcome of data collection, issue discovery and laboratory experiments should be the identification of social navigation scenarios which can be reliably detected in datasets, occur frequently in target environments, and can be replicated in controlled laboratory settings. While social navigation scenarios are not a substitute for in-the-wild data collection, using validated social navigation scenarios in laboratory experiments can ensure that experiments are ecologically valid, and can ensure that A/B tests are backed up by regression tests of  known social navigation issues. Scenarios also aid targeted data collection for both analysis of human behavior and generation of datasets for imitation learning.

\item \textbf{Social Benchmarking:} Social navigation scenarios can be composed to create benchmarks for social navigation. Most social navigation benchmarks consist, at least implicitly, of a set of social navigation scenarios, real or simulated, that are used to test robot social navigation behavior, along with metrics to gauge performance; many also define datasets of social navigation behavior for comparisons, and may also provide simulation environments where the scenarios are defined. From a lifecycle perspective, using reliable, validated scenarios frequently occurring in target environments would make a social navigation benchmark more valuable.

\item \textbf{Benchmark Challenges:} Finally, benchmarks can be publicly released as ``challenges" which include success criteria, a call for solutions, methods for collecting and evaluating solutions, and a leaderboard of ranked solutions. Benchmark challenges have been used for a wide variety of embodied AI tasks and have proved useful for promoting improvements in the field, sometimes leading to challenges being solved and retired (see discussion in \cite{deitke2022retrospectives}). The iGibson Challenge \cite{igibson_challenge} was one of the first publicly available social navigation challenges.

\end{enumerate}

\subsection{Guidelines for Real-world Studies}
Real-world social navigation studies have aspects that do not come up in simulated experiments or even traditional navigation experiments. Robots controlled by untested policies can damage themselves, other robots, humans, or their environments; human participants captured by robot sensors have privacy and consent concerns. Here we present guidelines for conducting social navigation experiments in the real:

\textbf{Guideline R1: Preserve Safety.} Real-world benchmarks should preserve the safety of humans, robots, and the environment through active measures such as experiment monitors and safety layers. In particular, policies which have been ablated to illuminate sources of power may have unintuitive behavior; if a safety layer is not available to prevent unsafe actions, either these policies should be tested in simulation, or an experiment monitor should be ready to stop the robot in case of issues.

\textbf{Guideline R2: Respect Human Participants.} The privacy, needs and time of human participants should be respected, along with their informed consent. For academic studies, institutional review board (IRB) approval may be required before proceeding with experiments; for industry studies, an early privacy review can help illuminate potential legal concerns.

\textbf{Guideline R3: Ensure Clear Scientific Objectives.} As real-world experiments are expensive, the purpose of real-world benchmark studies should be clearly defined and the use of the data specified. Conducting an experiment which costs time and money and exposes humans, robots and their environments to risk should be justified with a clear notion of what is to be learned from conducting the experiment.

\begin{figure*}
\centering
\includegraphics[width=0.95\textwidth]{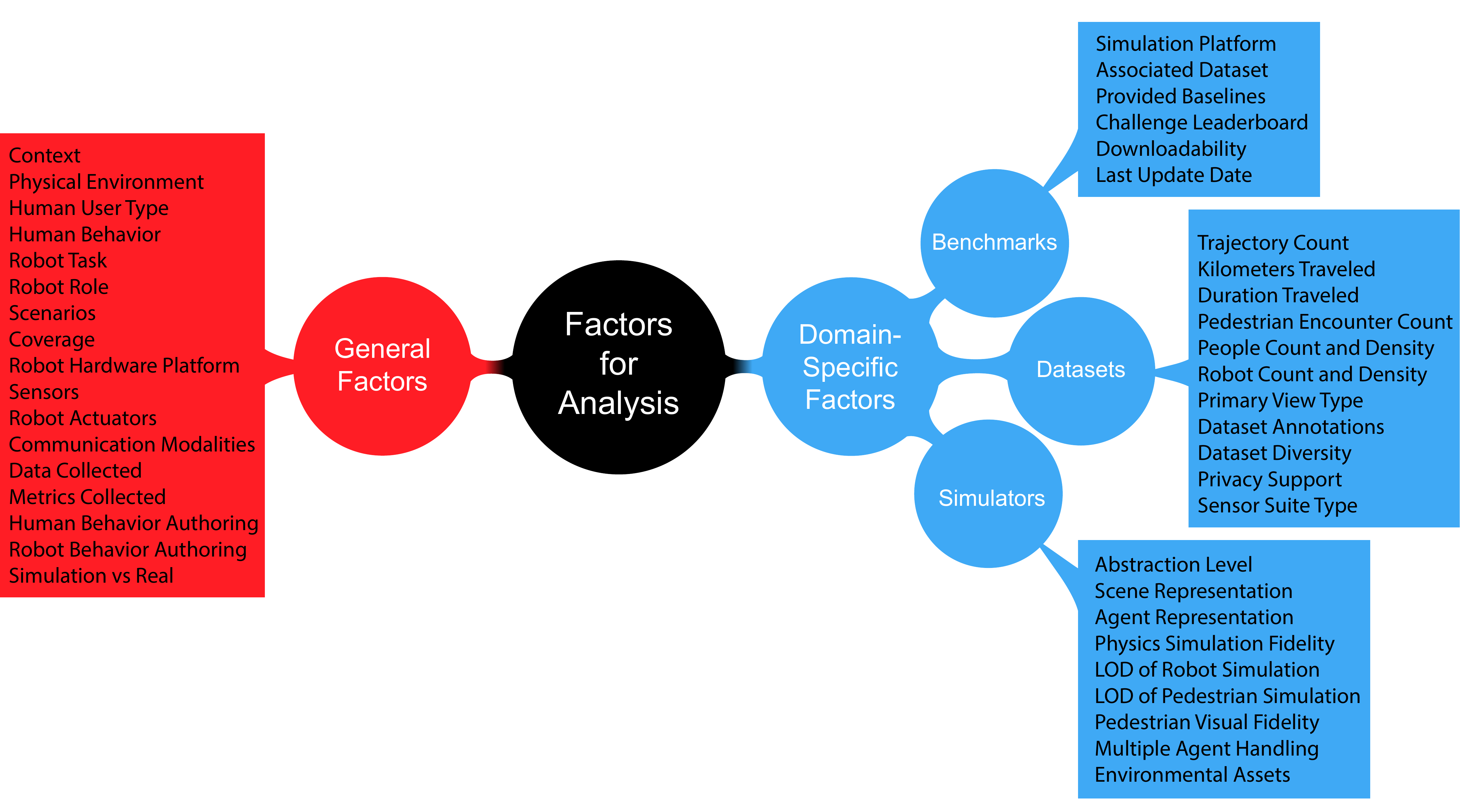}
\caption{A taxonomy of social navigation. Most social navigation instruments share common factors like overall context, physical environments, human user type, robot role and task, and so on. However, datasets, benchmarks and simulators have additional factors particular to them.}
\label{fig:factors}
\end{figure*}

\section{A Taxonomy of Social Navigation}
\label{s:taxonomy}

Creating principles and guidelines that are broadly useful to the social navigation community requires understanding the research efforts already underway. Therefore, 
we have developed a taxonomy for social navigation research in terms of the metrics, datasets, simulators, and benchmarks used, analyzed with a common vocabulary for factors of analysis.

\subsection{A Taxonomy for Analysis}

We propose that social navigation research instruments can be analyzed along a set of formal axes which include the metrics they collect, the datasets that they use, if any, the simulator platforms they use, if any, and any formalized scenarios or benchmarks they use for comparison.

\begin{itemize}

\item \textbf{Metrics:} Social navigation metrics enable us to quantify the performance of algorithms. Recent surveys of social navigation metrics have uncovered close to a hundred different metrics in use (see ~\cite{gao2021evaluation,mirsky2021prevention} for recent reviews). 
Some metrics are algorithmically computed, while others are gathered by surveying humans, either explicitly via questionnaires or implicitly through sensors measuring affect.
Algorithmic metrics in turn can be hand-crafted or learned from data gathered from surveys. Other axes of metrics include the type of variable(s) being modeled, and whether metrics cover the behavior of the robot at a specific point in time (step wise) or during a whole navigation task (episode wise). The nuances of metrics are discussed in depth Section~\ref{s:metrics}.

\item \textbf{Scenarios:} Social navigation studies include field studies of behavior in the wild, long-term robot deployments at particular sites, controlled laboratory experiments, social navigation scenarios that aim to create a particular in-the-wild behavior, and ``staged'' scenarios that attempt to recreate the chaos of crowd scenarios.  We have developed a ``scenario card'' which enables us to compare scenarios, discussed in further depth in Section~\ref{s:scenarios}. 

\item \textbf{Benchmarks:} Social navigation benchmarks involve an evaluation protocol for collecting metrics for social robot navigation methods 
in social navigation scenarios. 
Current benchmarks are discussed in further depth in Section \ref{s:benchmarks}.
``Challenges'' are benchmarks that are publicly available, include success criteria, and provide evaluation mechanisms along with leaderboards to compare solutions; challenges have shown success in other fields in promoting the improvement of the state of the art~\cite{deitke2022retrospectives}. 

\item \textbf{Datasets:} We have used these factors to analyze social navigation datasets, discussed in further depth in Section~\ref{s:datasets}. Note that datasets require additional parameters for analysis such as coverage, sampling distribution, annotations, and privacy and fairness handling.

\item \textbf{Simulators:} Social navigation simulators enable the evaluation of policies controlling agents around other agents in simulation, discussed in further depth in Section~\ref{s:simulation}. Many  simulators have different APIs and metrics. To enable clearer comparison across simulation environments, we are working to create a common simulator API.
\end{itemize}

We next unpack factors common to metrics, scenarios, benchmarks, datasets and simulators before drilling into these topics in more detail in Sections \ref{s:metrics}, \ref{s:scenarios}, \ref{s:benchmarks}, \ref{s:datasets}, and \ref{s:simulation}.

\label{s:factors}

\subsection{Factors Common to Social Navigation}

Participants in the workshop argued that it is important to explicitly define the setups of social navigation benchmarks. But this is equally true of datasets and simulators. Benchmarks, datasets, and simulators for social navigation all face similar challenges: characterizing contexts, representing environments, defining robot roles, tasks, and embodiments, and so on. Rather than analyzing benchmarks, datasets and simulators separately, we argue that many factors are shared among all three, and here present common factors in attempt to create a common vocabulary for analysis.

\begin{enumerate}

\item \textbf{Context:} Broadly speaking, the context of a social navigation endeavor refers to its scope, objective, and intended application. As discussed in Section~\ref{s:principles}, context is a complex construct, and symposium participants did not come to an agreement on a crisp definition, often preferring to use more specific terms when available. However, when it is used, the context of  a social navigation research tool often refers to factors implicit in its definition, e.g., a ``pedestrian outdoor dataset” or a ``benchmark for indoor environments”. Often, the generic concept of ``context'' is cashed out into the currency of environment, human behavior, or robot tasks; simulators may have aspects of the context embedded into their design.
Aspects of context include the scope of a dataset, what a benchmark tests and what it doesn't, and what the focus of experiments are: perception, trajectory forecasting, collision avoidance, algorithm benchmarking, human simulation testing, gesture and gaze interaction, body language and affect sensing, human-robot collaboration, indoors vs outdoors, and individuals vs crowds.

\begin{itemize}
\item Synonyms: Application, Scope
\item Related Factors: Robot Role
\end{itemize}

\item \textbf{Physical Environment:} Although it could be considered part of the ``Context", the physical space in which the robot(s) and humans operate is particularly relevant. The description of the physical environment includes high-level descriptions such as indoor or outdoor and can be as detailed as one desires. For example, "nearby a water cooler in an office space crowded with cubicles". Simulator environment definitions may be scanned from the real or authored. Environment definitions also include constraints such as the layouts and traversability of areas of the scene, as robot objectives and constraints are conditioned on the scene layout. Variations of environments can include indoor (relatively constrained environments) or outdoor (relatively spacious environments). The representation of this may be an explicit scene or map, or may be implicit in the physical layout of the experiment.

\begin{itemize}
\item Synonyms: Location, Scene Type, Context
\item Related Factors: Environmental Constraints
\end{itemize}

\item \textbf{Type of Human User:}
Specifying who the intended or expected human users are is also important. Key is gauging whether the humans are familiar with robots. Humans behave quite differently when they see the robot the first few times, then they get used to it. This type of behavior shift should be noted in a benchmark or dataset since benchmark results obtained by interacting with a group of roboticists may not be representative of when the robot interacts with the public.
\begin{itemize}
\item Synonyms: Human Role
\item Related Factors: Human Behavior, Robot Role
\end{itemize}

\item \textbf{Human Behavior:} A description of the actions taken by specific humans or groups of humans as they relate to the robot. In benchmarks, the desired agent behavior needs to be specified. In simulation, this means the algorithms and scripts that guide the movements of simulated pedestrians. In the real, this means the instructions to human participants; these could range from a scripted setting, where humans are instructed to perform a specific task or trying to go to a specific location, to unscripted scenarios, where the humans are not explicitly instructed how to move. Examples of behavior descriptions include humans navigating to specific  waypoints, humans blocking the robot, or passing. These range from in-the-wild behaviors to carefully specified tasks and everything in-between. Simulated human behavior is currently far more constrained than behaviors in the wild. 
\begin{itemize}
\item Synonyms: Pedestrian Behaviors, Human Tasks
\item Related Factors: Robot Task, Robot Role
\end{itemize}

\item \textbf{Robot Task:} The piece of work assigned to the robot. The typical robot task is navigation from the robot's current location to a goal location. Further, higher-level tasks could be specified, such as the delivery of an object, or guarding of an area in the physical environment. 
\begin{itemize}
\item Synonyms: Robot Behaviors
\item Related Factors: Robot Role, Human Behavior
\end{itemize}

\item \textbf{Robot Role:} The relationship intended between the robot and the humans, e.g., servant, companion, or fellow pedestrian in a space.
\begin{itemize}
\item Related Factors: Robot Task, Type of Human User
\end{itemize}

\item \textbf{Scenarios:} A specific configuration of physical environment, human behavior, and robot task. Scenarios combine three other factors into a package to enable specific configurations of environment, behavior and task of research interest to be shared in the community. A scenario can be as detailed as a scripted interaction, although free-form scenarios, which are unscripted, are also possible. Robot role may be specified as part of a scenario, or it may be a variable which is changed and tested.
\begin{itemize}
\item \textbf{Scenario Classifiers} and \textbf{Behavior Graphs} are methods to automatically extract scenarios from data and/or to provide an unambiguous way of labeling
\end{itemize}

\item \textbf{Coverage:} The breadth and frequencies of scenarios is also important. Datasets, benchmarks and simulators can focus on narrowly specified scenarios, a suite of scenarios, or a broad range of cases. Even if the coverage is broad, the distribution of the tests are important, as is explicit coverage of \textit{corner cases}, such as tests or data collection that include erratic, non-cooperative humans.
\begin{itemize}
\item Synonyms: Edge Cases, Regression Tests
\end{itemize}

\item \textbf{Robot Hardware Platform:} The specific robot morphology, including its shape, sensors, effectors, displays and communications modalities.
Robot hardware platforms can be instantiated in the real world or in simulation or both; while many robots have associated simulators, not every robot is represented in every simulator.
Unifying robot embodiments is unnecessary and likely impossible, as different robot embodiments are used in different contexts. For this reasons, while some benchmarks specify robot embodiments, others are embodiment agnostic.  

\begin{itemize}
\item \textbf{Synonyms:} Form Factor, Platform, Embodiment
\end{itemize}

\item \textbf{Sensors:} The devices which detect or measure physical properties and record, indicate, or otherwise respond to them. Sensing can include on-board sensors only, or may include external sensors or trackers.

\begin{itemize}
\item Synonyms: Inputs, Observation Space
\item Major Divisions: \textbf{Robot Sensor}s on the robot and \textbf{Third-Person Sensor}s in the environment.
\end{itemize}

\item \textbf{Robot Actuators:} What is the action space of the robot? Conceivably, this may also include third-party actuators such as automatic doors, but this usage is rare.
\begin{itemize}
\item Synonyms: Effectors, Action Space
\end{itemize}

\item \textbf{Communications Modalities:} How can humans and robots communicate? Not at all, the robot speaking but not hearing, the robot hearing but not speaking, or two-way? Some possible modalities of communication includes but are not limited to communicating through visual and audio signals, through the body and head motion, or no communication at all.

\item \textbf{Data Collected:} In addition to any robot sensation, actuation, and communication,  benchmarks, datasets or simulators may collect other data such as people tracks, maps of the spaces, and so on. This can include information about pedestrians, such as access to explicit pedestrian states (e.g., position, velocity) or just sensor data; sensor data itself can include third-person sensors like external cameras, or be restricted to the robot's observation space. Pedestrian and robot data can be ground truth (either from a simulator or from motion capture in the real) or noisily extracted with detection and tracking. The range of visibility of pedestrian is also important, as is whether the visibility is restricted to that of robot sensors (including range, occlusions, directionality, and sensing delay) or ground truth (again, from the simulator or non-robot sensors). This is further discussed in Section~\ref{s:datasets}.

\item \textbf{Metrics Collected:} Metrics transform raw collected data into standardized measures with shared definitions that can be compared across different algorithms, robots, and scenarios. Having shared metrics is important for communicating benchmarks, datasets and simulators and is being looked at by several research groups; we present a view of this field in Section~\ref{s:metrics}.

\item \textbf{Human Behavior Authoring Methods:} How are the human behaviors generated for the dataset or benchmark? E.g. real pedestrians, confederates of the experimenters, recordings, simulated via a standard social model, or generated by a policy.
For simulated environments, these behaviors may include non reactive (pedestrians driven by pre-recorded data), reactive (ORCA, social force, or generative models), and animated (character animations including static moving shapes and animated walking); for real environments, these may include natural behaviors, scripted behaviors, or randomized behaviors. For both simulated and real environments, goals of the movement may be random, context-relevant, or goal-directed.
\begin{itemize}
\item Synonyms: Pedestrian Simulation, Crowd Simulation, Microscopic Crowd Simulation
\end{itemize}

\item \textbf{Robot Behavior Authoring Methods:} These are similar to the human behavior authoring methods, except there is no “real robot” class corresponding to “real pedestrians”, just the robot policies under test.
\begin{itemize}
\item Synonyms: Agent Behaviors, Baseline Policies
\end{itemize}

\item \textbf{Simulation vs Real:} Whether the dataset or benchmark is in simulation, in the real, or some combination of both.
Sim and/or real: Is the benchmark operated in the simulation or in the real world? The participants noted that the simulation can be effectively used for issue discovery but cannot replace real world testing.
\begin{itemize}
\item \textbf{Subfactors:} Simulation Fidelity, which ranges from dots in an abstract geometrical space to fully rendered simulations. This includes both \textbf{Human Simulation Fidelity} and \textbf{Robot Simulation Fidelity}, as robots are simulated more often than full humans.
\end{itemize}

\end{enumerate}

While the above factors are common across many social navigation instruments, one size does not fit all: there are additional factors particular to benchmarks, datasets or simulators:

\begin{itemize}
\item \textbf{Dataset Properties} include: trajectory count, kilometers traveled, duration traveled, number of pedestrian encounters, people count and density, robot count and density, primary view type (robot POV, pedestrian POV, third-person POV), dataset annotations, dataset diversity, privacy support, and sensor suite type (moving robot / stationary robot / third-person sensor suite)
\item \textbf{Benchmark Properties} include: the simulation platform, associated dataset, provided baselines, challenge leaderboard, downloadability, and the most recent update.
\item \textbf{Simulator Properties} include: abstraction level, scene representation, agent representation, physics simulation fidelity, level of detail of robot simulation (points, cylinders, robot morphologies), level of detail of pedestrian simulation (with or without gait), pedestrian visual fidelity (basic meshes, movements, photorealistic), handling of multiple agents (flow-based crowd, agent-based individuals), environmental assets (realistic scenes or simulated layouts; indoors, outdoors or abstract scenes).
\end{itemize}

\section{Social Navigation Metrics}
\label{s:metrics}

Unlike traditional navigation, where the community largely agrees on a few evaluation metrics, such as Success weighted by Path Length (SPL)~\cite{anderson2018evaluation}, finding a consensus for social navigation metrics is challenging. 
One reason for this difficulty is that we care about multiple aspects of human-robot encounters in social robot navigation, e.g., how safe a robot's behavior is nearby people and how well the robot communicates its intent in order to facilitate motion coordination. Measuring any one of these factors from a human perspective is difficult, let alone deciding how to combine them into a single metric.

For example, while safety is a generally agreed upon factor that drives the implementation and evaluation of social navigation systems, safety is a complex construct~\cite{lasota2017survey}. While one can think of physical safety in terms of collisions, as is often the case in the broader robot navigation literature (and is captured in Principle P1, Safety), safety also can be viewed from a psychological standpoint~\cite{kamide2012new} (which might be captured in Principle P2, comfort), or even in terms of not disrupting social and moral values~\cite{boddington2017epsrc} (which might be captured in Principle P5, Social Norms). Careful thought must be put into even obvious terms as the context of their usage may change their meaning (Principle P7, Contextual Appropriateness).

The next section provides a taxonomy of social navigation metrics, followed by a discussion of the challenges of measuring social navigation. We then present recommendations on metrics for social navigation, along with guidelines for using metrics to evaluate the success of social navigation systems.

\subsection{Taxonomy of Existing Social Navigation Metrics}

In the past years, a wide range of metrics have been proposed to quantitatively measure key aspects of social robot navigation and allow for fair comparison among social navigation solutions (see ~\cite{gao2021evaluation,mirsky2021prevention} for recent reviews). We describe three ways social navigation metrics can be classified according to, a) their nature, b) the variable being modeled and, c) their temporal scope. To fully classify a metric, it should therefore be classified according to the three taxonomies.

\begin{figure}
    \includegraphics[width=0.95\columnwidth]{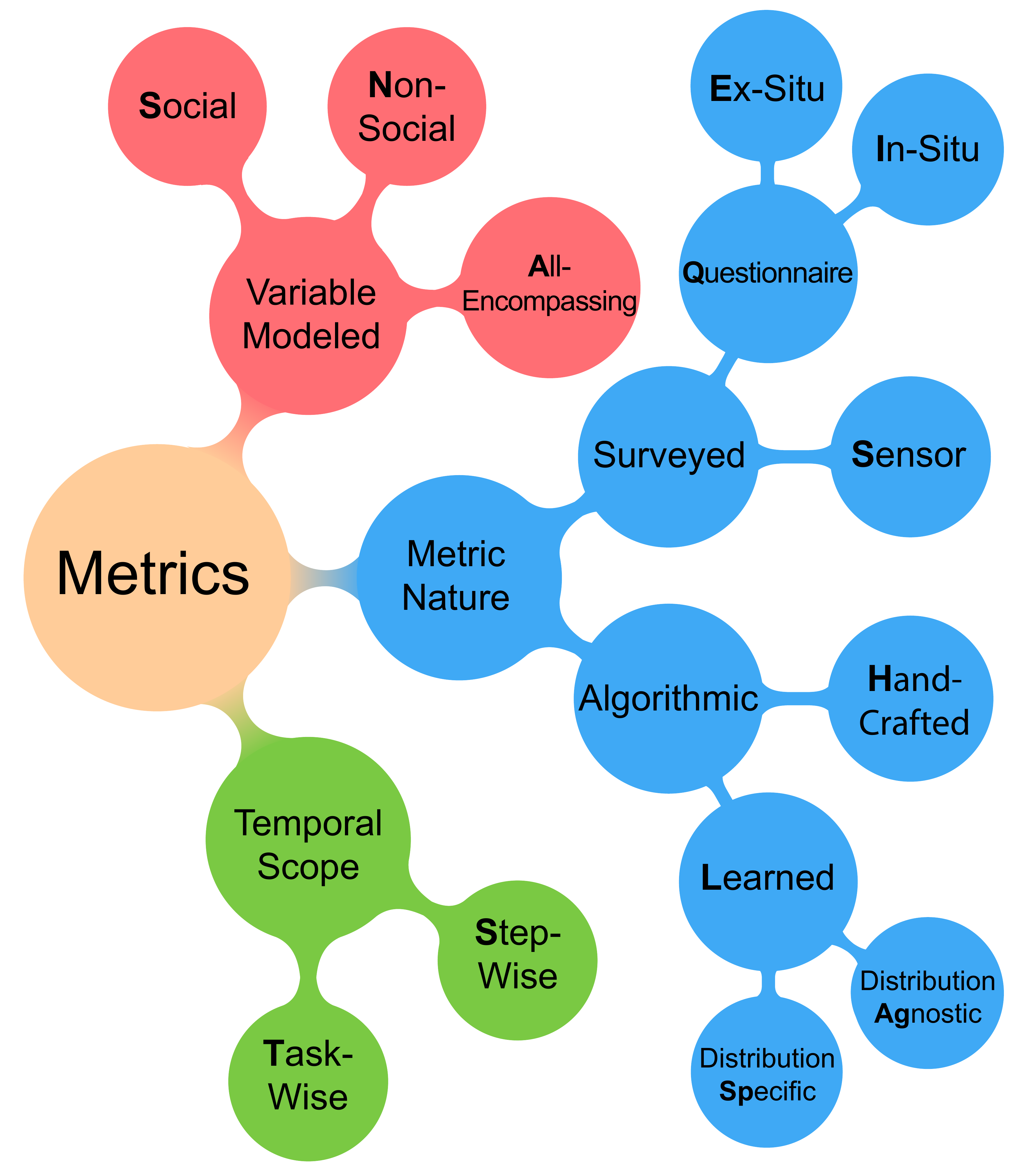}
    \caption{The proposed taxonomy suggests classifying metrics according to three aspects: the type of variable (or variables) they model, their nature, and their temporal scope. To quickly identify metric types, we suggest using a three-letter code based on these factors. For example Success Rate (SR) is a Non-Social, Hand-Crafted, Task-Wise metric, or NHT metric; a sensor metric gauging moment-to-moment human facial reactions to robot behavior would be a Social, Sensor, Stepwise metric, or SSS metric; and a questionnaire asking about the overall quality of a robot's navigation would be an All-Encompassing, Questionnaire, Task-Wise metric, or AQT metric. }
    \label{fig:metricTaxonomy}
\end{figure}

\subsubsection{Taxonomy based on their nature}
We can distinguish two main groups of metrics, those that are algorithmic, and those that are not computed but surveyed (see Fig.~\ref{fig:metricTaxonomy}).

\textbf{Surveyed} metrics are usually human ratings of desired properties of social robot navigation, e.g. safety, comfort, or legibility. They can be classified into \textbf{questionnaire-based} (in-situ or ex-situ), where the ratings are explicitly requested, or \textbf{sensor-based}, where the ratings are transduced from sensor data.
Although surveyed metrics are (arguably) the best way to measure social navigation success, they are expensive, difficult to scale, and time-consuming.
While small-scale human studies are commonly conducted, results can have high variance and can be non-reproducible.
To address this shortcoming, researchers have also created a variety of \textbf{algorithmic} metrics that serve as proxies for surveyed metrics. These algorithmic metrics are cheap to compute and reproducible, properties that are key for benchmarking. Unlike traditional navigation, where SPL~\cite{anderson2018evaluation} is a commonly accepted metric, social navigation has no single metric of reference. Instead, method comparison is usually performed using multiple metrics.
\par
Algorithmic metrics can be subsequently classified into hand-crafted and learned, based on whether they are the result of intuition and experience, or modeled using statistical analysis or machine learning.
\textbf{Hand-crafted} metrics are objective (\textit{i.e.}, in what they compute, not necessarily their interpretation), scalable and can be easily computed given certain assumptions, yet oftentimes they cannot fully capture the desired property of social robots.
\textbf{Learned metrics} can be considered a compromise between survey-based and hand-crafted metrics.
These evaluative models can be trained on large-scale offline datasets of human ratings and then used to score robot behavior. They are reproducible and have minimal inference cost, but compiling the necessary datasets can be very time consuming.
Learned metrics can be further split into \textbf{distribution-specific} metrics, which rely on assumptions on the properties of the dependent variable to model~\cite{perez2014robot}, and \textbf{distribution-agnostic} metrics, which aim to model these variables without making any relevant assumptions~\cite{manso2020socnav1,bachiller2022graph}.

\subsubsection{Taxonomy based on the variable being modeled}
Regardless of the nature of the metrics, algorithmic or surveyed, learned or hand-crafted, the variables they model can refer to different phenomena. Most common metrics assess either social or non-social aspects, but a metric could also combine both into an all-encompassing metric (see Fig.~\ref{fig:metricTaxonomy}).
\textbf{Non-social} metrics have generally been developed with PointGoal navigation in mind and focus on aspects such as path and energy efficiency or success rate. They generally have the advantages of being objective, reproducible, and are usually fast to compute, but do not provide any social performance information.
\textbf{Social} metrics focus on one or more social aspects of robot navigation, such as comfort, acceptance, trustworthiness, or predictability.
\textbf{All-encompassing} metrics aim to model the overall scores humans would provide in robot navigation, considering both social and non-social aspects. Although these metrics would arguably be the most desirable, very few have been proposed~\cite{chiang2019learning,bachiller2022graph}. 

\subsubsection{Taxonomy based on temporal scope}
It is also useful to consider metrics' temporal scopes, as they determine where a metric can be applied.
Here we distinguish task wise and step wise metrics (see Fig.~\ref{fig:metricTaxonomy}).
\textbf{Step wise} metrics provide a score per time step, and are well suited for path planning.
\textbf{Task wise} metrics are the most appropriate for benchmarking social navigation algorithms, as they provide a single score per task.
Step wise metrics can potentially be combined into task wise metrics, such as by averaging across all steps within a task.
However, not all moments within a social navigation task have an equal impact on social performance.
To address this, task wise metrics can also combine step wise data with temporal data to capture features such as reversals in step wise metrics over time, more heavily weight task-relevant periods of the task, or measure how long a robot was able to navigate with high quality step wise metrics \cite{taylor2022observer}.
For reinforcement learning-based social navigation, task wise metrics (AAT specifically, according to the proposed taxonomy) can be preferable over step wise metrics, depending on their properties. Although using task wise metrics would produce delayed rewards, a step wise metric would only be advisable if its cumulative value reflects task performance, which is generally not the case. 

\subsection{The Challenges of Measuring Social Navigation}
The evaluation of robot navigation has evolved as the field has matured, moving from success metrics to quality metrics to social metrics. Early work focused on success metrics that gauged whether the robot did its task, such as success rate or kilometers without incident \cite{marder2010office}. Later work proposed quality metrics that gauged how well the robot did its task, such as SPL~\cite{anderson2018evaluation}. More recent work such as \cite{wang2022metrics} proposes social metrics that gauge how the robot behavior affects other agents, such as personal space compliance (PSC)~\cite{igibson_challenge} or questionnaire-based metrics~\cite{pirk2022protocol,xiao2022learning,cuan2022gesture2path}. 

However, because social metrics involve robots interacting in complex real-world environments with humans whose learning changes their behavior over time, several additional factors must be considered to evaluate these social metrics accurately and reliably. These include (1) the challenges of dynamic environments, (2) the impact of long-term exposure on study participants, (3) how robot behavior may be changed by deployments, and (4) limitations of metrics themselves.

\subsubsection{The Challenges of Dynamic Environments}

When measuring the performance of a social robot, an important consideration is the dynamic nature of the \textbf{environment} and of the other \textbf{pedestrians} around. These elements are often controlled when performing in-lab studies, but evaluation in the wild is much more intricate, especially when looking at longer periods of time, as robots become more and more capable of long-term deployment \cite{biswas20161}.
Results of performance measures like accuracy might be affected by simple changes such as lighting conditions and weather. Speed may be similarly affected by the percentage of remaining battery. While almost impossible to mitigate such effects, they should be acknowledged and highlighted when reporting relevant results.

\subsubsection{Challenges Based on How Robots Change People}
Yet, a more challenging aspect to measure is the dynamic nature of pedestrians when interacting, even casually, with a robot \cite{rosenthal2020forgotten}.
When people interact with a navigating robot for the first time, they adapt their beliefs and expectations to the robot's behavior, which can cause immediate improvement in various metrics such as fluency, time, and efficiency. 

This phenomenon can be leveraged to train pedestrians around robots, rather than adapting the robot to the pedestrians. One such use of passive demonstrations was shown to significantly reduce the number of conflicts between a person and a robot passing one another in a hallway, by having the robot demonstrate in advance how it signals its intentions \cite{fernandez2018passive}.

Beyond the novelty effect of first encounters, people will refine their behavior around a robot as they interact with it over time. A person will behave and react differently to a navigating robot on the tenth interaction than the hundredth \cite{gockley2005designing, hart2022longitudinal}. More research is needed on how people adapt to the presence of navigating robots  \cite{leite2013social}, but studies of other social interactions such as asking favors \cite{saunderson2021robots} and information delivery \cite{kim2013effects} indicate that adaptation is likely.

When conducting research on social navigation in academia, it is not uncommon to rely on students, especially those with STEM and robotics backgrounds, as participants in an empirical study; this reliance on students has long been known to the psychological field as a potential source of bias \cite{smart1966subject,schultz1969human}. Measuring acceptance, animacy, and fluency can all be affected by this biased population that has been exposed to robots as part of their studies. Moreover, as robots are being deployed around campus, other students are also being exposed to these robots and thus over longer periods of time might also be biased in their expectations regarding the behavior of robots, based on their past encounters. Industry researchers in the symposium also reported differences of socialness ratings between naive subjects and robot researchers, and even between robot researchers and experienced robot ``wranglers'' who logged far more hours of direct robot time.

\subsubsection{Challenges Based on How People Change Robots}
When a robot is deployed for a long period of time, people may become familiar with it and thus more willing to accept risky behavior from it: it may be able to drive at higher velocities, which will affect speed measurements, or it might get closer to others, which may affect efficiency and acceptance. Some symposium participants noted multiple instances of robots colliding with visitors after a good track record of avoiding collisions around the development team; a postmortem revealed that this was likely due to the development team implicitly learning to keep a collision-free distance.

Moreover, people's attitudes towards the robot over longer periods may require additional metrics that better capture how people perceive a long interaction with a robot.
For example, in a long-term study of a socially assistive robot, the faults of the system did not affect the overall acceptance of the system by the participants \cite{feingold2021robot}. Similar phenomena are likely to be observed in a social navigation context and thus should be reasoned about when measuring long-term interactions. 

\subsubsection{The Limitations of Social Metrics Themselves}
\label{metrics:challenges}
In addition to these concerns, metrics themselves have challenges, including subjectivity and scaling, relevance and weighting, and the transferability of results between robot morphologies. 

\begin{enumerate}
	\item \textbf{Human ratings are subjective.} Human ratings are by their very nature subjective, and they depend on many factors such as cultural context, environmental context, goals or priorities within a scenario, or their overall familiarity with robots. It is important to account for the factors that all human participants experience, as well as attempt to characterize unique factors relevant to the scenario that can affect how they perceive the scenario.

	\item \textbf{Subjective metrics are difficult to scale.} Expanding to a larger participant pool can help to mitigate variations between individuals, but it can be hard to execute complicated scenarios with a large number of participants. Creating analytical models of certain sub-elements of human reactions, such how comfortable observers are with the proximity of the robot, can potentially be done with studies of a more targeted scope, and then used in broader models of human responses to robot behavior. 

    \item \textbf{Real-world evaluation is difficult to scale.} The closer a study can get to emulating a real-world scenario such as a busy street, a crowded airport, or a packed restaurant, the more it can capture the effectiveness of a robot in this domain. However, creating these scenarios in a laboratory environment is difficult. Eliciting natural behavior can be challenging, and many social environments have a large volume of people entering and exiting that can be hard to represent. Therefore, efforts are being made to record natural human behavior for use in simulations to address this issue, along with blending multiple metrics to account for the many aspects of a real-world deployment.

	\item \textbf{Choosing which variables are relevant.} Measuring all possible signals humans generate in response to a robot is difficult. However, selecting any subset can neglect other useful signals. For example, using only 2-dimensional poses disregards other very important inputs such as face expressions, gestures or gaze \cite{hart2020using}. Putting thought into which signals are most relevant to a scenario and able to be robustly collect them is important.

	\item \textbf{Weighting Multiple Metrics.} The variety of useful metrics and their context dependence suggest applying an ensemble of metrics, weighted to account for the parameters of a specific scenario. The optimal method for doing this, however, remains open. It is worth considering if this weighting may vary not only across different environments but also over the course of a single path as the audience or priorities of the robot change.

    \item \textbf{Non-homogeneous hardware.} Robots have varying sensors and actuators. While some only have access to their wheels' motors and a LiDAR, others can inform pedestrians of their presence and intentions, or share information using sound and visual cues. It is difficult to consider these additional aspects analytically, so standardized metrics do not take them into account. Unfairly, this limitation can make robots able to share such information appear less socially capable than they are.
 \end{enumerate}

\subsection{Assessment of Existing Social Navigation Metrics}

For all the reasons outlined above, quantifying the quality of different social navigation strategies is difficult. A social scenario can include many different stakeholders with varying priorities, and context is extremely important. For example, a passerby may primarily be focused on metrics of discomfort as the robot passes them, while the recipient of a handoff may prefer more detailed information about the robot’s movements. A warehouse may focus more on expediency, while in a restaurant excess speed or urgency may be unnerving. Social preferences also vary across cultures and groups.

In all of these cases, subjective metrics reported by humans directly experiencing these scenarios is the gold standard. This can be difficult, however, in terms of scaling the number of participants. A secondary issue is that the higher the density of feedback requested, the more disruption to the social scenario being measured. Both of these issues increase the demand for analytical or learned subjective metrics, and we discuss the considerations for this in section \ref{metrics:challenges}.

New metrics are often created to address issues that come up in new scenarios, and as social navigation is being deployed in increasingly many new environments, more metrics are being created to address these scenarios. It is also unsurprising that new metrics will be of particular value in the environments that demanded their creation. This means that the number of metrics available to assess performance can be daunting, and their value very context-dependent.

After reviewing the related literature, we did not find a convincing method to quantitatively compare different metrics to determine whether one is strictly better than the other. We suggest that when using any social navigation metric it is essential to note both the metric's original context and the current one it is being applied to. As mentioned, survey-based metrics are generally preferred for benchmarking, though their results are difficult to reproduce if they are not run correctly and they are resource intensive. All-encompassing learned metrics would be the next best option for benchmarking, but unfortunately, none of the existing ones (see \cite{perez2014robot,manso2020socnav1,bachiller2022graph}) satisfy the requirements of all applications and scenarios.
Metrics focusing on specific phenomena are of great importance when debugging and diagnosing an algorithm's flaws, but are sometimes are difficult to use to compare disparate algorithms.

\begin{table*}
\begin{center}
\begin{tabular}{ |L{0.2cm}|C{2.3cm}|C{0.7cm}|L{6cm}|C{0.8cm}|L{1.5cm}|C{1.0cm}|C{0.8cm}|C{0.7cm}|} 
\hline
    & \textbf{Metric} & \textbf{Short} & \textbf{Description} & \textbf{Class.} & \textbf{Parameters} & \textbf{Unit}  & \textbf{Range} & \textbf{Cited} \\
\hline
    \multirow{11}{*}{\rotatebox[origin=c]{90}{Success metrics\hspace{1cm}}}& Success  & $S$ & Binary variable describing whether the robot reaches the goal.  When averaged, it is referred as Success Rate (SR). & NHT & - & boolean          & $\{0, 1\}$  & \cite{anderson2018evaluation}   \\
\cline{2-9}
    & Collision & $C$       & Number of collisions in the trajectory. When averaged it is referred to as Collision Rate (CR).  & NHT  & Collisions~to terminate episode & collision & $[0,\infty)$ & \cite{kastner2022arena}  \\
\cline{2-9}
    & Wall Collisions & $WC$ & Number of collisions against walls.  & NHT                  & - & collision & $[0,\infty)$ & R@G \\
\cline{2-9}
    & Agent Collisions & $AC$ & Number of collisions against humans or robots.  & NHT                 & - & collision & $[0,\infty)$ & R@G \\
\cline{2-9}
    & Human Collisions & $HC$ & Number of collisions against humans. Also called H-collisions~\cite{cancelli2022exploiting}. & NHT                  & - & collision & $[0,\infty)$ &  R@G \cite{cancelli2022exploiting} \\
\cline{2-9}
    & Timeout~before reaching goal & $TO$ & Binary variable accounting for failures caused by a timeout.  & NHT & Time threshold & timeout &  $\{0, 1\}$ & R@G \\
\cline{2-9}
    & Failure to progress & $FP$ & Number of failures caused by not decreasing the distance to the goal for a given period of time.  & NHT & Distance~\& time thresholds & failure &$[0,\infty)$ & R@G \\
\cline{2-9}
    & Stalled time & $ST$ & Time where the magnitude of the speed of the robot falls within a given threshold. & NHT & Distance~\& time thresholds & $s$ &$[0,\infty)$ & \cite{tsoi2022sean} \\
\cline{2-9}
    & Time to reach goal & $T$ & Time between task assignment and completion. & NHT & - & $s$ & $[0,\infty)$ & \cite{francis2020long,kastner2022arena}  \\
\cline{2-9}
    & Path length & $PL$ &  Length of the trajectory. & NHT & - & $m$ & $[0,\infty)$ & \cite{francis2020long,kastner2022arena} \\
\cline{2-9}
    & Success weighted by path length & $SPL$ & Success weighted using normalized inverse path length, \textit{i.e.}, weighted using path length divided by the max of the min distance and path length~\cite{anderson2018evaluation}. & NHT & - & success & $[0,1]$ & \cite{anderson2018evaluation} \\
\hline
    \multirow{8}{*}{\rotatebox[origin=c]{90}{Quality and social metrics\hspace{1cm}}} &Velocity-based \mbox{features} & $V_{min}$, $V_{avg}$, $V_{max}$& Minimum, average and maximum linear velocity on a trajectory. & SHT & - & $m/s$ & $(-\infty,\infty)$ & \cite{kastner2022arena} \\
\cline{2-9}
    & Linear acceleration based features & $A_{min}$, $A_{avg}$, $A_{max}$& Minimum, average and maximum linear acceleration on a trajectory. & SHT & - & $m/s^2$ & $(-\infty,\infty)$ & \cite{kastner2022arena} \\
\cline{2-9}
    & Movement jerk & $J_{min}$, $J_{avg}$, $J_{max}$ & Minimum, average and maximum linear jerk (\textit{i.e.}, the second-order derivative of the linear speed). & SHT & - & $m/s^3$ & $(-\infty,\infty)$ & \cite{kastner2022arena} \\
\cline{2-9}
    & Clearing distance & ${CD}_{min}$, ${CD}_{max}$ & Minimum and average distance to obstacles in a trajectory. & SHT & - & $m$ & $[0,\infty)$ & \cite{kastner2022arena} \\
\cline{2-9}
    & Space compliance & $SC$ & Ratio of the trajectory with the minimum distance to a human under a given threshold. If the threshold is $0.5m$, it is referred to as Personal Space Compliance ($PSC$)~\cite{li2021igibson}. & SHT & Distance threshold & $m$ & $[0,1)$ & \cite{igibson_challenge} \\
\cline{2-9}
    & Minimum distance to human & $DH_{min}$ & Minimum distance to a human in a given trajectory. & SHT & - & $m$ & $[0,\infty)$ &  \\
\cline{2-9}
    & Minimum time to collision & $TTC$ & Minimum time to collision with a human agent at any point in time in the trajectory, should all robots and humans move in a linear trajectory. & SHT & - & $m$ & $[0,\infty)$ & \cite{biswas2022socnavbench}\\
\cline{2-9}
    & Aggregated Time & $AT$ & Time taken for a subset of cooperative agents to meet their goals. & SHT & Cooperative agents' set & $t$ & $[0,\infty)$ & \cite{wang2022metrics}\\
\hline
    \end{tabular}
\end{center}
\caption{Suggested hand-crafted metrics for the evaluation of social navigation systems. The first tranche in the table are traditional navigation metrics, included to ensure that social navigation systems do not regress on traditional navigation performance; the second tranche concern aspects of the quality and socialness of navigation. Citations refer to either papers or challenges defining the term, or R@G for metrics from an unpublished Robotics at Google\cite{robotics-at-google} robot deployment.}\label{t:metrics}
\end{table*}

\subsection{Recommendations for Metric Usage and Development}
\label{s:metric-recommendations}

While many in the symposium argued that surveyed metrics are the gold standard, others pointed out that they are challenging to get right, expensive to collect and sometimes inappropriate (e.g., for evaluating ablation studies where safety cannot be guaranteed). Learned metrics have been proposed as a solution, but are not ready for adoption. Therefore, to measure social robot navigation, we recommend a balanced approach, involving a common subset of hand-crafted metrics, recommendations for the iterative validation of surveys, and suggestions for future metric development.

\subsubsection{Recommendations for Hand-crafted Metrics}

Although surveyed metrics are the arguably the most meaningful and reliable metrics if the required surveys are carried out adequately, they are expensive and time consuming. Additionally, no task-wise learned metrics are yet available. 

Therefore, to ensure a systematic and objective comparison of social navigation algorithms we suggest using a subset of existing hand-crafted navigation metrics. The suggestion includes success-related metrics accounting for success itself, collisions, and failures, as well as metrics related to trajectory properties and social aspects. These recommended metrics can be found in Table~\ref{t:metrics}, along with descriptions of the phenomena accounted for, their required parameters, units, ranges and references where a full mathematical definition can be found.

A relevant characteristic of many of these metrics is that their values, and more importantly what would be considered good ones, heavily depend on the task and the context where the experiments take place and the parameters of the metric (see Table~\ref{t:metrics}). It is therefore good practice to explicitly state the parameters used and context when reporting results.
\par

It is also worth noting that the metrics in Table~\ref{t:metrics} are frequently reported as averages for a number of experiments rather than for a single trajectory (\textit{e.g.}, Success (S) is often found as the Success Rate (SR)). When reporting experimental results for multiple trajectories, providing distributional information in addition to averages allows to show valuable information, including outliers. This is key when consistency is important, as it is the case of safety. Distributional information can be provided, for instance, as histograms.

\subsubsection{Recommendations for Survey Development}
Gathering human perception with surveys has a long history in human-robot interaction (see for example the discussion in \cite{spatola2021perception}), but there is not yet a unified approach to questionnaire development in social robot navigation. Following the social scenario development approach of \cite{pirk2022protocol,xiao2022learning,cuan2022gesture2path}, we recommend an iterative approach in which versions of questionnaires are proposed and then empirically tested to determine their validity ~\cite{morling2014research}.

While survey validity is a complex topic worthy of its own book \cite{litwin1995measure}, several concerns for the design of questionnaires include assessing test-retest reliability (whether a survey gives stable results over time), construct validity (whether a survey measures what it purports to measure), and sources of bias (distorting factors that make the results hard to interpret). Assessing these factors involves reviewing both individual questions and the design of the survey as a whole.

For surveys as a whole, the longer a survey is, the less reliable the answers are \cite{backor2007estimating,galesic2009effects}, and the more frequently surveys are given, the less likely people are to participate \cite{porter2004multiple} a phenomenon known as \textit{survey fatigue} or more generally \textit{response burden} on participants. Reducing response burden is important not just to improve the quality of results but to respect the time of participants; nevertheless, issuing surveys multiple times can help measure test-retest reliability, issuing surveys to multiple populations can help measure bias, and including redundant questions can help measure construct reliability and question utility.

For individual survey questions it is important to ask them using techniques which have been validated. For example, Likert scales \cite{likert1932technique} are a widely-used technique which provide a range of options like ``Strongly Agree, Agree, Disagree, or Strongly Agree.'' While it is tempting to use consistent wording between questions, to reduce cognitive load on participants it is arguably better to formulate Likert scale responses so they form direct responses to each question, along with an option to indicate the question is not applicable. For example, to assess Principle P1, Safety, a question might ask ``How safe was the robot's motion?'' and give the responses ``Unsafe, Somewhat Unsafe, Somewhat Safe, Safe, or Not Applicable.'' To evaluate overall navigation quality, some researchers have explored Likert scales similar to performance-based employee rating systems (e.g., ``Outstanding, Very satisfactory, Satisfactory, Unsatisfactory, Poor''\footnote{\url{https://helpjuice.com/blog/employee-evaluation-form}}) but no consensus yet exists here.\footnote{\url{https://www.performyard.com/articles/performance-review-ratings-scales-examples}}

Statistical analysis of experiments is discussed in depth in standard textbooks such as \cite{cohen1995empirical,morling2014research}, but we highlight some key concerns for social navigation. Terms such as ``significance'' often refer to \textit{statistical significance}, a specific and contentious term in psychological literature~\cite{lykken1968statistical} which should not be used unless the proper statistical tests are conducted. To do so, experimental conditions tested should be properly balanced counts (especially if questions are presented in multiple orders to reduce first-response bias, which creates sub-conditions within the experiment). Properly balanced experiment conditions enable the analysis of variance with tools like ANOVAs~\cite{cohen1995empirical,tabachnick2007experimental,morling2014research} and Cronbach's alpha \cite{tavakol2011making,morling2014research}. Cronbach's alpha in particular can help determine whether a given question is a reliable factor (see for example the discussion in Appendix D.4 of \cite{xiao2022learning}) or should be dropped in future surveys in favor of more reliable questions.

\subsubsection{Recommendations for Future Metric Development}
Because conducting human surveys is expensive, symposium participants expressed interest in finding hand-crafted or learned algorithmic proxies. For example, to gauge safety, some benchmarks measure `time-to-collision' (TTC)~\cite{biswas2022socnavbench}. To gauge comfort, some researchers \cite{truong2017toward, truong2017approach} have proposed some metrics to measure and limit the unnecessary motion and direction changes by the robot in the presence of humans; others have proposed `visibility indices' which gauge the distance and angle at which robots first impinge on a human's field of view~\cite{singamaneni2022thesis,singamaneni2023towards}. Legibility is also highly connected to field of view, as observers need to be able to see a robot to make inferences about its movements and goals\cite{taylor2022observer}.

Future metric development should continue to explore learned or hand-crafted algorithmic proxies for surveyed metrics which can be efficiently computed, enabling the development of more efficient, repeatable and scalable benchmarks. Validating these metrics might require collecting and annotating a large-scale dataset with both algorithmic and surveyed metrics, which could be used to compute the correlations between algorithmic proxies and their surveyed counterparts. This dataset could also be used to learn metrics which capture the surveyed results, as done in \cite{manso2020socnav1,bachiller2022graph}. Another approach to learning social metrics could be AutoRL~\cite{chiang2019learning}, which learns dense reward functions useful for learning based on a sparse true objective; conceivably, data from surveys could be used as the true objective to train a learned social reward.

\subsection{Metric Guidelines}

In general, social navigation systems should not just be good social systems, but robust navigation systems, with a high success rate, low collision rate and a good SPL to ensure efficient experiments and the safety of human participants. Many of these features can be determined in simulation before deploying policies on potentially dangerous robots, but how social these policies are can only be determined with reference to human reactions to robot behavior - either through direct human surveys, or learned metrics derived from human data.

Our recommendations for social metrics expand on these insights and summarize our broader recommendations from Section~\ref{s:metric-recommendations}: use a broad set of navigation metrics to ensure robustness, attempt to use human survey metrics where feasible to evaluate socialness, validate those metrics with standard tools, guard against sources of bias, but use metrics appropriately in each stage of development.

\begin{enumerate}
    \item \textbf{M1 - Ensure robustness using standard metrics:} To ensure social navigation algorithms are good navigation systems, evaluations should report as many of the standard metrics of Table~\ref{t:metrics} as feasible. 
    \item \textbf{M2 - Validate policies with algorithmic metrics in simulation:} Prior to deployment, algorithmic metrics such as those in Table~\ref{t:metrics} can enable fast evaluation to filter out bad policies prior to deployment.
    \item \textbf{M3 - Parameterize metrics appropriately in context:} Social metrics with parameters, such as failure to progress or space compliance, should be appropriately parameterized given the current context, and parameters should be reported for those metrics that require them (see Table~\ref{t:metrics}).
    \item \textbf{M4 - Use learned metrics to help iterate on behavior:} Where learned metrics based on human data are available, they can provide insights to improve robot behavior, or acceptance tests prior to  deploying policies on robot.
    \item \textbf{M5 - Use validated surveys to evaluate social performance:} Human surveys using validated instruments should be used to test the social navigation scenarios once the system is sufficiently robust and reliable.
    \item \textbf{M6 - Set up experiments consistently to avoid bias:} Environmental complexity, subject selection, robot familiarity, survey fatigue and differing experimental setups can all distort metrics. Use well-designed scenarios  (see Section~\ref{s:scenarios}) to make metrics easier to compare.
    \item \textbf{M7 - Analyze experiments iteratively:} Social contexts are complex and getting metrics and surveys right are difficult; therefore, researchers should analyze experiments and iteratively improve them.
    \item \textbf{M8 - Report results in depth:}  Pointwise estimates of single metrics can provide a distorted view of the performance of a system. Experimenters should report a battery of traditional, learned, and surveyed metrics, including both stepwise and taskwise metrics, as well as histograms or other distributional information.
\end{enumerate}

\section{Social Navigation Scenarios}\label{sec:social-scenarios}
\label{s:scenarios}

Social navigation scenarios are specifications of categories of human-robot interactions that facilitate the collection of data on human-robot behaviors and the communication of that data between researchers in a common language. 

Fundamentally, social robot navigation involves robots interacting with humans. The situations in which we study these interactions range from controlled scenarios in the laboratory with small numbers of humans and robots to large-scale in-the-wild studies with dozens of robots in many uncontrolled pedestrian encounters. Following the symposium, participants engaged in substantial discussion regarding the relative importance of studies along this spectrum.

\begin{itemize}
    \item Proponents of large-scale in-the-wild studies argue they have good ecological validity, uncover long-tail behaviors, enable more reliable assessments of human perceptions of robot behavior, and enable data collection for unsupervised and reinforcement learning. These studies can have good statistical reliability and can generate large datasets; however, they are expensive, time-consuming, require heavyweight software architectures, and are suitable for policies that are already reliable.
    \item Proponents of controlled in-the-lab scenarios argue they can also have good ecological validity, prevent regressions on known issues, enable scientific analysis of algorithms and behavior, and enable data collection for supervised and imitation learning.  These studies are cheaper to run, generate data quickly, require less complex software, and are more appropriate for iterating on policies in an earlier stage of development; however, it is harder to uncover long-tail behaviors or to generate large datasets.
\end{itemize}

Social navigation scenarios are a research tool to help bridge the gap between in-the-wild studies and controlled laboratory experiments by defining a clearly specified set of scenarios which can be identified in data collected in the wild, set up as experiments in the lab, and analyzed consistently based on the common definition. For example, the commonly-used \textsc{Frontal Approach} scenario (Table III), which involves a robot and a human traveling in opposite directions in an environment large enough for them to pass each other, could be used to in field studies, robot deployments, laboratory experiments, and even imitation learning:

\begin{itemize}
    \item \textbf{Field Studies}: A \textsc{Frontal Approach} definition could be used to identify human-robot interactions in data collected from an in-the-wild field study, perhaps using the Behavior Graph method for analysis to distinguish them from other interactions such as intersections or overtaking. This suggests social navigation scenarios should be construed broadly so that long-tail behavior can be analyzed. For example, if during a \textsc{Frontal Approach} a pedestrian trips and is helped up by the robot, the pedestrian and robot may not exit the environment normally, but this is nevertheless an event that happens in \textsc{Frontal Approach} scenarios and should be captured in the data.
    \item \textbf{Robot Deployments}: A \textsc{Frontal Approach} definition could be used to set up a deployment to elicit desired behaviors - for example, a robot could be deployed traveling back and forth on a well-trafficked corridor. Thus, social navigation scenarios should be well-specified enough to eliminate counterexamples (for example, a corridor must be wide enough for both robot and human to pass to be considered \textsc{Frontal Approach}).
    \item \textbf{Laboratory Experiments}: A \textsc{Frontal Approach} definition can be used to set up a laboratory experiment (or regression test) to evaluate the performance of a given policy compared to alternatives - for example \cite{pirk2022protocol}. For the statistical analysis of this experiment to be successful, both metrics and criteria for a successful test need to be defined. For example, in a laboratory experiment, both the robot and the human need to attempt to cross the scenario environment, whereas in a robot deployment or field study, humans may stop to take a phone call, or a robot navigation stack may crash.
    \item \textbf{Dataset Generation}: When creating datasets for social navigation, scenario definitions can be used to curate existing data for inclusion into the dataset, or to guide the setup of robot deployments or laboratory experiments designed to build that data. This scenario categorization can then be used to capture information about the dataset. For example, a pedestrian dataset could collect episodes each with a single scenario like \textsc{Frontal Approach}. In contrast, a crowd dataset, with larger numbers of pedestrians interacting in a larger area, might have episodes with several scenarios happening at once, like \textsc{Frontal Approach}, \textsc{Intersection}, and \textsc{Blind Corner}.
    \item \textbf{Imitation learning}: A \textsc{Frontal Approach} definition can also be used to collect episodes for imitation learning. For this to be successful, additional criteria must be defined - for example, which behaviors are considered successes or failures, or quality metrics which enable rating episodes as better or worse - so a high-quality set of episodes can be collected to enable training of a policy. In other words, while creating imitation learning is dataset generation, not all datasets are good for imitation learning. In the imitation learning use case, \textsc{Frontal Approach} episodes in which the robot or human fail to cross the scenario environment may be marked as failures so they can be excluded by the learning algorithm.

\end{itemize}

In the next section, we outline a methodology for identifying and specifying social navigation scenarios which supports this breadth of usage, present a “Social Navigation Scenario Card” which enables scenarios to be clearly defined and disseminated, list common scenarios in the literature, 
and conclude with guidelines for scenario development and usage.

\begin{table}[t]

\begin{tabular}{ |R{0.3\linewidth} |L{0.6\linewidth} | } 
    \hline
    \multicolumn{2}{| c |}{\textbf{Social Navigation Scenario Card}}\\
    \hline
    \multicolumn{2}{l}{\textbf{Scenario Metadata}} \\
    \hline
    Scenario Name & \textsc{Frontal Approach} \\ 
    \hline
    Scenario Description & A robot and a human approach head-on in a passable space. \\ 
    \hline
    Scientific Purpose & Low-density pedestrian scenario applicable indoors and outdoors. \\ 
    \hline
    \multicolumn{2}{l}{\textbf{Scenario Definition}} \\
    \hline
    Geometric Layout & A space wide enough for the robot and human can pass each other. \\
    \hline
    Intended Robot~Task & The robot navigates from one side of the space to the other \\
    \hline
    Intended Human~Behavior & The human navigates in the opposite direction of the robot. \\
    \hline
    \multicolumn{2}{l}{\textbf{Scenario Usage Guide}} \\
    \hline
    Success Metrics & Success Rate, (No) Collisions \\
    \hline
    Quality Metrics & Comfort, Politeness \\
    \hline
    Ideal Outcome & Robot goes around human in a socially acceptable manner. \\ 
    \hline
    Failure Modes & 1. Robot collides with human \\
    & 2. Robot fails to exit in time limit \\
    \hline
     Labeling Criteria & 1. Robot and human face each other \\ 
    & 2. Robot and human move towards each other at start of episode\\ 
    & 3. Sufficient clearance exists for robot and human to pass each other \\
    \hline
\end{tabular}

\caption{Scenario Card for \textsc{Frontal Approach}}
\label{tab:frontal}

\end{table}

\subsection{Scenario Design Methodology}
\label{s:scenario-methodology}
Interactions occur between humans and robots wherever robots are deployed. Many of their interactions are unique, but others are common enough or important enough to warrant special treatment - whether we are looking for them in data collected from field studies, trying to recreate them in robot deployments and laboratory experiments, or trying to make them happen at scale for dataset generation or imitation learning. Having a clear definition of what behavior we want to identify, recreate or scale can ensure that we have good data and can communicate it to other researchers.

To facilitate this, we propose the use of scenarios defining human-robot interactions, and propose the following methodology for defining scenarios relevant to social navigation. This consists of a three-step process:
\begin{enumerate}[leftmargin=*]
    \item \textbf{Define the Scenario} Scenario definitions should be clearly specified enough to be identified in data or set up as an experiment. Thus, the scenario designer should consider:
        \begin{enumerate}[leftmargin=*]
            \item \textbf{Intended Research Context}: the research topic the scenario is designed to explore, for example, low-density indoor pedestrian navigation or high-density outdoor crowd navigation. Many scenarios general enough to apply to most research contexts.
            \item \textbf{Intended Robot Task}: the high-level objective of the robot, for example, navigation between two points, visual navigation, or guiding a person to a goal location.
            \item \textbf{Intended Human Behavior}: the high-level objectives of nearby people, for example, navigating between two points, delivering a package, or following the robot to a goal location.
            \item \textbf{Success Metrics}: the criteria which define the successful completion of the robot's task. While scenarios may play out in the wild in a variety of ways, the robot's task should be well-specified enough that it is unambiguous whether it succeeded.
        \end{enumerate}
    \item \textbf{Evaluate the Definition}: The way scenarios are designed affects the aspects of robot behavior that they evaluate and what behaviors they elicit in humans, sometimes in unexpected ways Therefore, we propose that designers should evaluate scenarios after their initial design, assessing their ability to measure the desired robot behaviors. Well-designed scenarios should have the properties of commonality, flexibility, and fitness to purpose.
    \begin{enumerate}[leftmargin=*]
        \item \textbf{Commonality}: Well-designed scenarios should be designed to evaluate the designer's intended criteria while maintaining identifiable characteristics that allow it to be grouped and compared with similar scenarios in use in the community. Common categories of scenarios are listed as sections of Table~\ref{tab:scenarios}, and include ``approach'' or ``hallway'' scenarios involving robots approaching people or objects from specific directions, ``intersection'' scenarios where robots and humans cross paths, and ``interpersonal'' scenarios such as robots leaving or joining conversational groups. Scenario designers should compare their scenarios with these common scenarios to avoid introducing redundant scenarios when existing scenarios are available.
        \item \textbf{Flexibility}: Well-designed scenarios should be broadly specified enough to capture the full range of behavior which occurs in the wild. It is important to avoid ``solutionizing'' in which scenarios prescribe intended robot or human behavior so narrowly that naturally occurring variants are included. Instead, scenarios should have broad, flexible definitions that enable them to capture behaviors that happen, along with clear success metrics to evaluate whether that behavior came out as intended.
        \item \textbf{Fitness to Purpose}: Well-designed scenarios should allow the scenario designer to evaluate the adaptations of robot behaviors with which they are concerned. For example, researchers have explored how proactive robot behaviors can improve social interactions during navigation~\cite{khambhaita2020viewingRachid}. To evaluate robots that exhibit proactive cooperation, the scenario must be flexible enough to allow proactive cooperativeness interactions to occur. Early drafts of scenarios should be piloted to confirm that desired behaviors can be detected and elicited and that success metrics measure what is intended.
    \end{enumerate}
    \item \textbf{Communicate the Definition}: Once a scenario has been evaluated, it should be communicated clearly and consistently. A scenario definition should be specific enough to replicate, so other researchers can identify occurrences of the scenario in their data, recreate it in the laboratory, and determine whether instances of a scenario correspond to the intended outcome for the human or robot.
\end{enumerate}

Social navigation scenario development can be seen as a step towards the more formal scenarios engineering approach being adopted in intelligent vehicles research~\cite{10109502, li2022features, li2022novel}. 
To facilitate communicating scenarios, we propose a Social Navigation Scenario Card, presented next.

\begin{table*}[t]
    \centering\scriptsize
    \begin{tabular}{|R{0.09\linewidth} | L{0.13\linewidth}| L{0.05\linewidth}| L{0.05\linewidth}| L{0.06\linewidth}|L{0.05\linewidth}| L{0.05\linewidth}|L{0.06\linewidth}|L{0.08\linewidth}|L{0.07\linewidth}|L{0.045\linewidth}|}
    \hline

     \textbf{Scenario Name} & \textbf{Scenario Description} & \textbf{Physical Env.} & \textbf{Geom. Layout} & \textbf{Scientific Purpose} & \textbf{Robot Role} & \textbf{Robot Task} & \textbf{Human Behavior} & \textbf{Ideal Outcome} & \textbf{Related Scenarios} & \textbf{Cited In} \\
    \hline
    \multicolumn{11}{l}{\textbf{Hallway Scenarios}} \\
    \hline

    \textsc{Frontal Approach} & A pedestrian and robot approach head-on. & Generic & Passable Space & Pedestrian Interaction & Any & Navigate A to B & Navigate B to A & Robot / humans pass & \textsc{Ped. Obstruct} & \cite{pirk2022protocol, gao2021evaluation, wang2022metrics} \\

    \hline
    \textsc{Pedestrian Overtaking} & A pedestrian overtakes a moving robot. & Generic & Passable Space & Pedestrian Interaction & Any & Navigate A to B & Navigate A to B & Human passes robot & \textsc{Down Path} &  \cite{chen2017sociallyawareDRLmit_iros17}\\
    \hline
    \textsc{Robot Overtaking} & A robot overtakes a moving pedestrian. & Generic & Passable Space & Pedestrian Interaction & Any & Navigate A to B & Navigate A to B & Robot passes human &  & \cite{gao2021evaluation, wang2022metrics} \\

    \hline
    \textsc{Intersection (No~Gesture)} & A robot and a human cross at an intersection. & Indoor & Intersection & Pedestrian Interaction & Any & Navigate A to B & Cross Navigate & Both pass no collision &  & \cite{tsoi2022sean, gao2021evaluation, wang2022metrics, chen2017decentralized} \\
    \hline
    \textsc{Intersection (Gest.~Wait)} & A robot is told to wait at an intersection. & Indoor & Intersection & Pedestrian Interaction & Servant & Navigate A to B & Cross Navigate & Human goes then robot & \textsc{Gesture Proceed} & \cite{pirk2022protocol} \\
    \hline

    \textsc{Blind Corner} & A robot and human meet at a blind corner. & Indoor & Corner & Pedestrian Interaction & Any & Navigate A to B & Navigate B to A & No collision / obstruction &  & \cite{pirk2022protocol, xiao2022learning} \\

    \hline
    \multicolumn{11}{l}{\textbf{Doorway Scenarios}} \\
    \hline

    \textsc{Narrow Doorway} & A robot and human pass a narrow doorway. & Indoor & Room \& Door & Pedestrian Interaction & Any & Navigate A to B & Navigate B to A & No collision / obstruction & \textsc{Narrow Arch} & \cite{pirk2022protocol} \\
    \hline
    \textsc{Entering Room} & A robot enters a room occupied by a human & Indoor & Room \& Door & Pedestrian Interaction & Any & Navigate out to in & Navigate in to out & Robot allows human exit & \textsc{Entering Elevator} & R@G \\
    \hline
    \textsc{Exiting Room} & A robot exits a room while a person enters. & Indoor & Room \& Door & Pedestrian Interaction & Any & Navigate in to out & Navigate out to in & Robot exits first & \textsc{Exiting Elevator} &  R@G \\

    \hline
    \multicolumn{11}{l}{\textbf{Interpersonal Scenarios}} \\
    \hline

    \textsc{Joining a Group} & A robot joins a group of robots or people. & Generic & Open Space & Group Interaction & Any & Navigate to group & Continue conversing & Robot joins group & \textsc{Leaving a Group} & \cite{tsoi2022sean, gao2021evaluation} \\
    \hline
    \textsc{Following} & A robot follows a person. & Generic & Walking Space & Joint Navigation & Servant & Follow human & Lead robot & Robot follows person & \textsc{Accompany Peer} & \cite{gao2021evaluation} \\
    \hline
    \textsc{Leading} & A robot leads a person. & Generic & Walking Space & Joint Navigation & Leader & Lead human & Follow robot & Robot leads person & \textsc{Tour Guide} & \cite{gao2021evaluation} \\

    \hline
    \multicolumn{11}{l}{\textbf{Crowd Scenarios}} \\
    \hline
    \textsc{Crowd Navigation} & A robot navigates through a crowd. & Generic & Passable Space & Crowd Navigation & Any & Navigate thru & Mill about & No collision / obstruction & \textsc{Robot Crowding} &  Various \\
    \hline
    \textsc{Parallel Traffic} & Crowd moves parallel to the robot. & Generic & Passable Space & Crowd Navigation & Any & Navigate A to B& Mill from A to B & No collision / obstruction & \textsc{Circular Crossing} & \cite{wang2022metrics} \\
    \hline
    \textsc{Perpendicular Traffic} & Crowd moves perpendicular to robot. & Generic & Intersection & Crowd Navigation & Any & Cross Navigate & Mill from A to B & No collision / obstruction & \textsc{Plaza Crossing} & \cite{wang2022metrics} \\

    \hline
    \multicolumn{11}{l}{\textbf{Specialized Scenarios}} \\
    \hline

    \textsc{Object Handover} & A robot hands an object to a human. & Generic & Passable Space & Interactive Navigation & Servant & Deliver object & Receive object & Human takes object & \textsc{Robot Courier} & \cite{tsoi2022sean} \\
    \hline

    \textsc{Crash Cart} & Robot delivering a medical product. & Indoor & Passable Space & Interactive Navigation & Leader & Deliver object & Receive object & Delivery of medicine & \textsc{Food Delivery} & This paper \\

    \hline
    \end{tabular}
    \caption{Example social navigation scenarios. For illustrations of the geometric layout, see Figure~\ref{fig:scenarios}. Closely related scenarios are listed in the second-to-last column. Citations refer to either papers or challenges defining the scenario, or R@G for scenarios from an unpublished Robotics at Google\cite{robotics-at-google} deployment, developed according to the protocol in \cite{pirk2022protocol} .}
    \label{tab:scenarios}
    
\end{table*}

\subsection{Social Navigation Scenario Cards}
Ideally, a social navigation scenario consists of a well-defined social interaction including robots performing tasks, people performing behaviors, and relevant features of their environment. 
This definition should be specific enough that an encounter can be labeled as an instantiation of the scenario, but loose enough that it captures a wide variety of behaviors.  For ease of reusability, scenarios should ideally be realistic in that they represent real-world scenarios, scalable in that they can be set up at low cost, and repeatable in that the same scenario could be conducted many times under similar conditions. However, scenarios may encompass a wide variety of situations, from a simple \textsc{Frontal Approach} of a robot and human passing each other up to the complexity of a robot navigating a crowd exiting a stadium, and Scenario Cards should remain flexible enough to capture these use cases.

Following work on ``model cards'' in the machine learning community~\cite{mitchell2019model}, we propose a ``Scenario Card'' approach to defining scenarios which 
which labels the scenario with a set of features that unambiguously define it. Scenario cards have the following three major elements: (a) Scenario Metadata that define the name, description, and scientific purpose of the scenario; (b) Scenario Definition which clearly describes the environment, intended human behavior, and intended robot task; and (c) a Scenario Usage Guide, which provides additional information for specialized usages such as evaluation metrics, success and failure criteria.

\subsubsection{Scenario Metadata}
The Scenario Metadata identifies a scenario in an unambiguous way for other researchers, including the type of the scenario (doorway, hallway, etc.), its name, its description, and its scientific purpose (crowd navigation, low-density pedestrian, interactive, etc.). For example, a head-on pedestrian approach scenario might be labeled \textsc{Frontal Approach}, which we will use as a running example.
\begin{itemize}
    \item \textbf{Scenario Type:} Scenarios can be grouped into broad classes such as head-on approaches versus intersections, doorways and elevators, crowd versus group, interactive and accompanying, and so on. Identifying the group a scenario belongs to can help researchers decide whether to include it for coverage or exclude it as redundant.

    \item \textbf{Name:} The scenario should be given a unique name which does not conflict with existing scenarios used within the community.

    \item \textbf{Description:} The scenario should have a brief description that communicates what is intended to happen in it.

    \item \textbf{Research Context:} Scenarios often are targeted at specific scientific purposes along various dimensions of research interest - for example, indoor low-density pedestrian scenarios or outdoor high-density crowd scenarios. Key elements which are often distinguished include:
    \begin{itemize}
        \item \textbf{Location: Indoor, Outdoor or General} Indoor and outdoor navigation have different constraints and are often studied separately; however, some scenarios, like \textsc{Frontal Approach}, can occur in many contexts.
        \item \textbf{Density: Pedestrian or Crowd} Low-density pedestrian studies (where robots encounter only a few individuals at a time) are often studied separately than high-density crowd scenarios (in which people exhibit qualitatively different behavior).
        \item \textbf{High-Level Task: Navigation, Delivery or Interaction} Many scenarios focus on pure navigation tasks, but others involve object delivery, interacting with humans, leaving and joining groups, and so on.
    \end{itemize}
\end{itemize}

\subsubsection{Scenario Definition}
The Scenario Definition defines roughly what is meant by the scenario, in a precise but broad way that allows scenarios to be identified but not so restrictive as to prevent recording important behaviors. For example, \textsc{Frontal Approach} sceanario definition should enable us to recognize that a robot and human are approaching head-on, but at the same time capture an interaction where the human changes their direction or stops to answer their phone.
\begin{itemize}
    \item \textbf{Geometric Layout:} Scenarios often occur in specific physical environments, such as corridors, doorways, at blind corners, or near elevators. The important features of the environment should be noted; features that can vary should also be noted so the scenario is not overspecified.
    
    \item \textbf{Intended Robot Task:} The number of robots and their desired behaviors should be recorded. A robot simply navigating around a pedestrian has different behaviors than one which is specifically attempting to navigate to a given target. Typical robot tasks are a robot heading to a pre-defined position, a robot guiding a person to a destination, or a robot delivering an item.        
    \item \textbf{Intended Human Behavior:} The expected human behavior should be specified. In the scenario definition, behaviors should be specified clearly enough to recognize the behavior in data or to enable a human to attempt to perform it, but not too specific that diverse behaviors could not be collected.
\end{itemize}

\subsubsection{Scenario Usage Guide}
The Scenario Usage Guide specifies how the scenario is used in practice, and contains additional information that goes beyond the definition, such as idealized outcomes or instructions for human confederates for experimental setups. This is the place where a \textsc{Frontal Approach} scenario would express that the ideal outcome is that the robot and human pass each other without incident and exit on the opposite sides of the scenario area.

\begin{itemize}
    \item \textbf{Labeling Criteria:} A clear set of criteria should be provided so that scenarios can be labeled in logs data or rejected in the event of a structured run. For example, for an intersection scenario, one could demand that the robot passes within two meters of the human and that their intended paths at least potentially cross.
    \item \textbf{Success and Quality Measures:} To evaluate how well the robot performed in the scenario, we may also want to specify ``Success Measures" and ``Quality Measures" specific to a scenario such as the ability of the robot to ensure legibility of its behavior, to limit and control disturbance, to facilitate human action and situation understanding, etc.
    \item \textbf{Ideal Outcome and Failure Modes:} To enable researchers to evaluate robot performances in episodes for imitation learning or data analysis, the ideal outcome should be outlined, for example, that a robot should not collide with a human at a blind corner. Also, to help debug scenarios and guard the safety of human participants, failure modes such as colliding with walls, or stopping dead after a near-collision, should be outlined. We include failure modes in ideal outcomes in the scenario usage guide and not the definition because researchers interested in data collection do not want to artificially exclude arbitrary outcomes that can occur in the wild; however, this is critical information for imitation learning researchers trying to craft behavior. 
    \item \textbf{Human Behavior Playbook:} If a scenario is designed to be created in a repeatable way as part of an experiment, a specific script or rubric should be provided so that the participants can perform their roles appropriately. For example, intended human behavior might be travelers in a crowded railway station, or workers going alone or with colleagues in an office context. These could include variations in the behaviors: for instance, some travelers might be in a hurry while others have more time. Also, there could be several categories of users in a given context that might act and react differently. 
    \item \textbf{Contextual Information:} Principle P7 notes that a robot's behavior should depend on context: tor instance, a robot should behave differently if the place is very calm and needs silence or if it is a busy place, so ideally robots should recognize in which contextual situation a scenario is happening. Success metrics, ideal outcomes, failure modes, human behavior and more can be altered by the context, so it can be useful to outline any important contextual variants of the scenario and how they affect intended robot or human behavior.
\end{itemize}

\begin{figure*}[tb!p]
    \centering
    \begin{subfigure}[b]{0.19\textwidth}
     \centering
     \includegraphics[width=\textwidth]{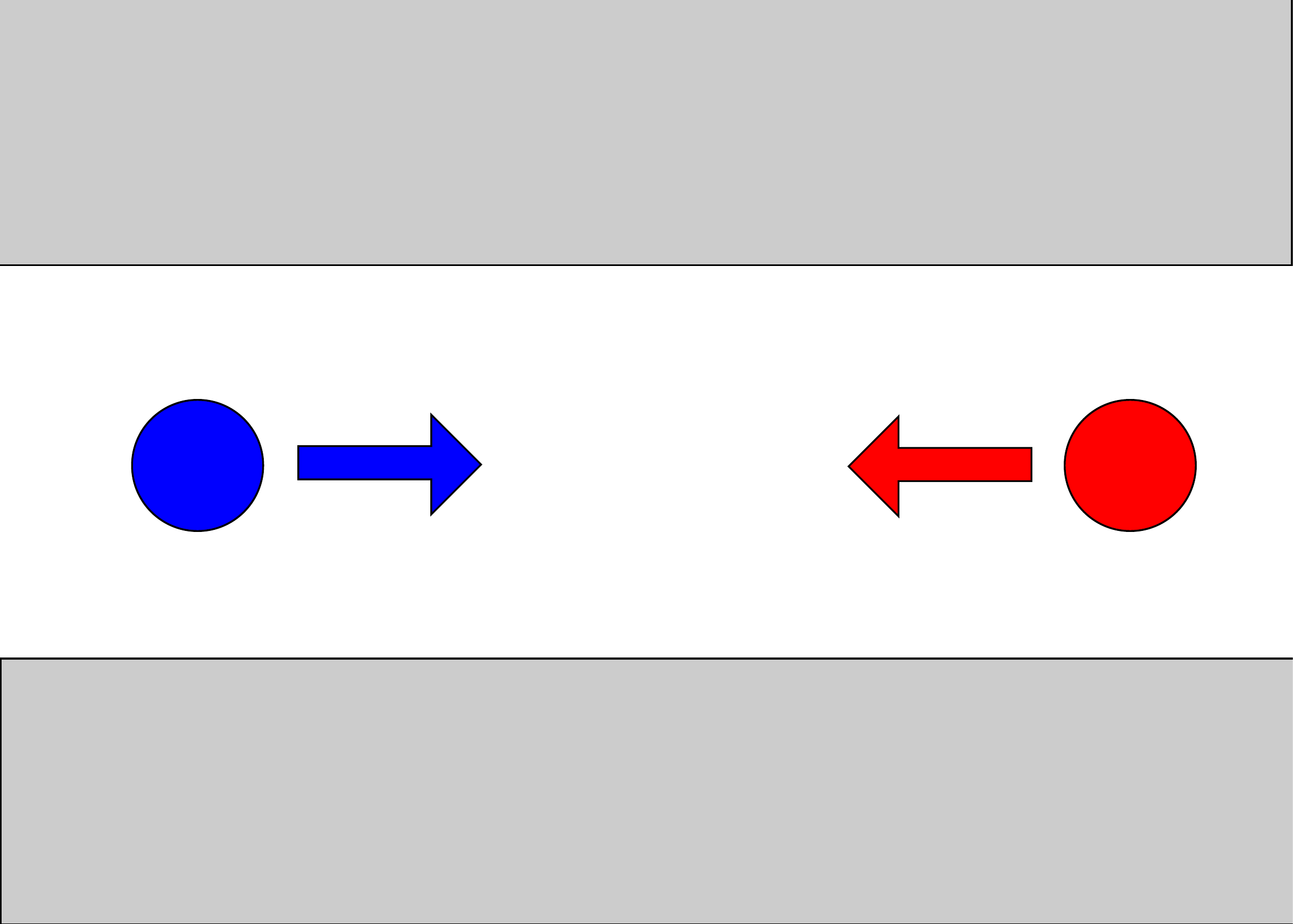}
     {\small \caption{Frontal Approach}}
    \end{subfigure}
    \hfill
    \begin{subfigure}[b]{0.19\textwidth}
     \centering
     \includegraphics[width=\textwidth]{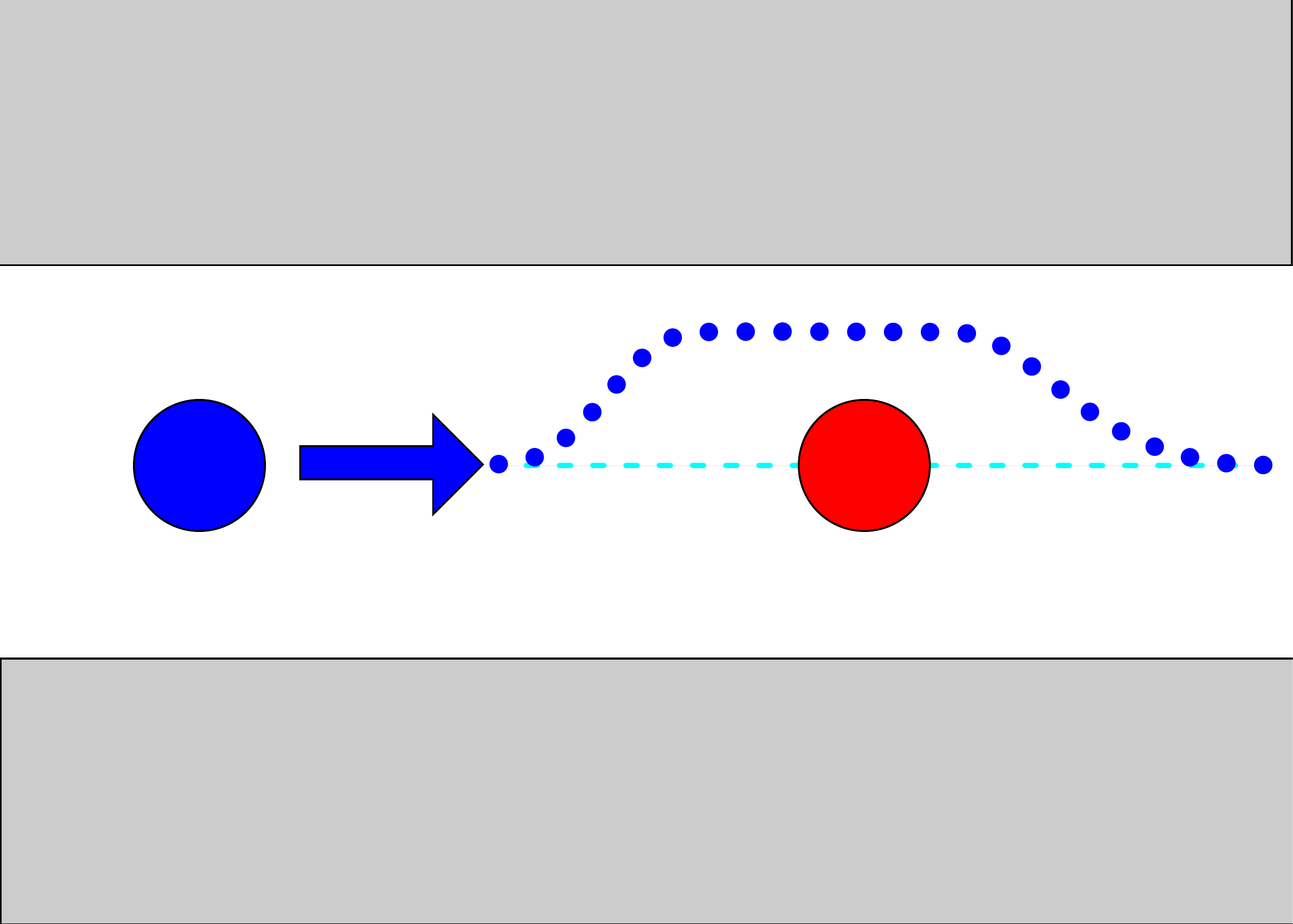}
     {\small \caption{Pedestrian Obstruction}}
    \end{subfigure}
    \hfill
    \begin{subfigure}[b]{0.19\textwidth}
     \centering
     \includegraphics[width=\textwidth]{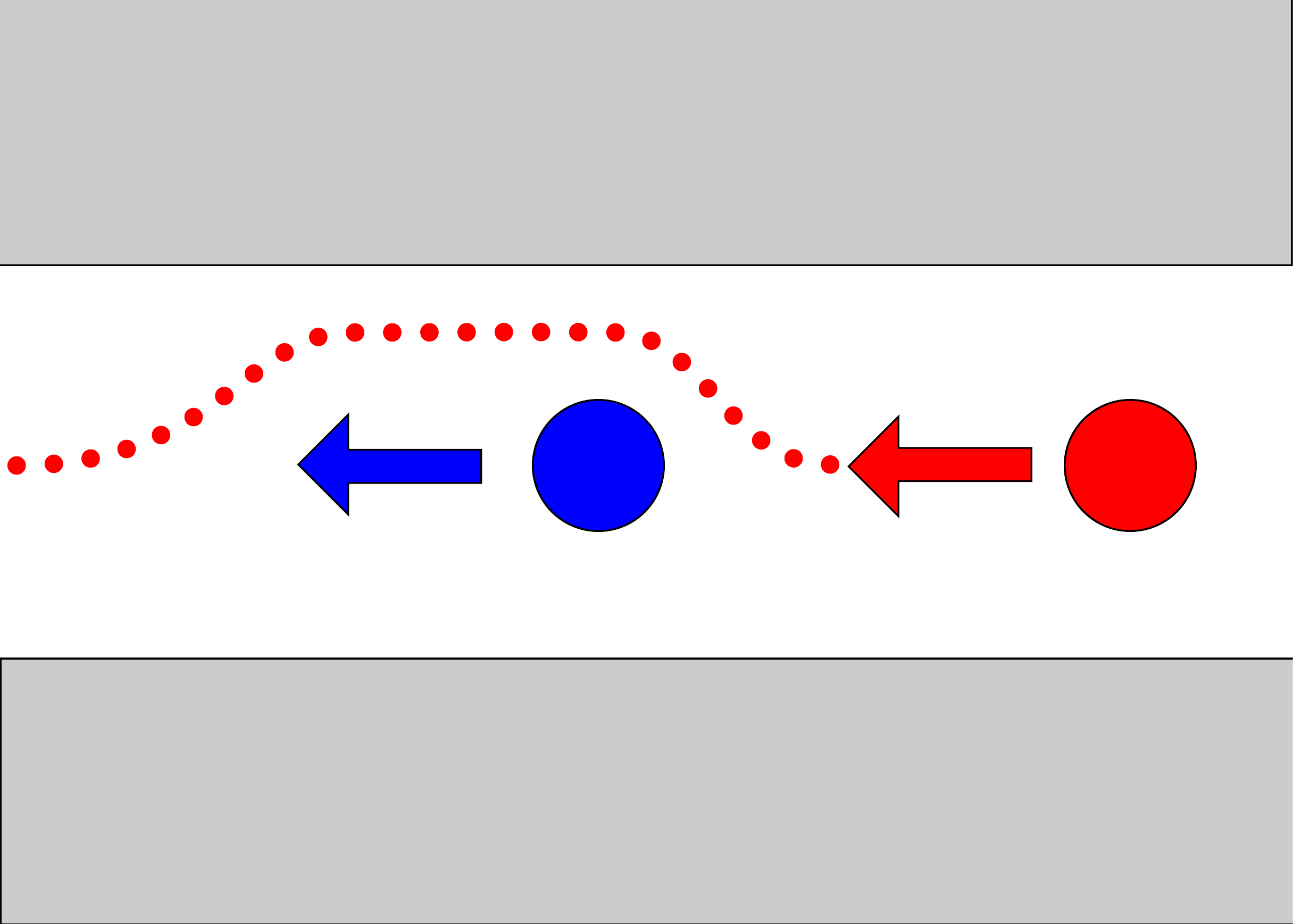}
     {\small \caption{Pedestrian Overtaking}}
    \end{subfigure}
    \hfill
    \begin{subfigure}[b]{0.19\textwidth}
     \centering
     \includegraphics[width=\textwidth]{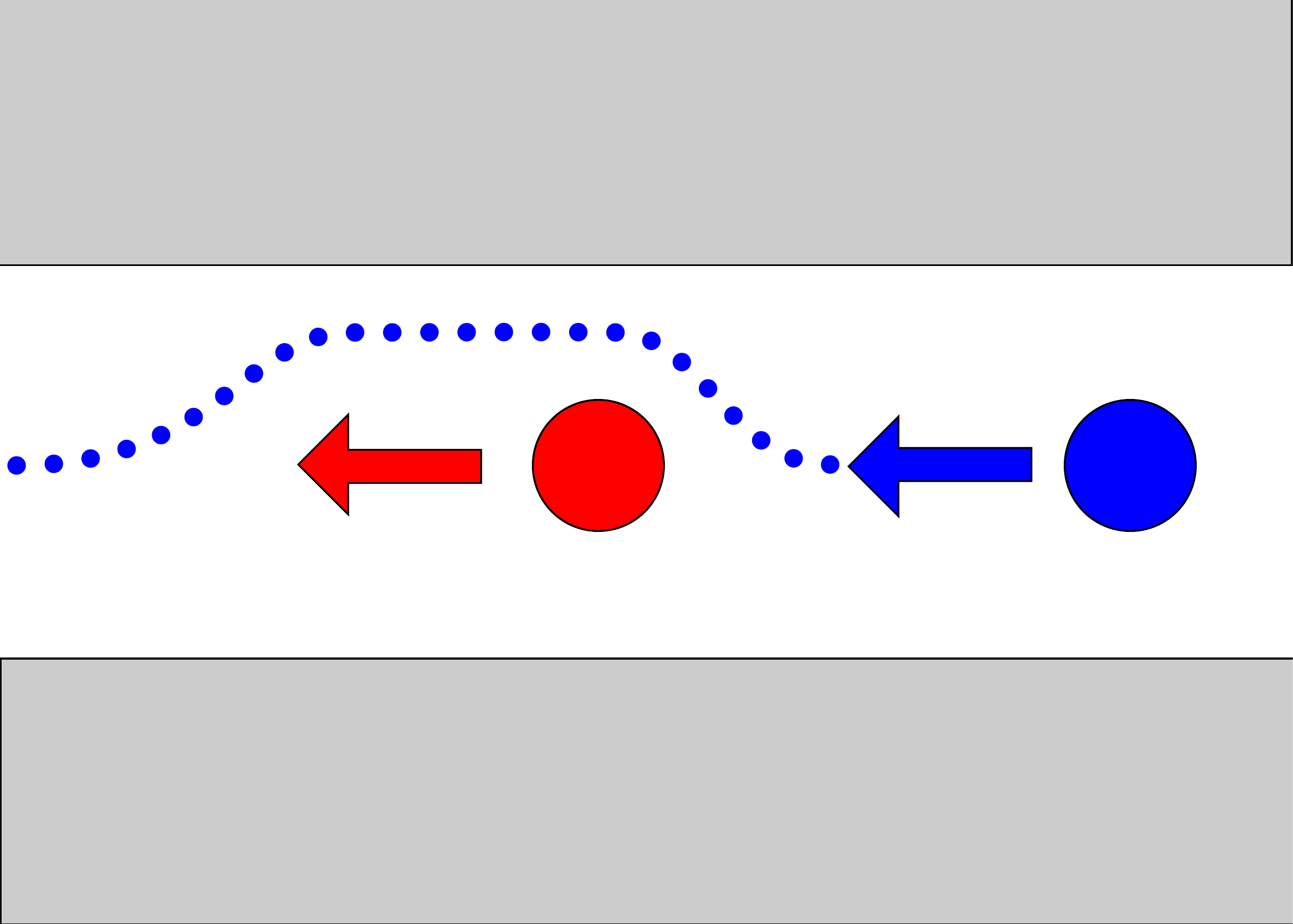}
     {\small \caption{Robot Overtaking}}
    \end{subfigure}
    \hfill
    \begin{subfigure}[b]{0.19\textwidth}
     \centering
     \includegraphics[width=\textwidth]{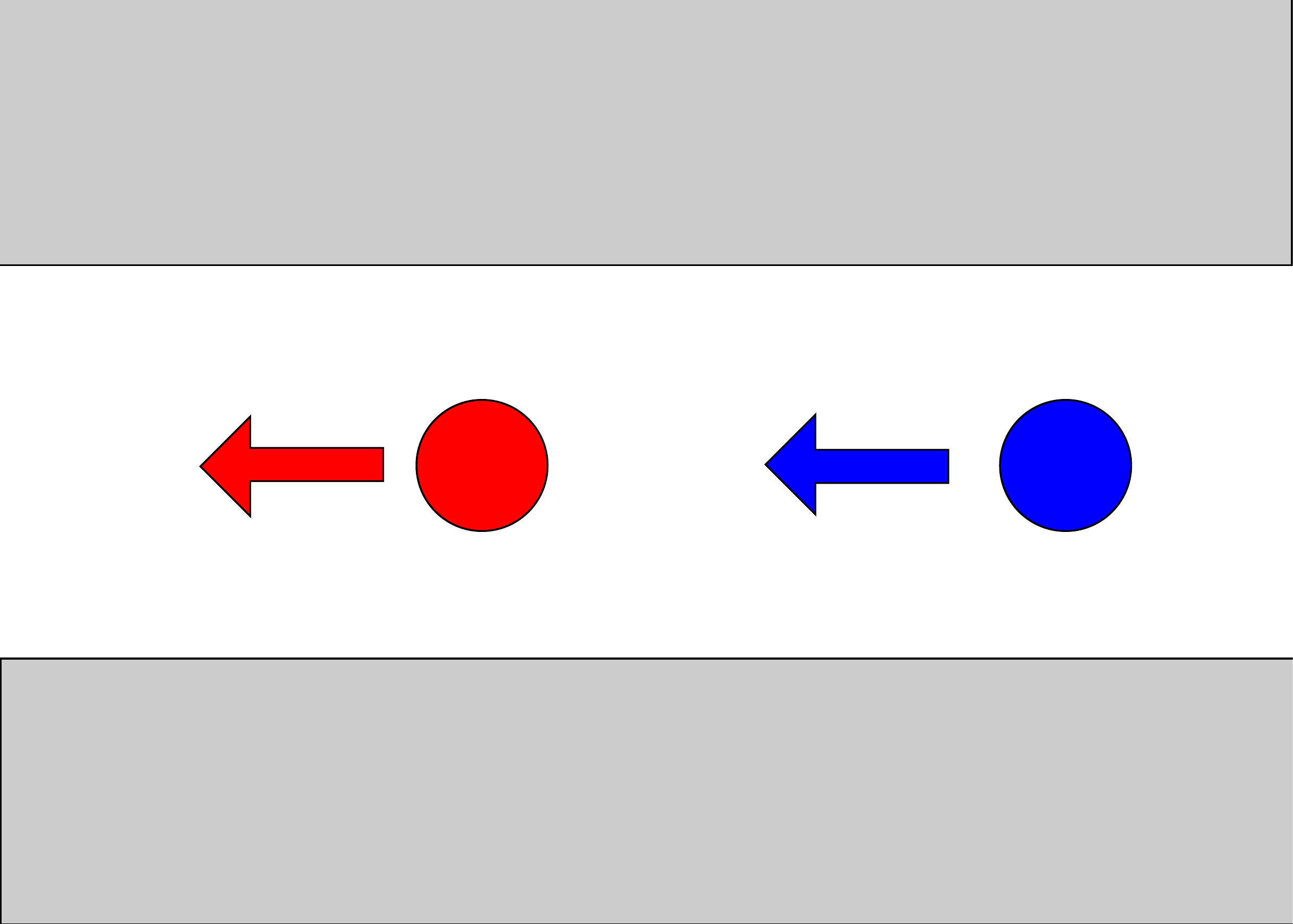}
     {\small \caption{Down Path}}
    \end{subfigure}
    \bigskip

    \begin{subfigure}[b]{0.19\textwidth}
     \centering
     \includegraphics[width=\textwidth]{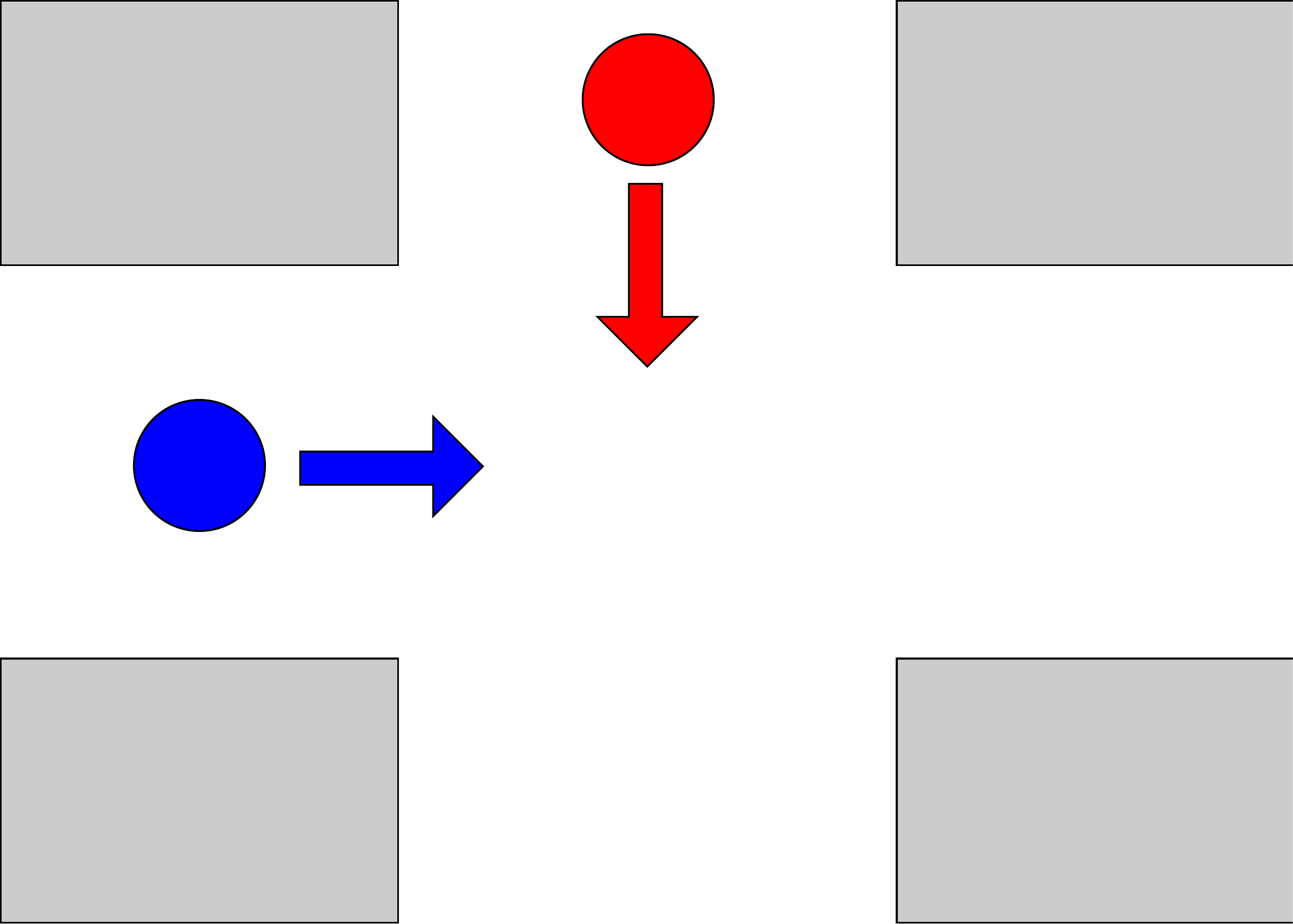}
     {\small \caption{Intersection No Gesture}}
    \end{subfigure}
    \hfill
    \begin{subfigure}[b]{0.19\textwidth}
     \centering
     \includegraphics[width=\textwidth]{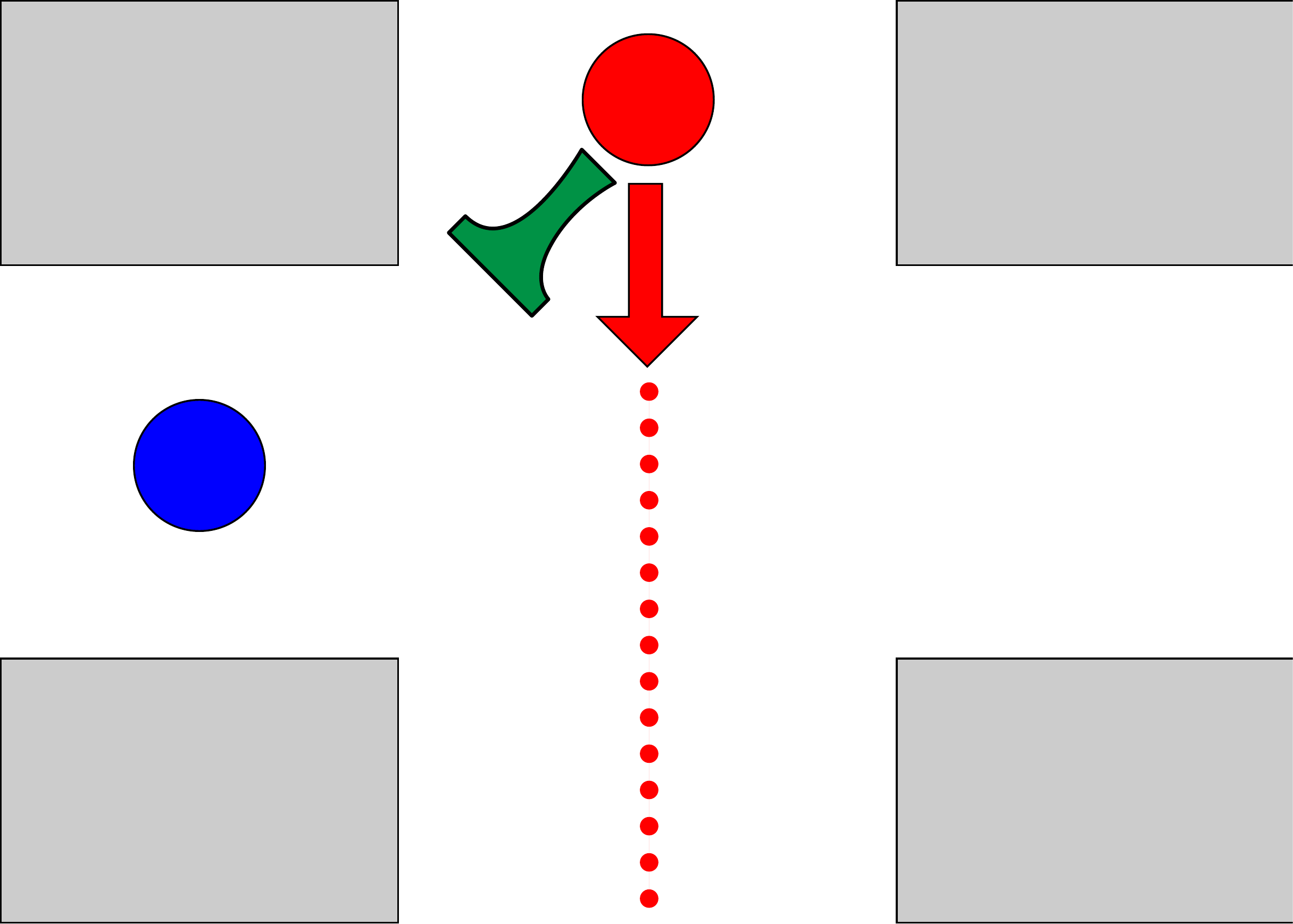}
     {\small \caption{Intersection Wait}}
    \end{subfigure}
    \hfill
    \begin{subfigure}[b]{0.19\textwidth}
     \centering
     \includegraphics[width=\textwidth]{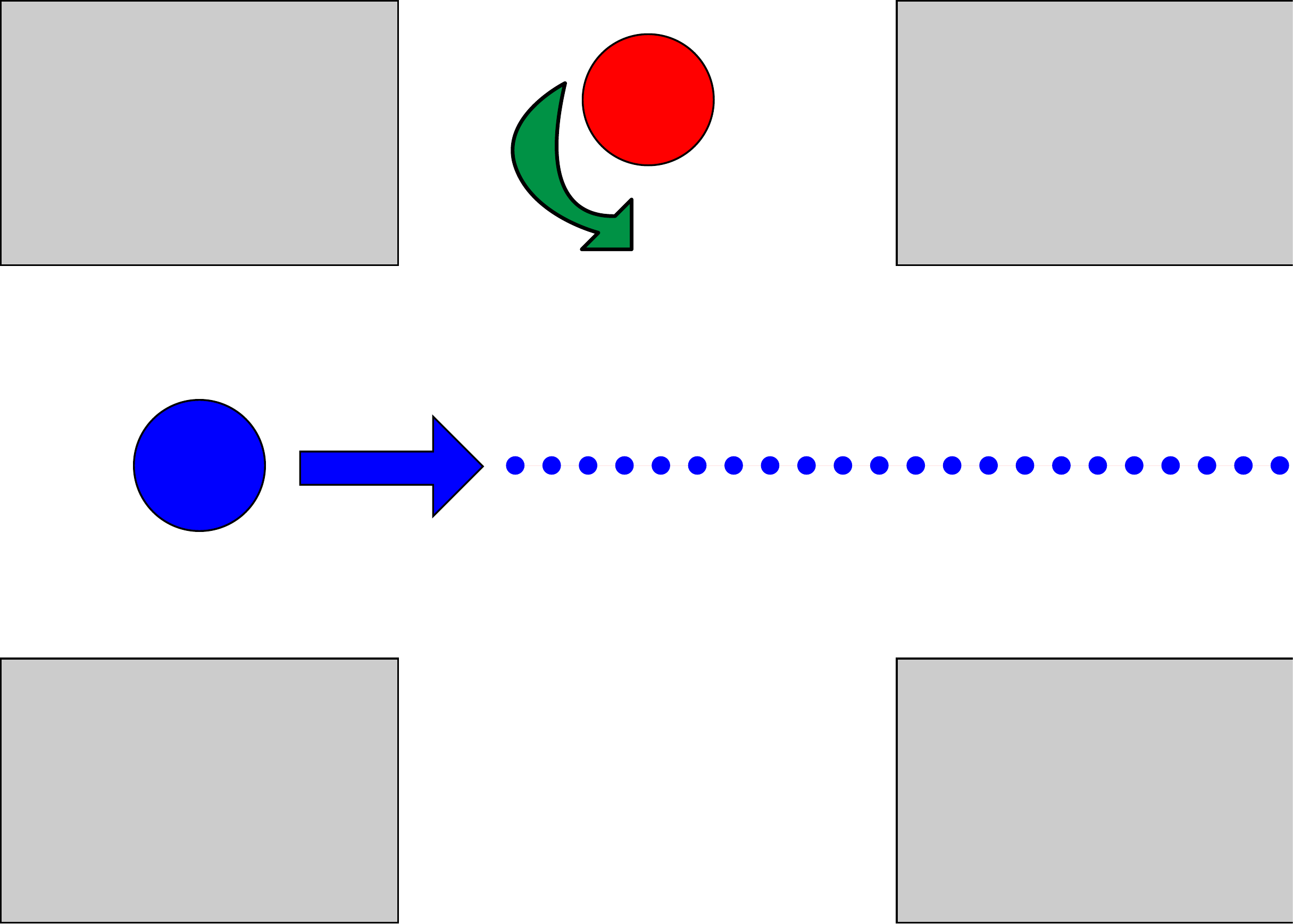}
     {\small \caption{Intersection Proceed}}
    \end{subfigure}
    \hfill
    \begin{subfigure}[b]{0.19\textwidth}
     \centering
     \includegraphics[width=\textwidth]{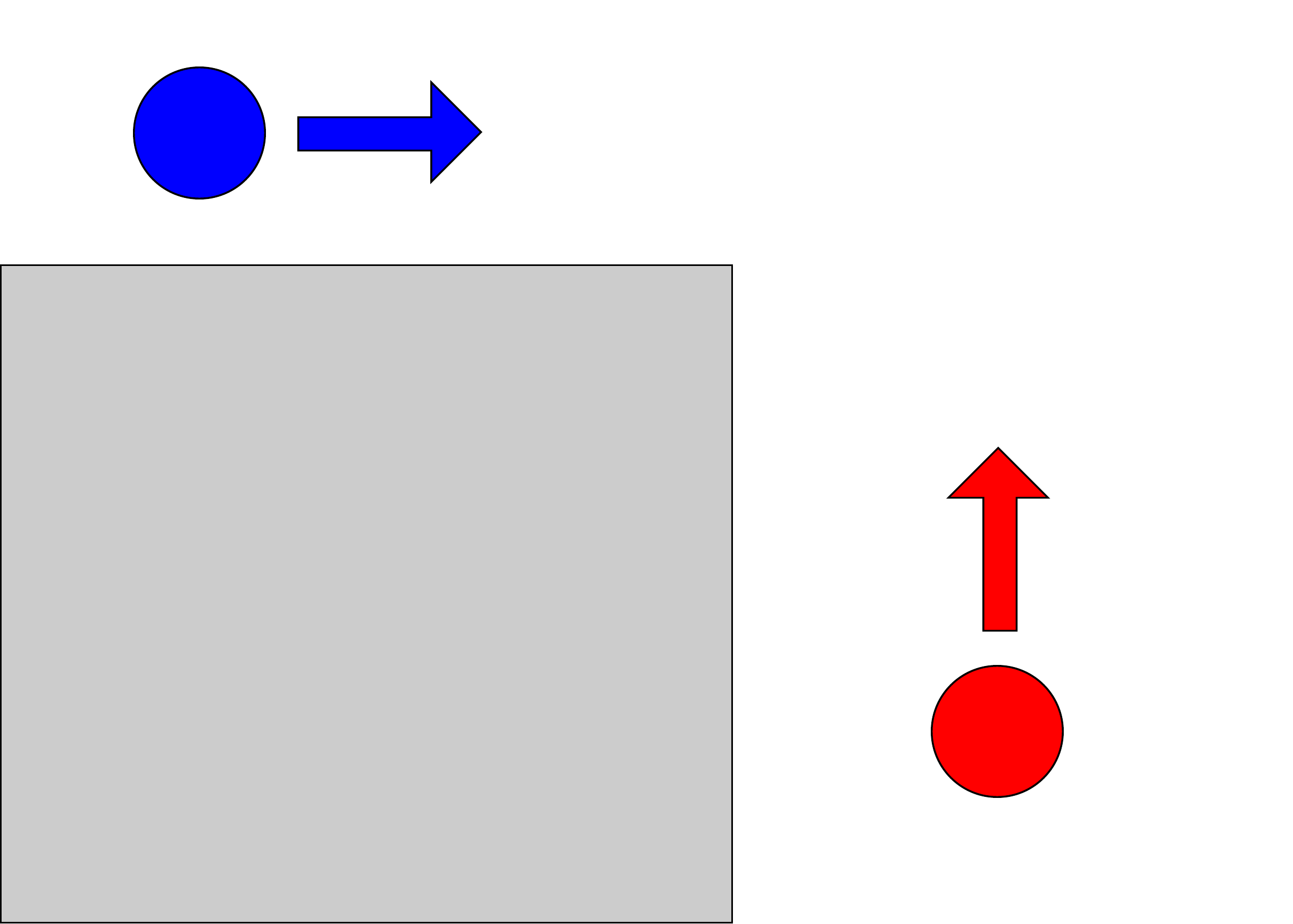}
     {\small \caption{Blind Corner}}
    \end{subfigure}
    \hfill
    \begin{subfigure}[b]{0.19\textwidth}
     \centering
     \includegraphics[width=\textwidth]{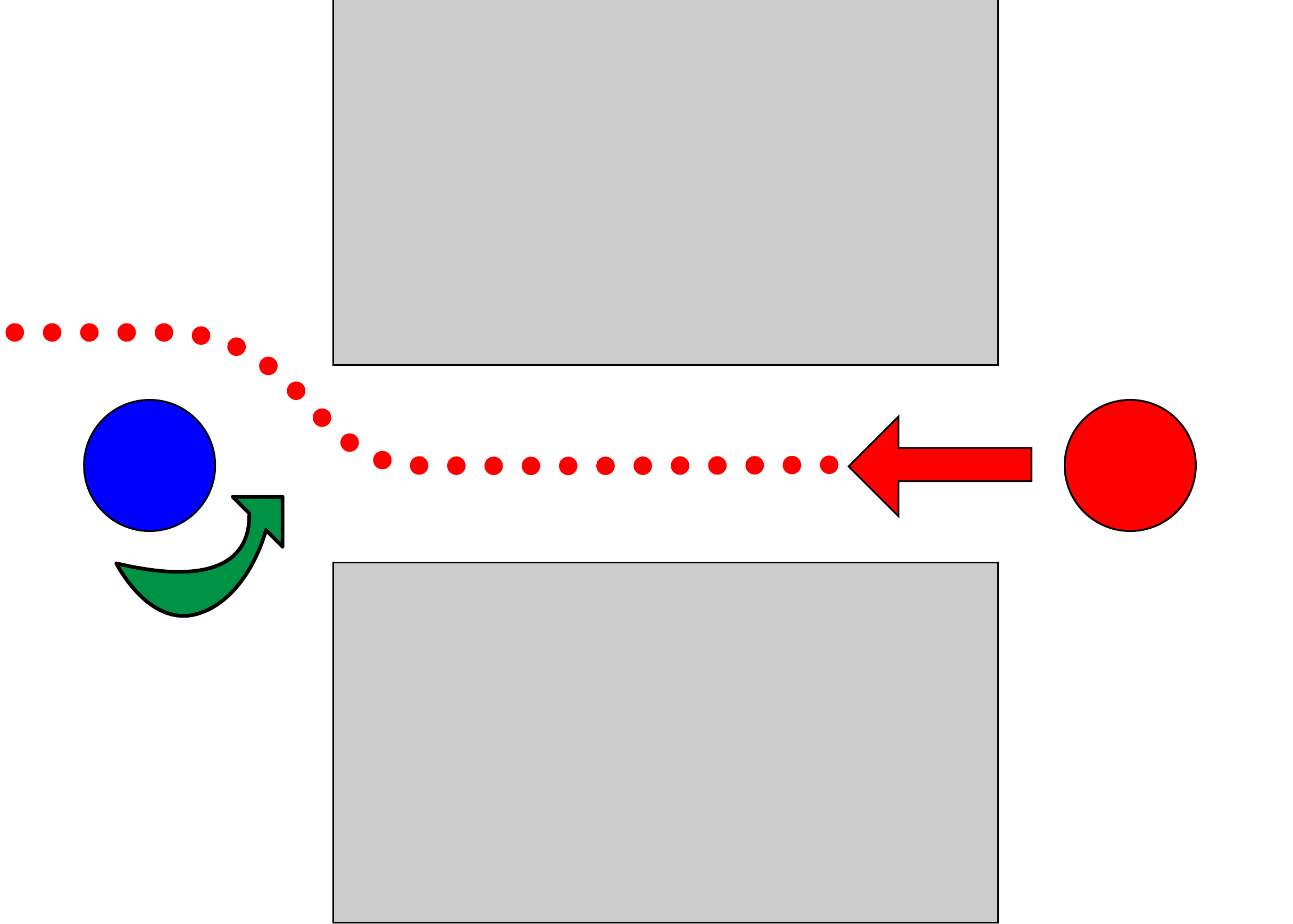}
     {\small \caption{Narrow Hallway}}
    \end{subfigure}
    \bigskip

    \begin{subfigure}[b]{0.19\textwidth}
     \centering
     \includegraphics[width=\textwidth]{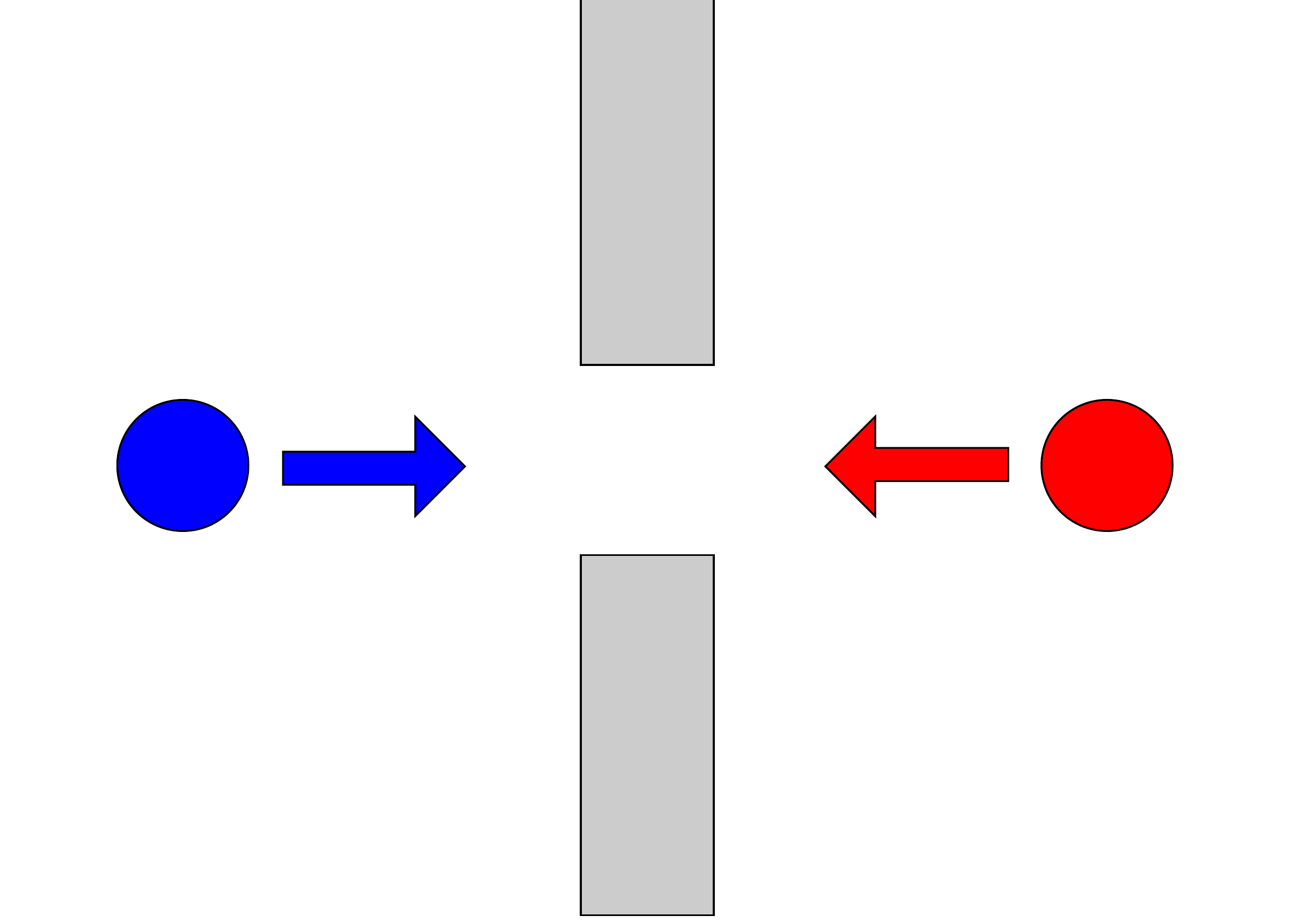}
     {\small \caption{Narrow Doorway}}
    \end{subfigure}
    \hfill
    \begin{subfigure}[b]{0.19\textwidth}
     \centering
     \includegraphics[width=\textwidth]{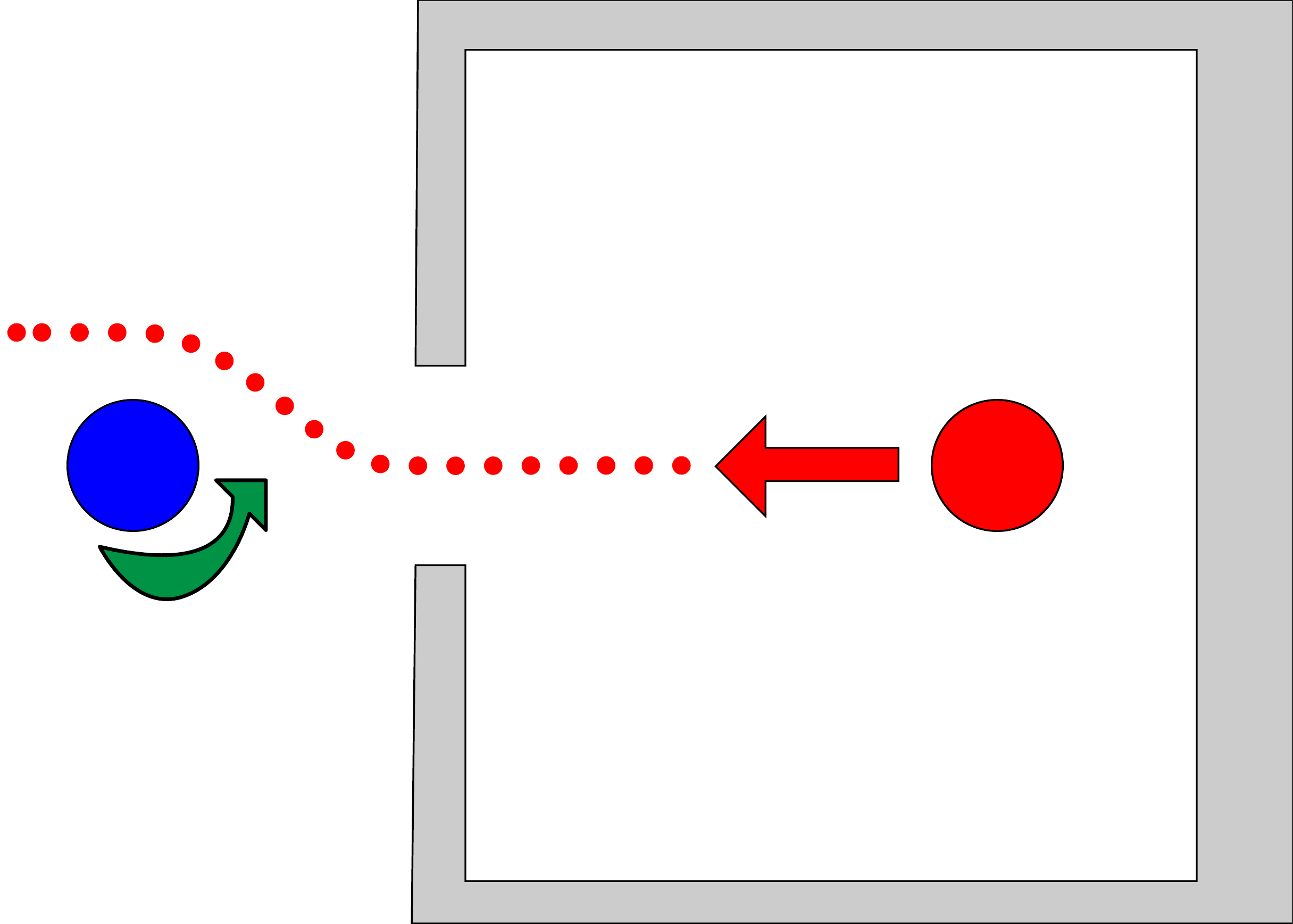}
     {\small \caption{Entering Room}}
    \end{subfigure}
    \hfill
    \begin{subfigure}[b]{0.19\textwidth}
     \centering
     \includegraphics[width=\textwidth]{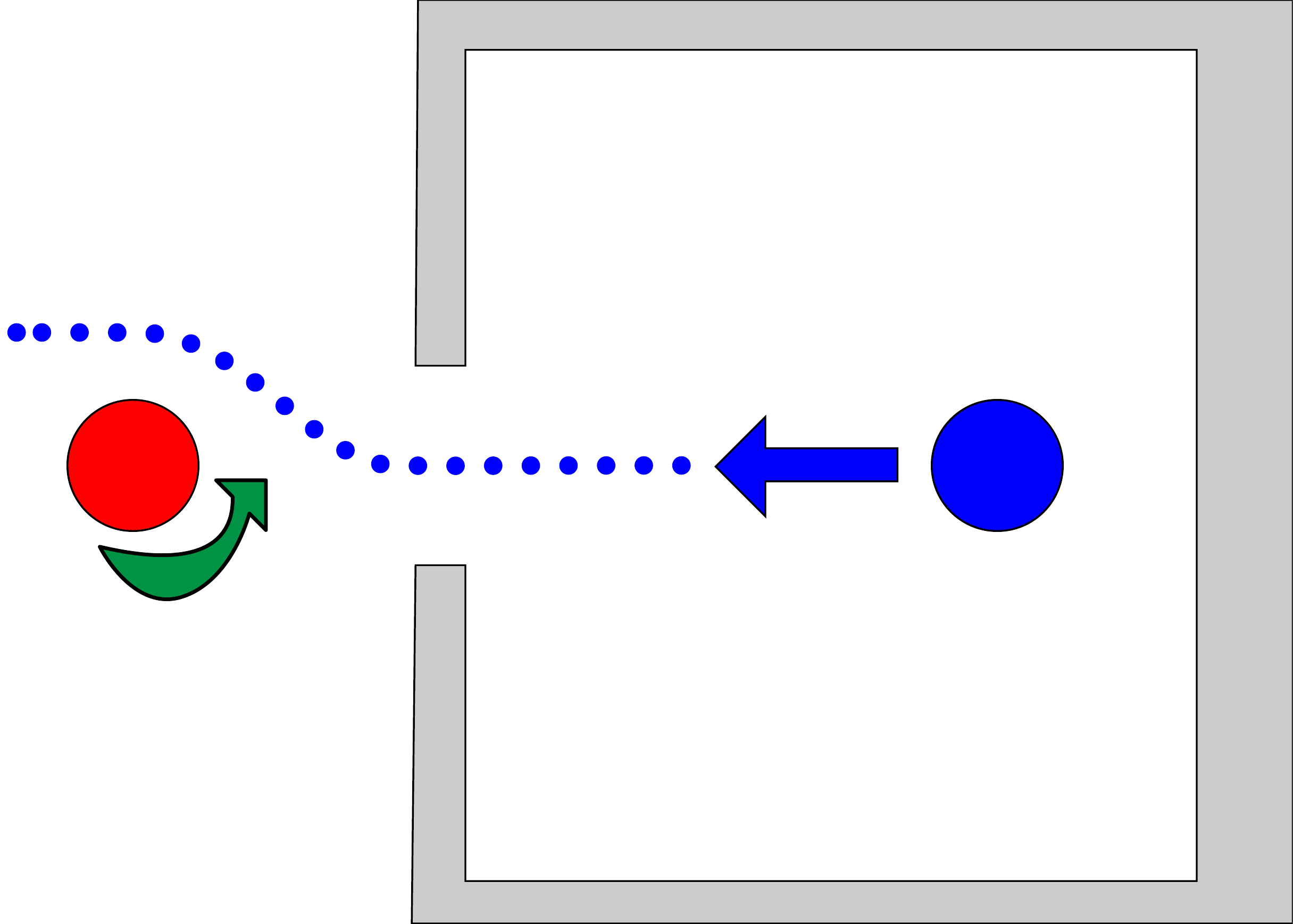}
     {\small \caption{Exiting Room}}
    \end{subfigure}
    \hfill
    \begin{subfigure}[b]{0.19\textwidth}
     \centering
     \includegraphics[width=\textwidth]{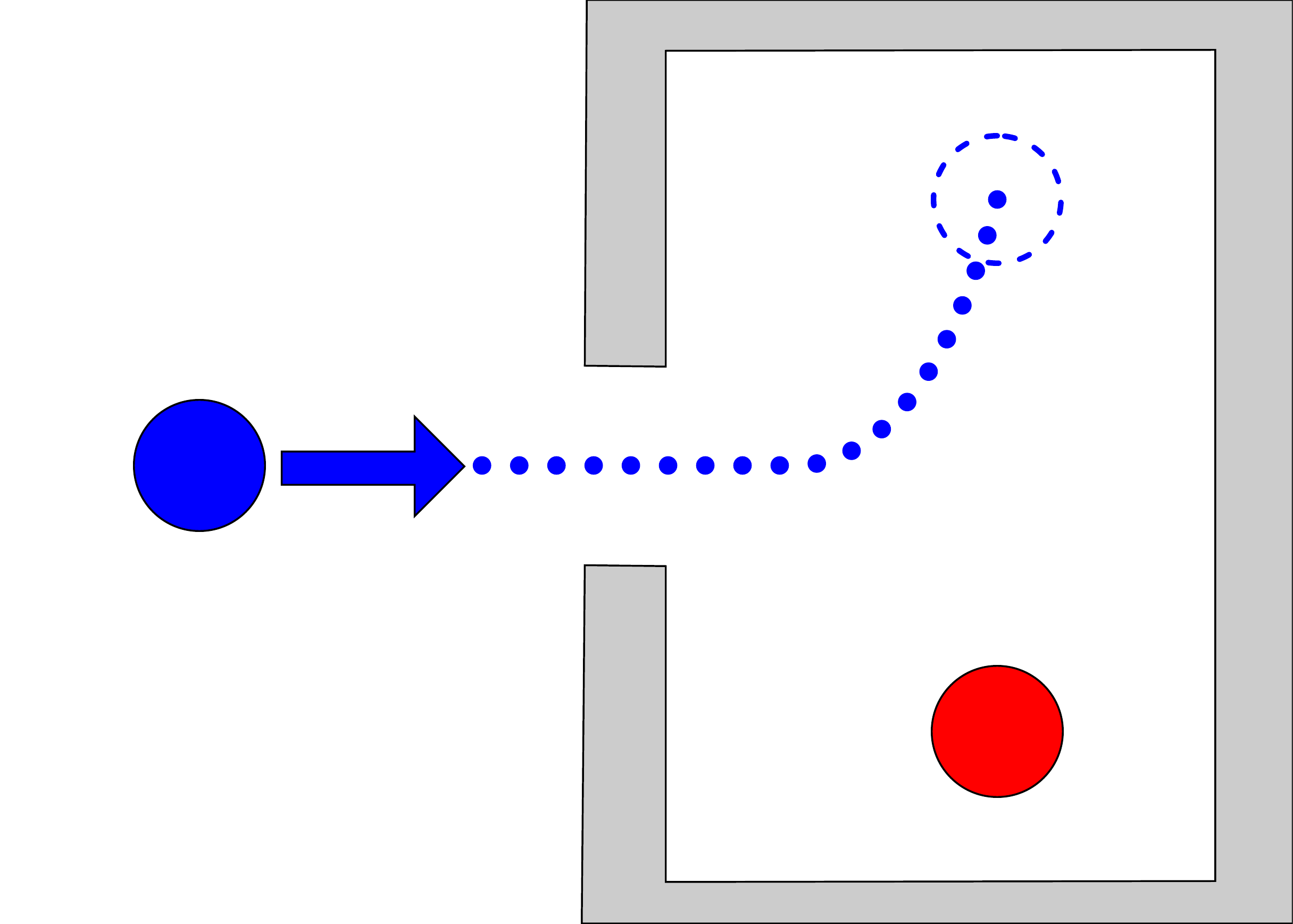}
     {\small \caption{Entering Elevator}}
    \end{subfigure}
    \hfill
    \begin{subfigure}[b]{0.19\textwidth}
     \centering
     \includegraphics[width=\textwidth]{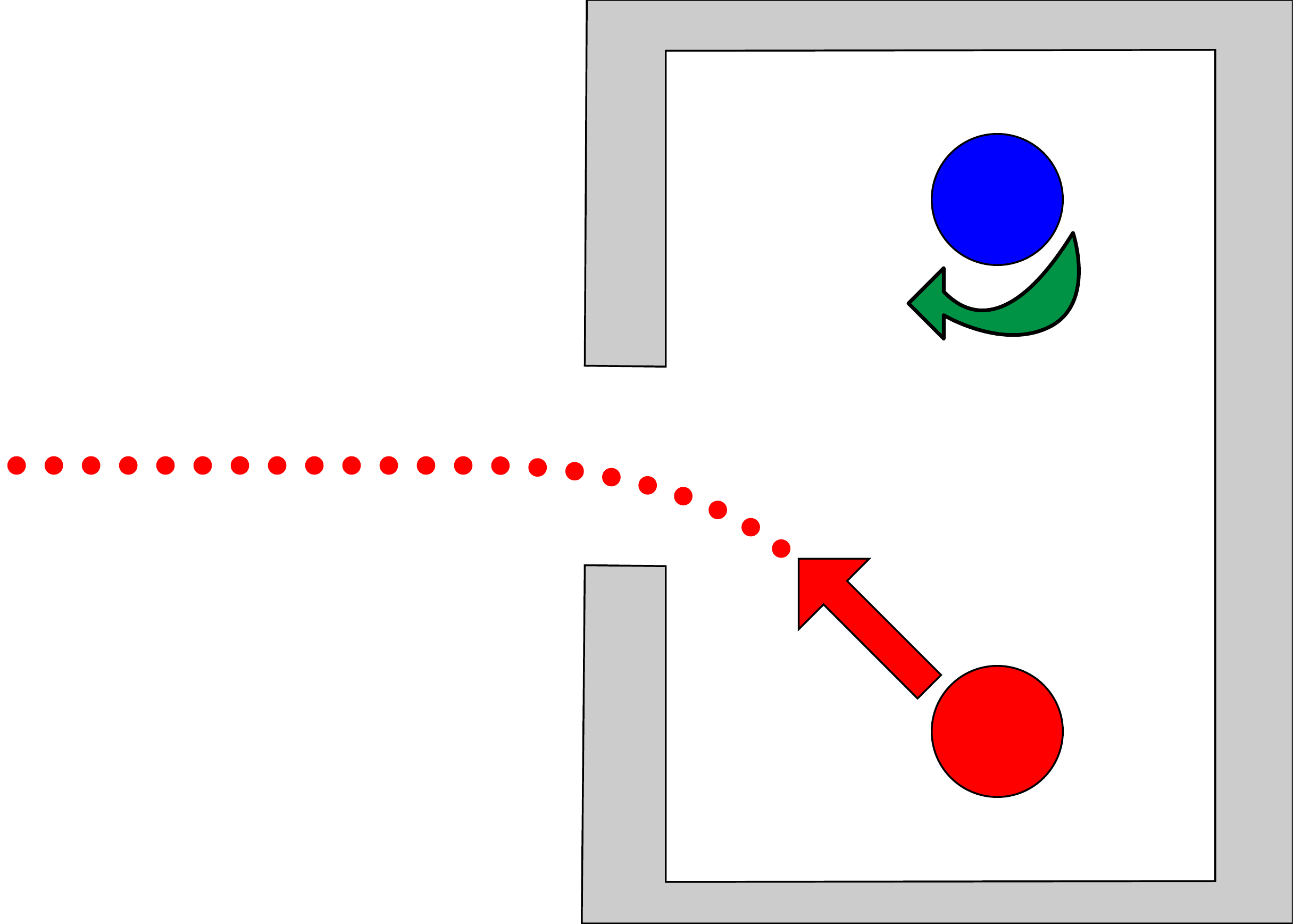}
     {\small \caption{Exiting Elevator}}
    \end{subfigure}
    \bigskip

    \begin{subfigure}[b]{0.19\textwidth}
     \centering
     \includegraphics[width=\textwidth]{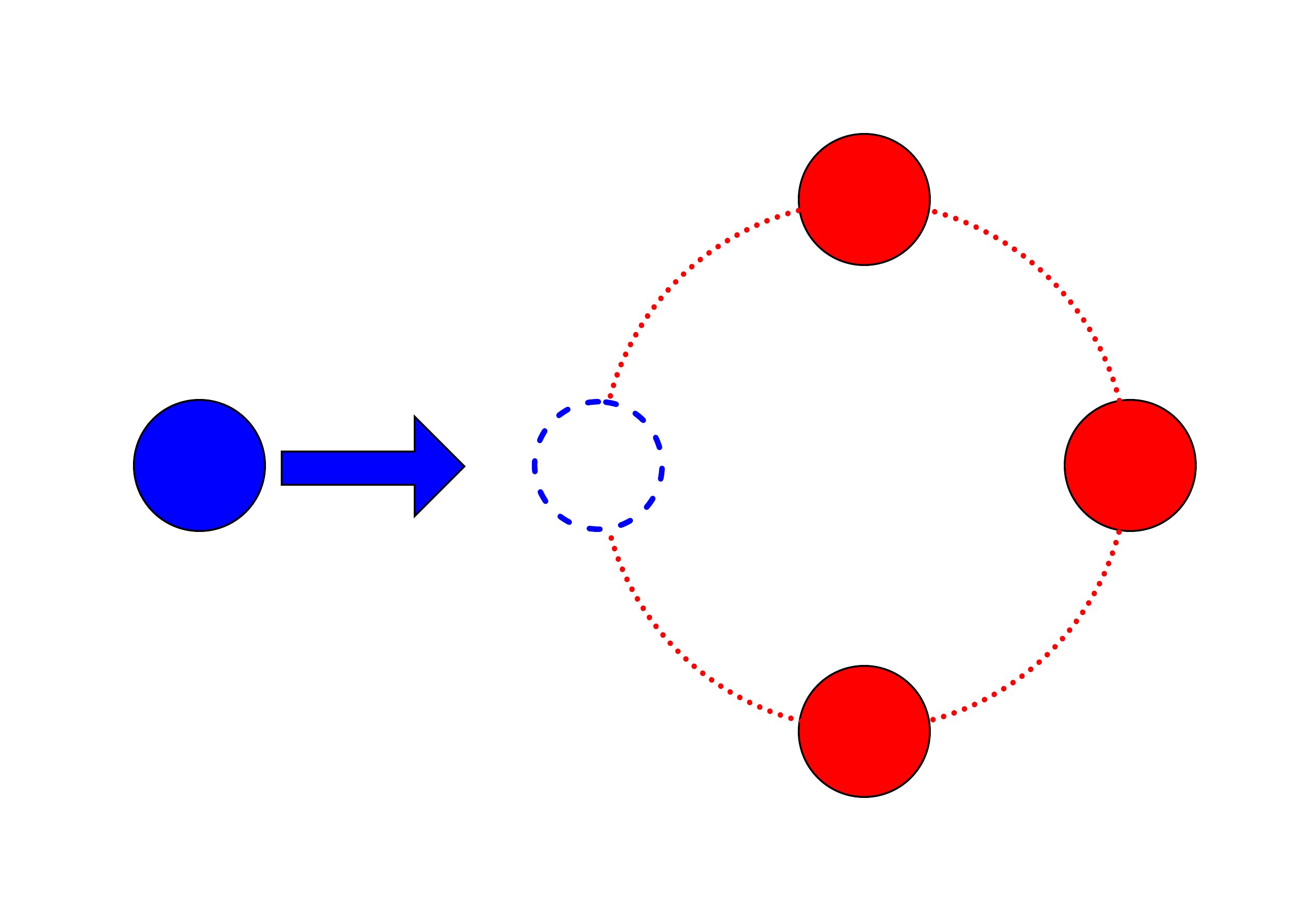}
     {\small \caption{Join Group}}
    \end{subfigure}
    \hfill
    \begin{subfigure}[b]{0.19\textwidth}
     \centering
     \includegraphics[width=\textwidth]{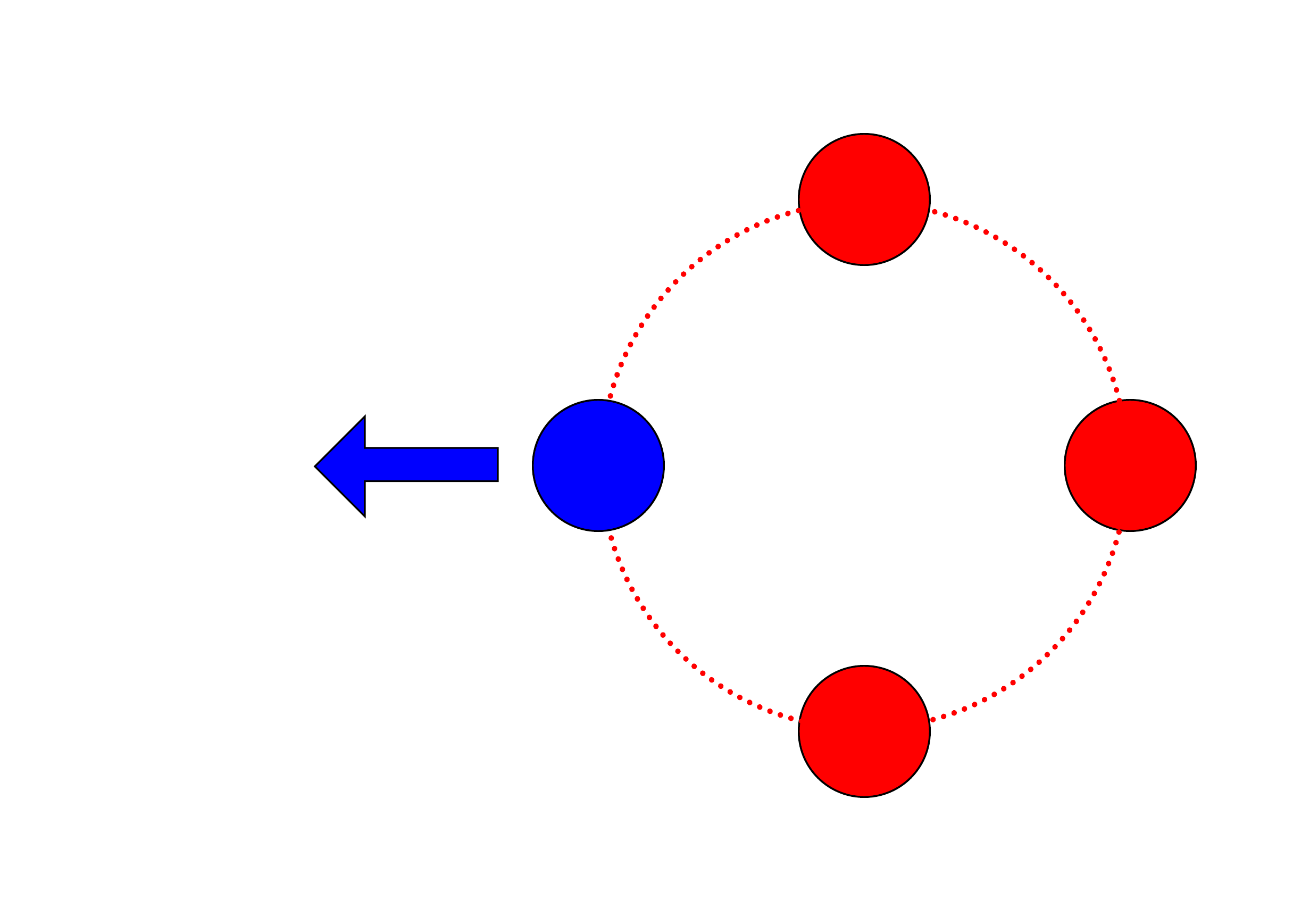}
     {\small \caption{Leave Group}}
    \end{subfigure}
    \hfill
    \begin{subfigure}[b]{0.19\textwidth}
     \centering
     \includegraphics[width=\textwidth]{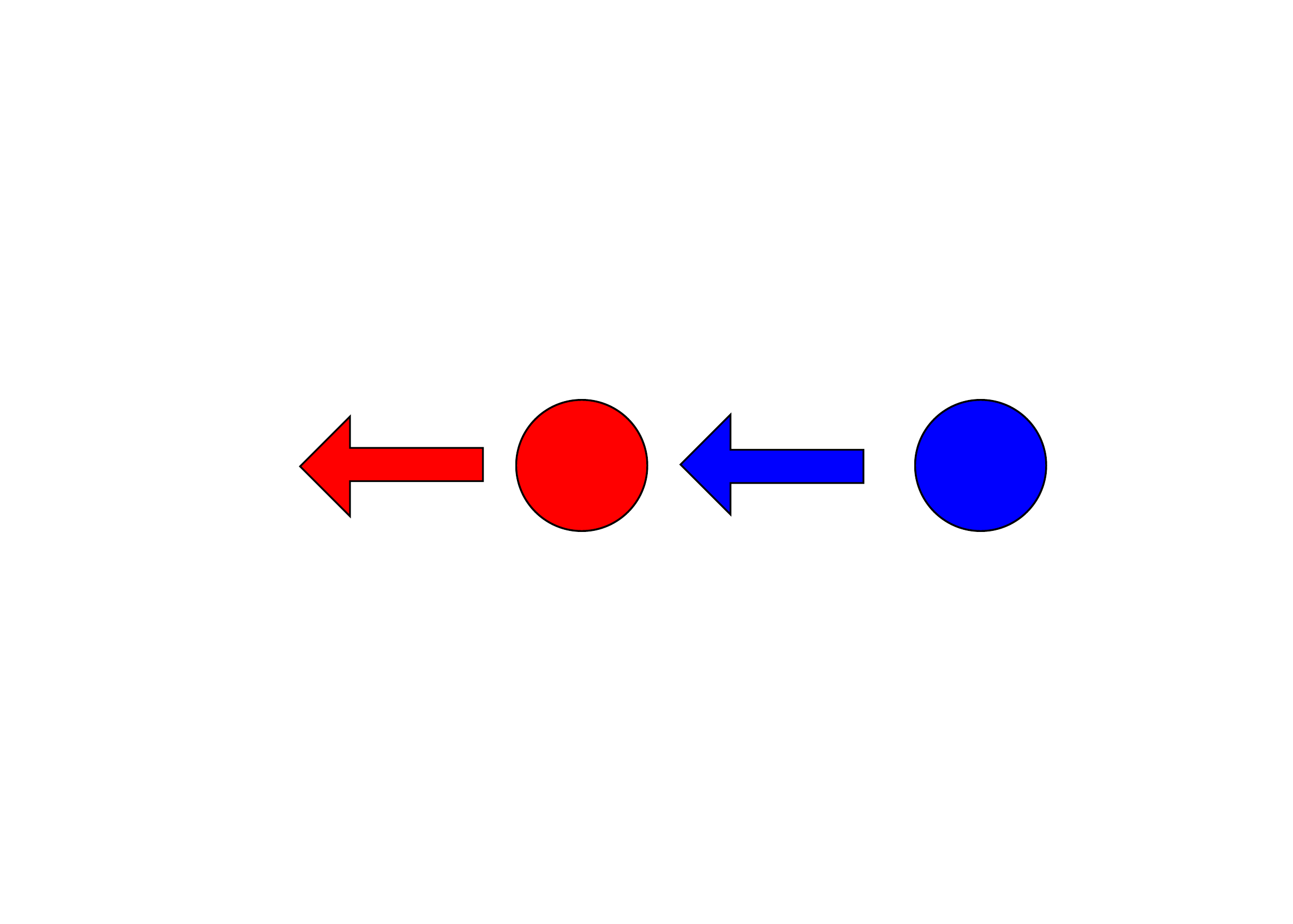}
     {\small \caption{Followng Human}}
    \end{subfigure}
    \hfill
    \begin{subfigure}[b]{0.19\textwidth}
     \centering
     \includegraphics[width=\textwidth]{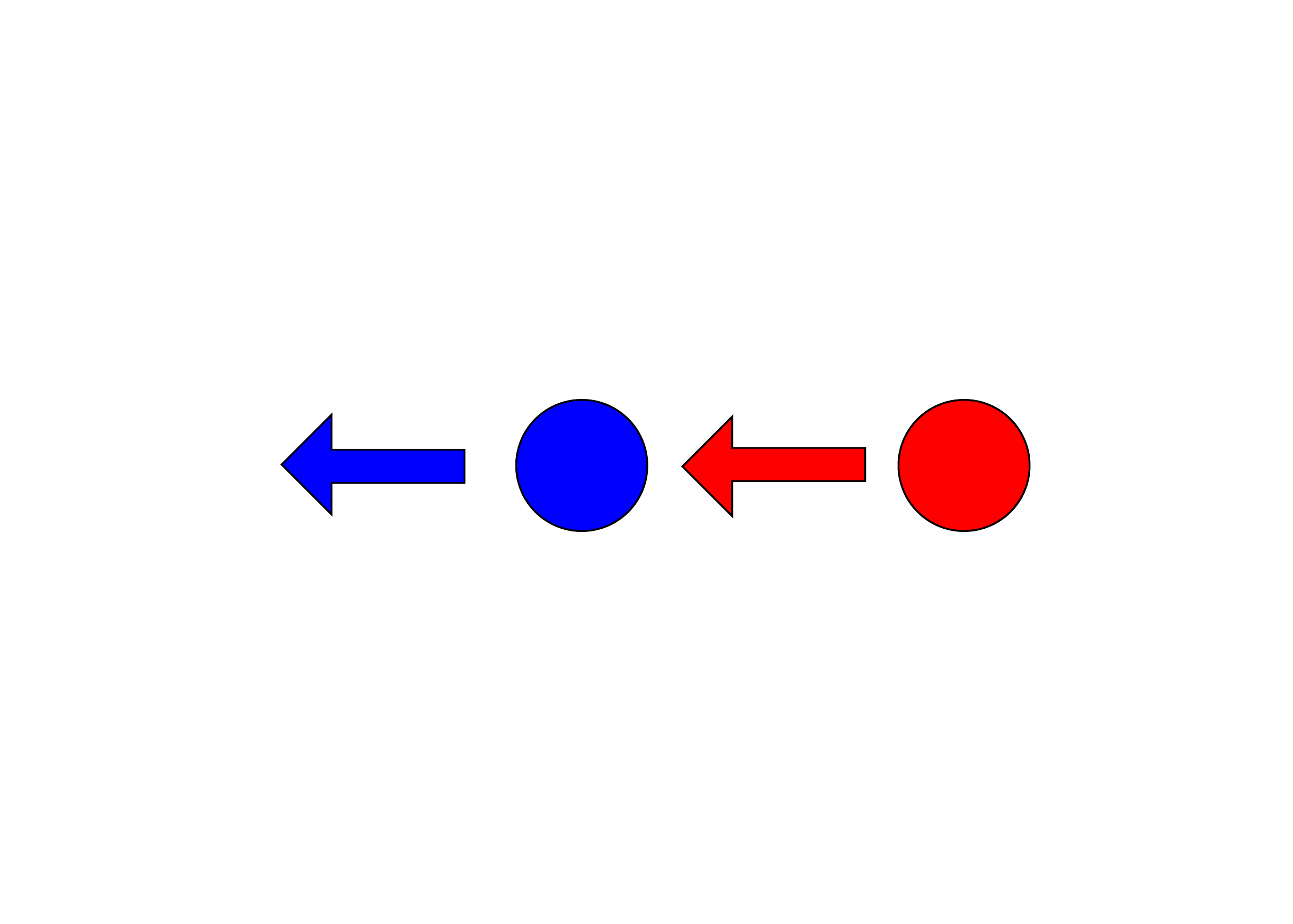}
     {\small \caption{Leading Human}}
    \end{subfigure}
    \hfill
    \begin{subfigure}[b]{0.19\textwidth}
     \centering
     \includegraphics[width=\textwidth]{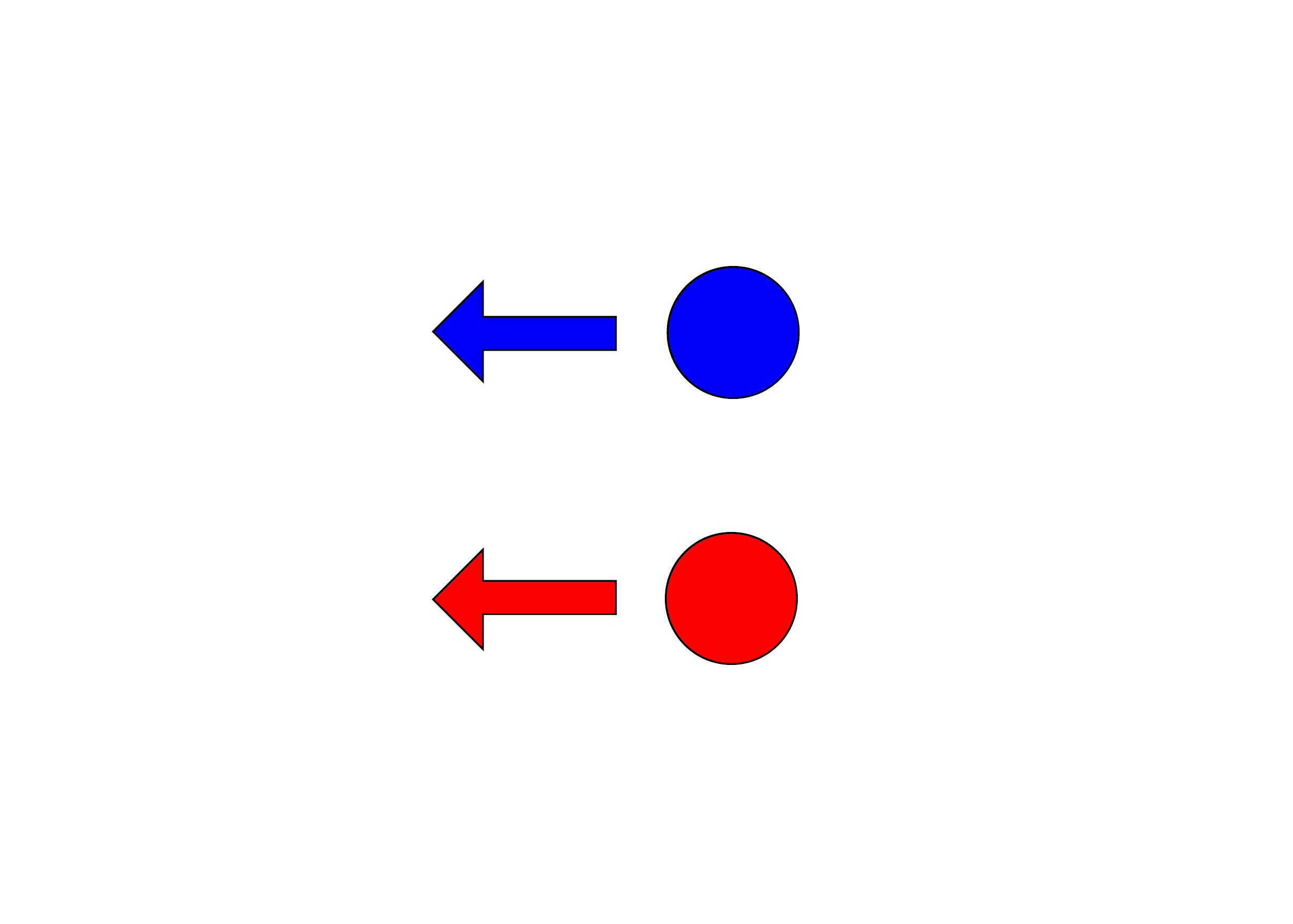}
     {\small \caption{Accompanying Peer}}
    \end{subfigure}
    \bigskip
 
    \begin{subfigure}[b]{0.19\textwidth}
     \centering
     \includegraphics[width=\textwidth]{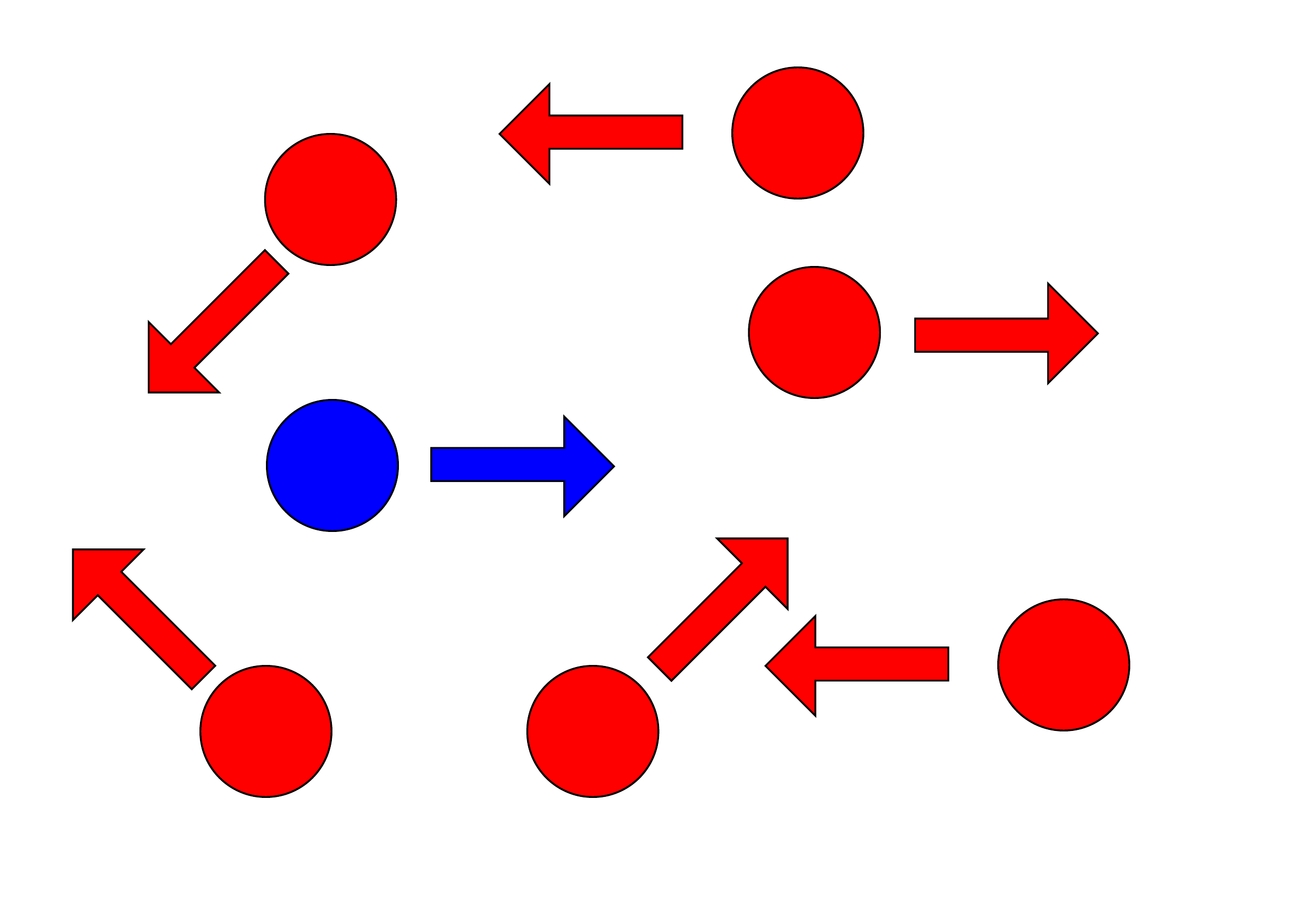}
     {\small \caption{Crowd Navigation}}
    \end{subfigure}
    \hfill
    \begin{subfigure}[b]{0.19\textwidth}
     \centering
     \includegraphics[width=\textwidth]{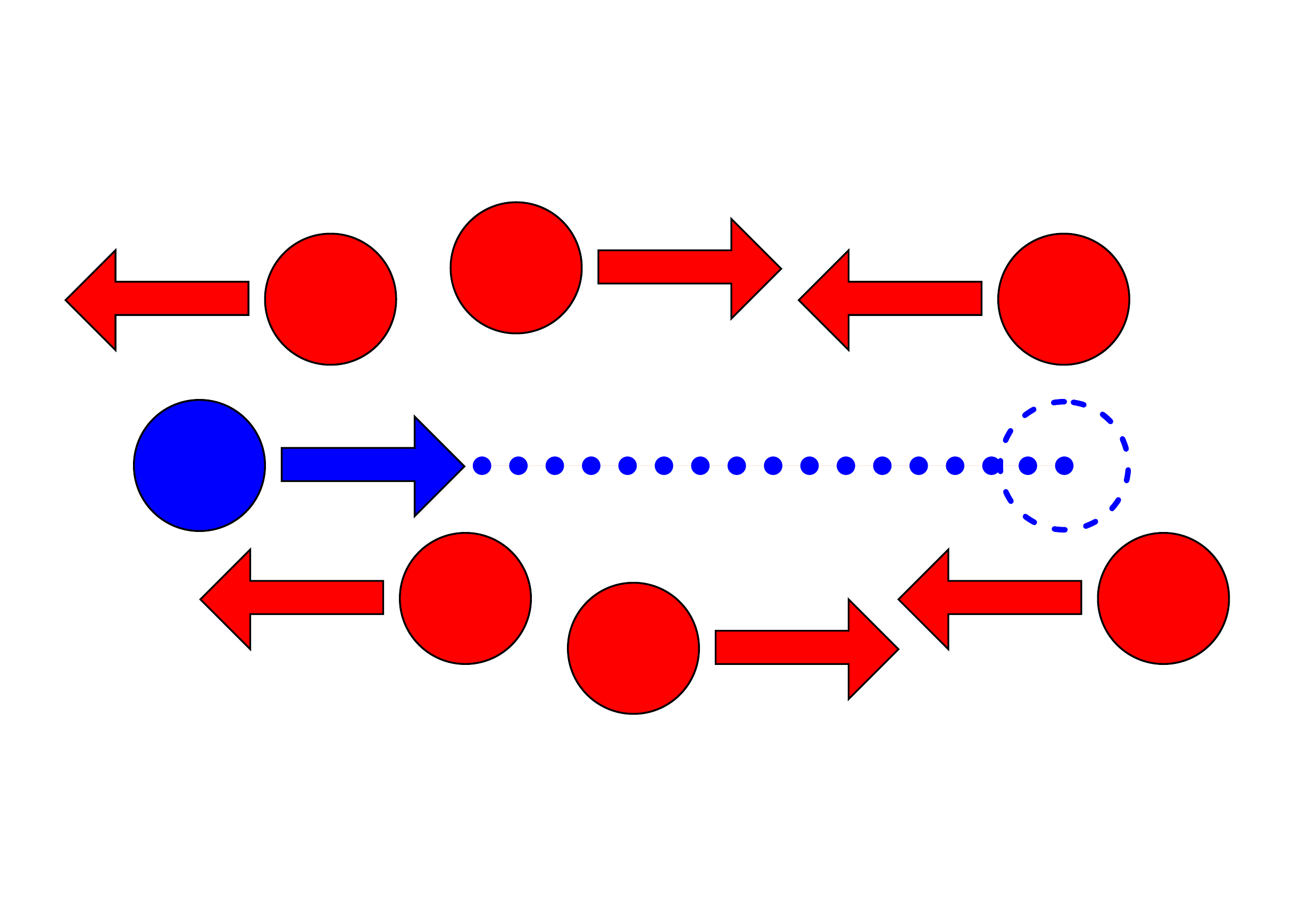}
     {\small \caption{Parallel Traffic}}
    \end{subfigure}
    \hfill
    \begin{subfigure}[b]{0.19\textwidth}
     \centering
     \includegraphics[width=\textwidth]{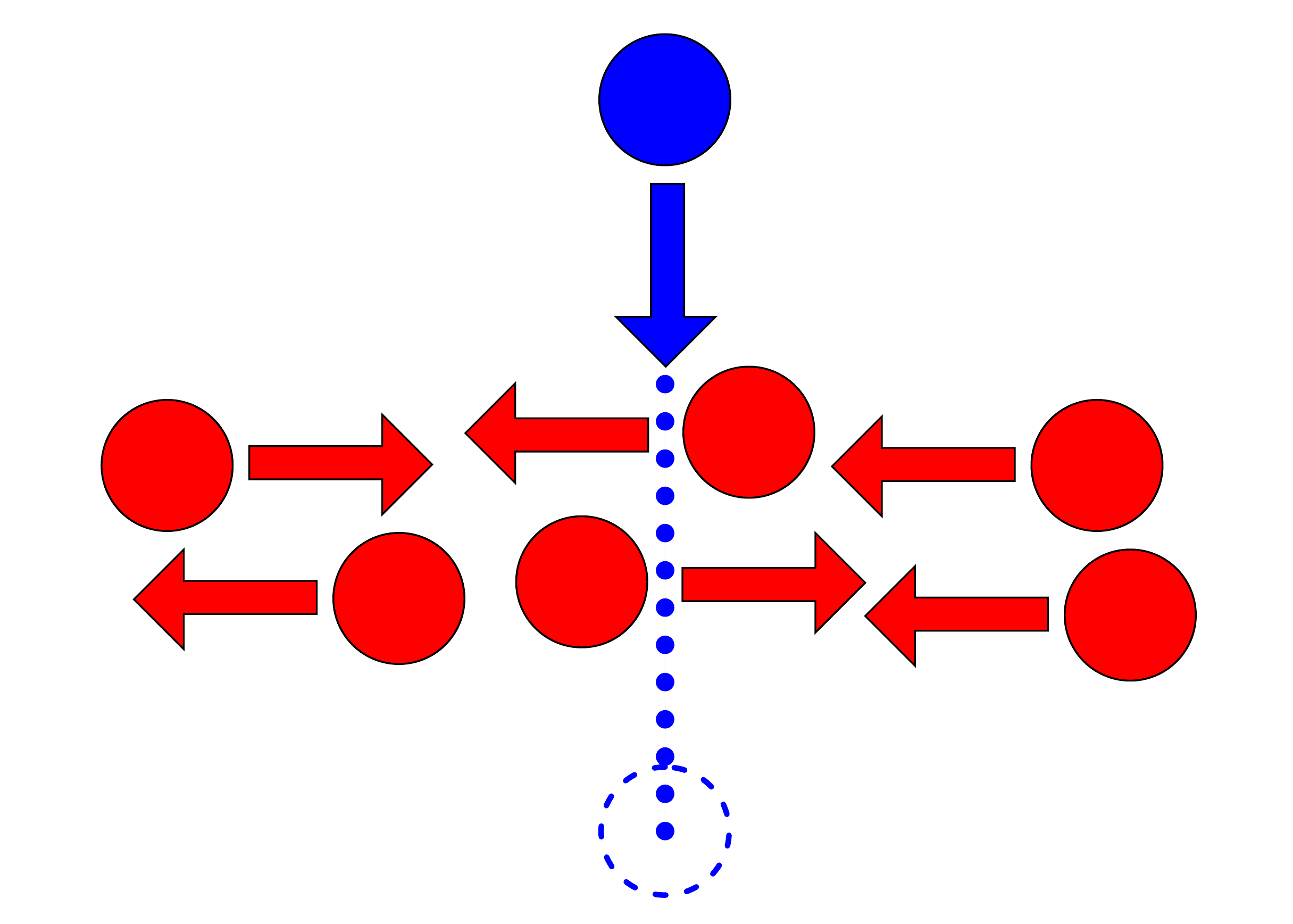}
     {\small \caption{Perpendicular Traffic}}
    \end{subfigure}
    \hfill
    \begin{subfigure}[b]{0.19\textwidth}
     \centering
     \includegraphics[width=\textwidth]{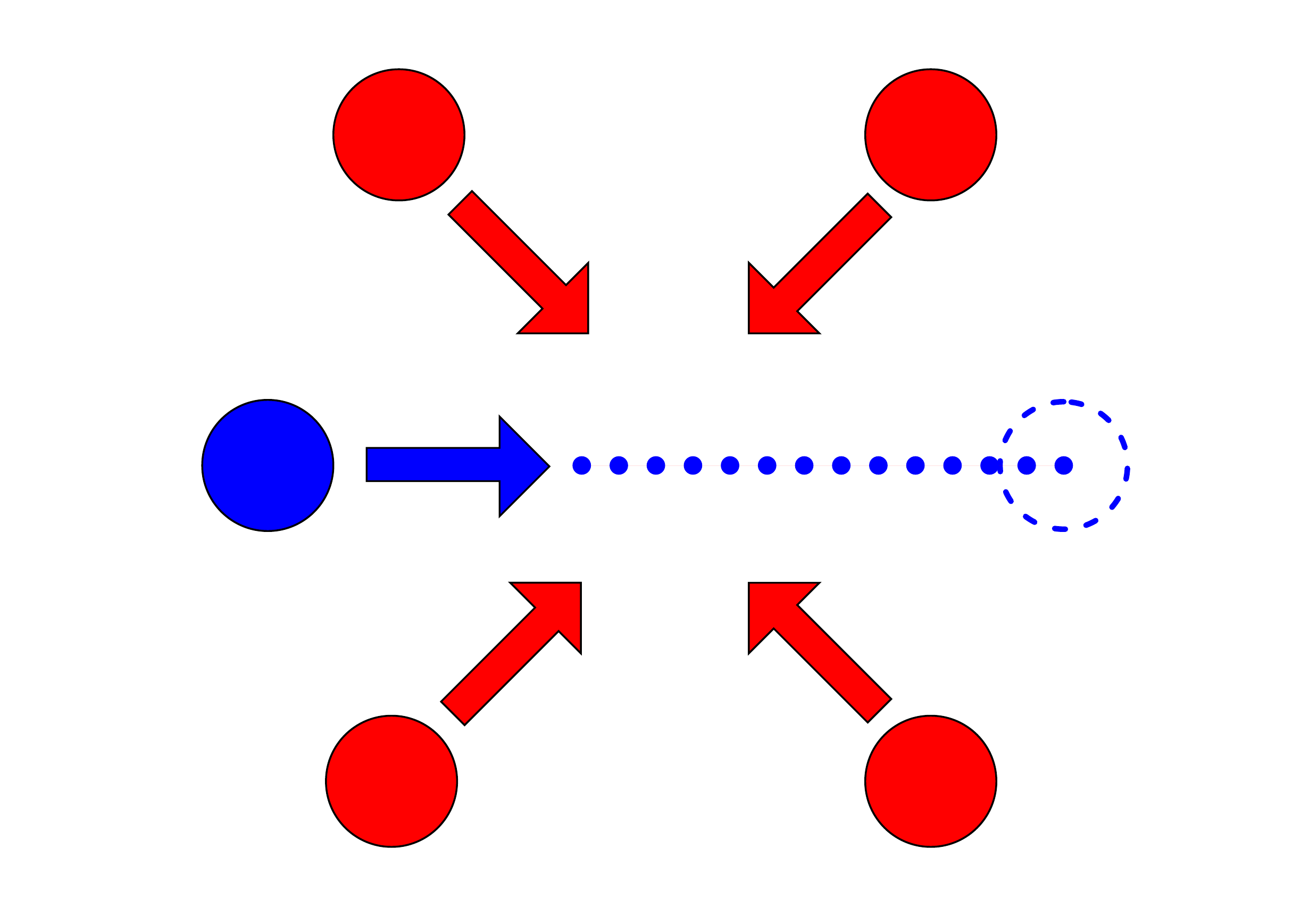}
     {\small \caption{Circular Crossing}}
    \end{subfigure}
    \hfill
    \begin{subfigure}[b]{0.19\textwidth}
     \centering
     \includegraphics[width=\textwidth]{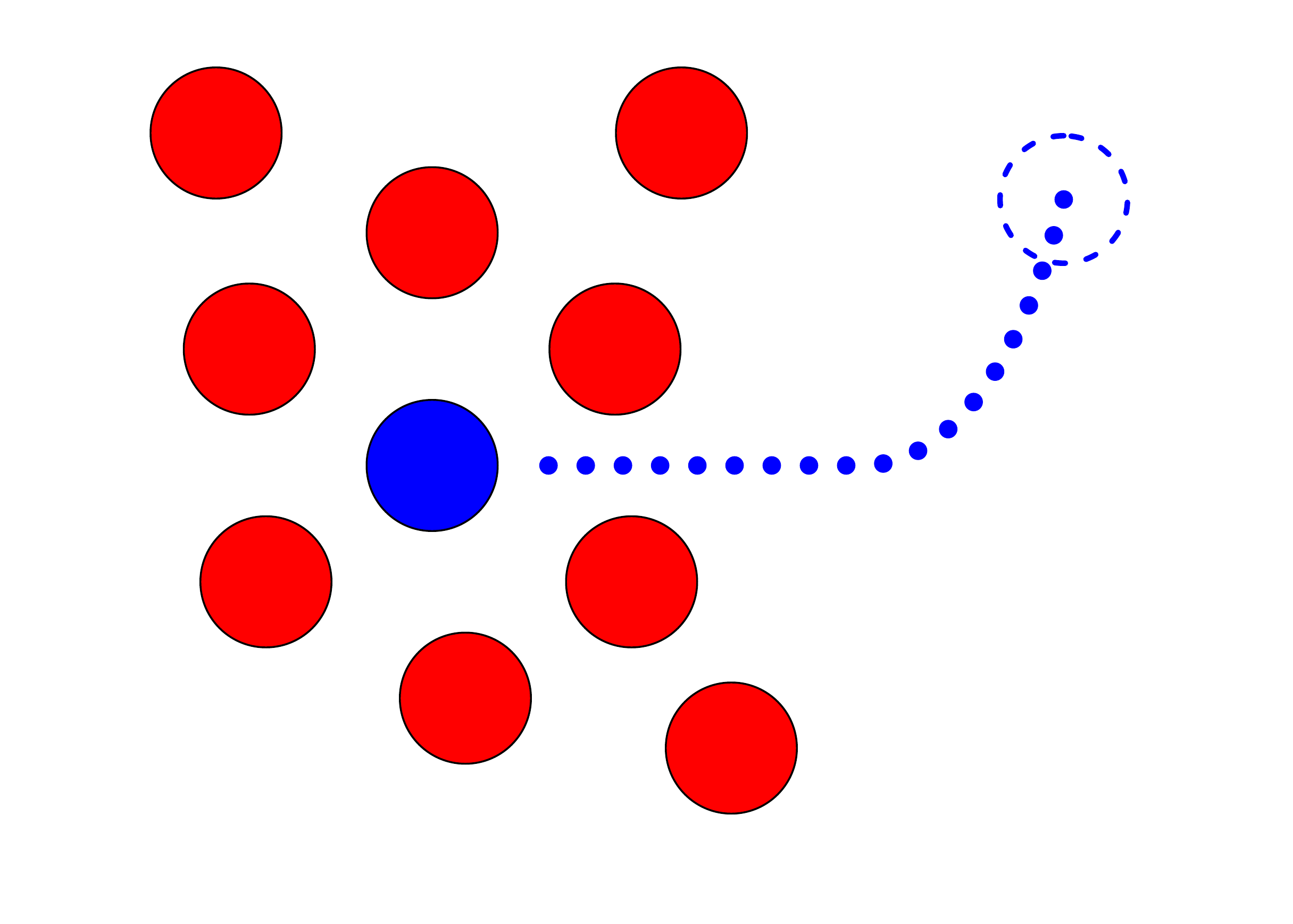}
     {\small \caption{Robot Crowding}}
    \end{subfigure}

    \caption{Geometric layout and intended human and robot behavior for example social navigation scenarios from Table~\ref{tab:scenarios}. The blue circles, arrows, dotted lines, and dotted circles represent the robot, its direction of motion, its intended path and its intended destination, respectively; red figures represent the corresponding items for humans. Grey backgrounds represent obstructions, while green figures represent signals or gestures emitted by an agent such as a gesture to stop or go ahead. Note sample paths and gestures are provided as examples to make the graphics clear; the actual scenario card definition should be flexible enough to capture a range of behaviors.}
    \label{fig:scenarios}
\end{figure*}

\subsection{Example Social Navigation Scenarios}
To effectively evaluate social navigation policies, they should be exercised in a set of scenarios which address the common use cases that come up in their intended context. For example, policies for interacting with pedestrians in low-traffic areas should be tested in common hallway and doorway scenarios, policies designed to navigate through crowds should handle common scenarios like traveling parallel to or perpendicular to the flow of traffic, and policies for interaction should handle scenarios like leaving and joining groups.

Ideally, policies should be evaluated using standard benchmarks as discussed in Section~\ref{s:benchmarks}; however, for a new research purpose a suitable benchmark may not yet exist. Nevertheless, researchers should try to find scenarios are already in use in the field and apply them as comprehensively as possible so that the evaluation of policies is meaningful and cam be reasonably compared to other work in the literature.

To facilitate this process, we summarize common social navigation scenarios in Table \ref{tab:scenarios}. Dozens of social navigation scenarios have been proposed in the literature, and we cannot list them comprehensively; however, we provide references above to relevant prior work using these scenarios where available. Also note that scenarios can be grouped into a broad variety of scientific purposes, including pedestrian navigation, crowd navigation, and interaction scenarios, which can help guide researchers in their selections. 
\begin{enumerate}
    \item \textbf{Pedestrian Navigation} Low-density pedestrian navigation scenarios study how pedestrians interact with robots a few at a time, and include common hallway and door interaction scenarios. Common pedestrian scenarios often include \textsc{Frontal Approach} where a human pedestrian approaches a moving robot head-on, \textsc{Robot Overtaking} where a robot overtakes a slower-moving human, \textsc{Intersection} where a robot passes a human at a right angle, \textsc{Blind Corner} where a robot and a human pass each other at an angle with poor visibility, \textsc{Narrow Doorway} where which a robot and a human attempt to exit a doorway in opposite directions, and so on.
    \item \textbf{Crowd Navigation} High-density crowd navigation scenarios study how robots can navigate dense human crowds, which many researchers argue exhibit qualitatively different behavior than pedestrians \cite{}. Common scenarios for crowd navigation include \textsc{Parallel Traffic} where a robot is going with or against the flow of moving pedestrians, \textsc{Perpendicular Traffic} where the robot must cross a flow of pedestrians, \textsc{Circular Crossing} and \textsc{Random Crossings} where pedestrians are crossing a plaza or room, and even \textsc{Robot Crowding} where a robot is surrounded by stationary pedestrians and must extricate themselves.
    \item \textbf{Interaction} Interaction scenarios involve a task that places constraints on robot navigation, such as group navigation skills like \textsc{Joining Groups} of pedestrians in conversations, \textsc{Leaving Groups} of pedestrians, or interactive skills such as \textsc{Object Handover} where robots deliver or receive an item, \textsc{Question Answering} where robots answer or ask questions, and \textsc{Continuous Monitoring} where a robot observes individuals exercising or performing another activity.
\end{enumerate}

The columns of Table \ref{tab:scenarios} capture many of the features of the Social Navigation Scenario card, though we cannot list all of them for space. Referring back to our running example, Table \ref{tab:frontal} shows an example of how the Social Navigation Scenario Card could be retroactively applied to one of the most common social navigation scenarios, \textsc{Frontal Approach}, which appears in \cite{pirk2022protocol, gao2021evaluation, wang2022metrics} among others.

\subsection{Scenario Guidelines}
Scenario guidelines can be broken into three groups following the methodology outlined above: guidelines for new scenario development, and guidelines for evaluating scenarios for research purposes, and guidelines for communication. For new scenarios, we propose the following guidelines:

\textbf{Guideline N1: Specify Research Context.} New social navigation scenarios should clearly define the research context under which they are expected to apply. 

\textbf{Guideline N2: Define Intended Robot Task.} New social navigation scenarios should clearly define the task the robot is expected to accomplish and not just start / end targets for navigation alone. 

\textbf{Guideline N3: Define Intended Human Behavior.} Scenarios should specify what human participants are intended to do in the scenario.

\textbf{Guideline N4: Define Success Metrics.} Scenarios should include metrics to gauge the success or failure of the task.

To evaluate the usefulness of scenarios, we recommend:

\textbf{Guideline N5: Cover Common Scenarios.} To adequately evaluate social navigation algorithms, researchers to try to include good coverage of scenarios which are used commonly in the field, such as those listed in Table~\ref{tab:scenarios}. 

\textbf{Guideline N6: Ensure Scenario Flexibility. } Scenarios should be broadly specified enough to capture the full range of behaviors that can occur.

\textbf{Guideline N7: Evaluate Fitness for Purpose. } Scenarios should identify or elicit the desired behaviors and enable the desirable properties of robot behavior to be evaluated.

Finally, we recommend the use of scenario cards as a standard communication format:

\textbf{Guideline N8: Use Scenario Cards.} When communicating scenarios - either new scenarios, or specializations of scenarios used for specific research purposes - use the scenario card format to clearly communicate scenario content.


\section{Social Navigation Benchmarks}
\label{s:benchmarks}

Social navigation benchmarks improve upon individual laboratory experiments or well-defined scenarios by collecting a set of scenarios into a benchmark suite with well-specified metrics, enabling the comparison of a variety of different methods against each other. However, existing benchmarks focus on different aspects of the social navigation problem outlined in Section~\ref{s:definition}, using different permutations of the factors we outlined in Section~\ref{s:taxonomy}. Hence, the results of these benchmarks may be more or less useful for researchers investigating different aspects of the social navigation problem. In this section, we advocate a set of criteria to make benchmarks useful across the social navigation community, and review existing benchmarks in use with regards to these criteria. 

First, we analyze benchmarks in use in the social navigation community, grouping them into benchmarking protocols, benchmarking environments, and benchmark challenges. Then, we analyze the strengths and weaknesses of these benchmarks, abstracting out criteria for good social navigation benchmarks, including evaluating social behavior using quantitative metrics and well-validated questionnaires grounded in human data. Finally, we make recommendations on how to improve the state of social navigation benchmarking, and discuss how social benchmarking could be integrated with standard navigation benchmarks as regression tests of navigation behavior, which ensure that previously successful behaviors do not degrade as changes are made \cite{okal2016learning,wahl1999overview}.

\subsection{Expanding the Factors for Benchmark Analysis}
In addition to the factors listed in Section \ref{s:factors}, additional aspects must be considered for benchmarks:

\paragraph{Simulation Platform} Benchmarks must specify how to set up an evaluation, but are more useful if that evaluation is already set up on a commonly available simulation platform.
\paragraph{Associated Dataset} Some benchmarks specify one or more datasets of reference behaviors used for comparisons.
\paragraph{Provided Baselines} Some benchmarks specify a set of baseline policies which can be used for comparisons.
\paragraph{Challenge Leaderboard} Benchmark challenges may also provide a leaderboard to enable policies among different teams to be compared publicly.
\paragraph{Downloadability} Ideally, a benchmark should include a downloadable software suite to enable replication of results.
\paragraph{Most Recent Update} Because software platforms evolve, benchmarks should be updated recently to ensure they are usable with current hardware and software.
\paragraph{Robot Hardware Platform} To make benchmarks most useful, they should support a wide variety of robot morphologies or custom robot morphologies so researchers have the best chance of generating comparisons for their target platform.
\paragraph{Human Behavior Authoring Methods} Benchmarks must include agents other than the robot, whether human or other robots. Support for realistic human behavior or replayed datasets can improve a benchmark's fidelity and usefulness.

\begin{table*}[t]
    \centering\scriptsize
    \begin{tabular}{|R{0.11\linewidth} |L{0.08\linewidth} |L{0.07\linewidth} | L{0.07\linewidth}| L{0.07\linewidth}| L{0.07\linewidth}| L{0.07\linewidth}| L{0.07\linewidth}| L{0.07\linewidth}| L{0.07\linewidth}|}
    \hline
    \textbf{Benchmark} & \textbf{ArenaBench} & \textbf{CrowdBot} & \textbf{DynaBarn} & \textbf{gym-collision-avoidance} & \textbf{HuNavSim} & \textbf{iGibson} & \textbf{Soc-NavBench} & \textbf{Sea-NavBench} & \textbf{Social Navigation Protocol}\\
    \hline
    \multicolumn{9}{l}{\textbf{Factors for Analysis}} \\
    \hline
    Benchmark Classification & Benchmark & Challenge & Benchmark & Benchmark & Benchmark & Challenge & Benchmark & Challenge & Protocol\\
    \hline
    Benchmark Context and Scope  & Dynamic Obstacle Benchmark & Crowd \mbox{Simulation} Benchmark & Dynamic \mbox{Obstacle} Benchmark & Collision Avoidance Benchmark & Human \mbox{Simulation} Benchmark & Social Navigation Benchmark & Social Navigation Benchmark & Social Navigation Benchmark & Human-Robot Expt. Design \\
    \hline
    Physical \mbox{Environment}  & Indoor & Indoor & Synthetic & Synthetic & Indoor & Indoor & Indoor \&~\mbox{Outdoor} & Indoor \&~\mbox{Outdoor} & Principally \mbox{Indoor}\\
    \hline
    Intended~Human User~Type  & Synthetic Pedestrian& Synthetic \mbox{Pedestrian} & Varied Human Motion & Synthetic \mbox{Pedestrian} & Varied \mbox{Resp} to Robot & Synthetic \mbox{Pedestrian} & Synthetic \mbox{Pedestrian} & Synthetic \mbox{Pedestrian} & Human \mbox{Coworkers} \\
    \hline
    Supported Robot~Tasks & Navigation & Navigation & Navigation & Navigation & Navigation & Navigation & Navigation & Navigation & Navigation \\
    \hline
    Social Scenarios Evaluated & 3 Worlds, 5/10 Peds& Basic Crowd Scenarios & 60 Crowd Scenarios & Multi-agent Scenarios & House, Cafe, Warehouse & 15 House Scenes & 5 Curated Environments & TBD & 6 Social Nav Scenarios\\
    \hline
    Coverage of Corner~Cases & Diversity, Random & Not Tested & Diversity, Random & Not Tested & Not Tested & Not Tested & Not Tested & TBD & Not Specified\\
    \hline
    Simulation Platform  & Flatland, Gazebo & Unity & Gazebo & Custom & Gazebo & iGibson & SocNavBench & SEAN 2.0 & None\\
    \hline
    Benchmarking Dataset  & None  & CrowdBot & None & None & None & None & UCY \& ETH & UCY \& ETH & None\\
    \hline
    Human Behavior Authoring  & Pedsim & UMANS & Multiple Algorithms & Baseline Policies & Soc.~Force, Behav.~Tree & ORCA & Replay, Planned & Replay, Soc~Force & Scripted\\
    \hline
    Human Simulation Fidelity  & Walking Humans  & Walking Humans & Moving Cylinders & Moving Cylinders & Walking Humans & Moving Humans & Walking Humans & Walking Humans & Real Humans \\
    \hline
    Supported Robot Embodiments  & Jackal, Burger, Robotino & Pepper, Wheelchair, CuyBot, Qolo & Custom Robots, ClearPath Jackal & Cylinders & ROS Gazebo-Compatible & 8~real, 2~Mujoco & Simulated Mobile Robot & Fetch, Jackal, Turtlebot, Warthog & Human-Scale Robots\\
    \hline
    Communication Modalities  & None & None & None & None & None & None & None & None & Human Gestures \\
    \hline
    Challenge Leaderboard  & None & None & None & None & None & 2021 & None & 2022 & None\\
    \hline
    Benchmark Last Updated  & 2022 & 2021 & 2023 & 2022 & 2023 & 2021 & 2022 & 2022 & 2022 \\
    \hline
    \multicolumn{9}{l}{\textbf{Guidelines for Benchmarks}} \\
   
    \hline
    B1: Evaluate Social Behavior  & Yes & Yes & Yes & Yes & Yes & Yes & Yes & Yes & Yes \\
    \hline
    B2: Quantitative Metrics Provided & Many & Many & Succ.~Rate & Succ.~Rate, Time2Goal & Many & Succ.~Rate, PSC & Many & Many & No \\
    \hline
    B3: Baseline Policies Provided  &SOA Nav, RL~Policies& No & SOA Nav, RL~Policies & SOA Social, Worst-Case & No & SOA RL & SOA Social, Worst-Case & SOA Social, Worst-Case & No \\
    \hline
    B4: Scalable, Repeatable  & Simulated, Download & Simulated, Download & Simulated, Download  & Simulated, Download & Simulated, Download & Simulated, Download & Simulated, Download & Simulated, Download & Setup~Req., Phys.~Eval\\
    \hline
    B5: Eval Grounded in Human Data  & No & No & Demo. pipeline & No & No & No & No & SEAN-EP extension & Yes \\
    \hline
    B6: Use Validated Instruments  & No  & No & No & No & No & No & No & No & Validation in Process \\
    \hline
    \end{tabular}
    \caption{Characteristics of existing social navigation benchmarks}
    \label{tab:benchmark-characteristics}
    
\end{table*}

\subsection{Existing Social Navigation Benchmarks}
In the social navigation literature, the term ``benchmark'' is sometimes applied to labeled datasets of reference behavior, which we discuss in Section~\ref{s:datasets}. In this section, we focus specifically on social navigation benchmarks which combine at least three components: (a) a social navigation system (such as a simulator) which can run algorithms and pedestrians (b) in well-defined scenarios (c) with metrics for evaluation; these benchmarks may optionally specify datasets of human or robot behavior for comparisons. Full benchmarks can be broken into three classes: 
1) \textit{benchmarking protocols} which enable the construction of experiments along well-specified principles, like the \textsc{Social Navigation Protocol}~\cite{pirk2022protocol},
2) \textit{benchmarking environments} which enable comparison of algorithms against baselines in environments, including \textsc{DynaBarn}~\cite{nair7dynabarn}, \textsc{gym-collision-avoidance}~\cite{Everett18_IROS}, \textsc{HuNavSim}~\cite{perez2023hunavsim}, and \textsc{SocNavBench}~\cite{biswas2022socnavbench}, and 
3) \textit{benchmark challenges} which also provide a platform or forum to share results, including \textsc{CrowdBot}~\cite{inria2020safe}, \textsc{iGibson}~\cite{li2021igibson, shen2021igibson}, and \textsc{SEANavBench}\footnote{\url{https://seanavbench.interactive-machines.com/}}. In the following, we describe these benchmarks; see Table~\ref{tab:benchmark-characteristics} for a side-by-side comparison based on the previously described factors and Figure~\ref{tab:benchmark-characteristics} for a visual description of some of the more commonly used benchmarks.

\subsubsection{Benchmarking Protocols}

The \textsc{Social Navigation Protocol} \cite{pirk2022protocol} is an industry benchmark proposed by Robotics at Google \cite{robotics-at-google} and used in \cite{pirk2022protocol}, \cite{xiao2022learning}, and \cite{cuan2022gesture2path} to evaluate the performance of a series of learning-based model predictive control policies for social robot navigation (though the protocol was intended to be applicable to the evaluation of any policy, learning or not).
This protocol involves selecting social navigation scenarios of interest, such as Frontal Approach, Blind Corner, Corridor Intersection, and so on. Each scenario's human-robot interaction is defined by the start and end of the robot trajectory and a short description of what is expected to happen for the human. This serves two purposes: enabling the collection of expert human trajectories for training social navigation policies, and evaluating policies on the same scenarios with low variability.  
Over the course of \cite{pirk2022protocol}, \cite{xiao2022learning}, and \cite{cuan2022gesture2path}, the protocol was iteratively improved. For example, the questionnaire proposed in \cite{pirk2022protocol} was analyzed in \cite{xiao2022learning} to identify reliable factors according to Cronbach's alpha, which  were used to update the questionnaire for \cite{cuan2022gesture2path}, which enabled more extensive analysis. While the \textsc{Social Navigation Protocol} can be applied to a wide variety of setups, it does not provide a downloadable, simulated environment, and must be manually set up for each experiment.

\begin{figure*}[tb!p]
    \centering
    \begin{subfigure}[b]{0.3\textwidth}
     \centering
     \includegraphics[width=\textwidth]{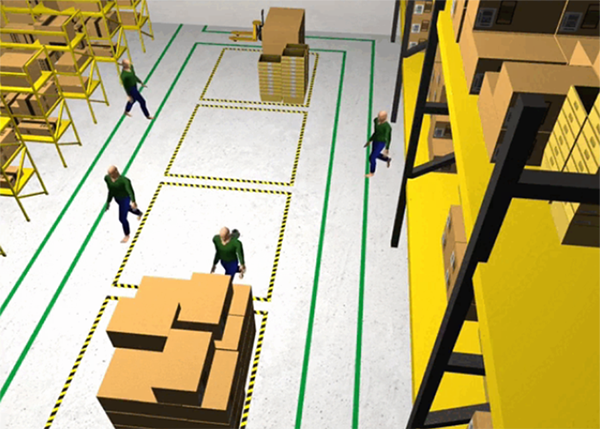}
     {\small \caption{ArenaBench}}
    \end{subfigure}
    \hfill
    \begin{subfigure}[b]{0.3\textwidth}
     \centering
     \includegraphics[width=\textwidth]{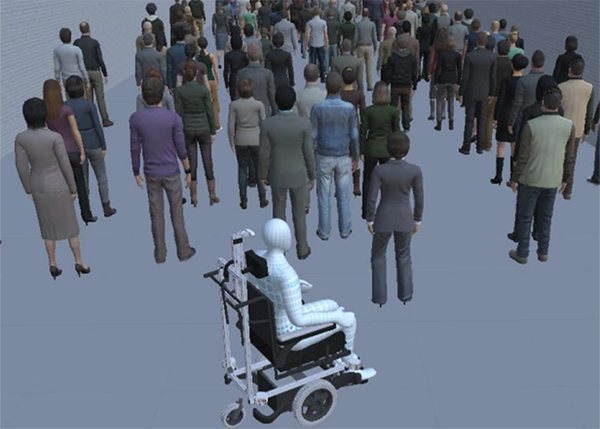}
     {\small \caption{CrowdBot}}
    \end{subfigure}
    \hfill
    \begin{subfigure}[b]{0.3\textwidth}
     \centering
     \includegraphics[width=\textwidth]{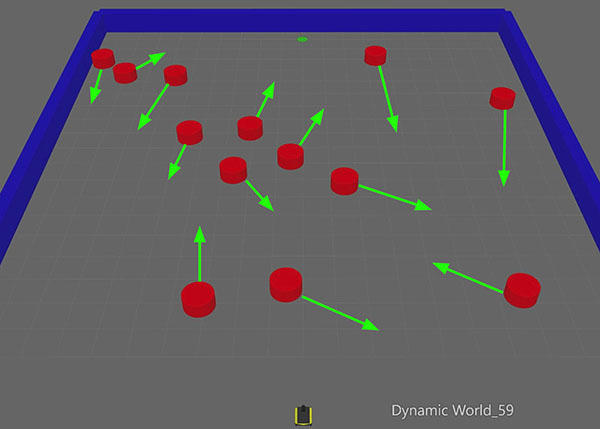}
     {\small \caption{DynaBarn}}
    \end{subfigure}
    \bigskip

    \begin{subfigure}[b]{0.3\textwidth}
     \centering
     \includegraphics[width=\textwidth]{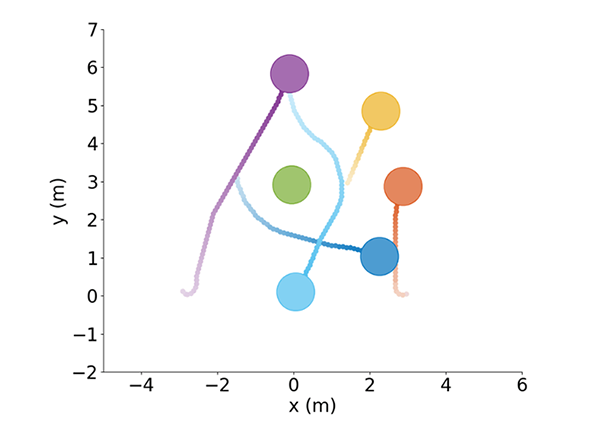}
     {\small \caption{gym-collision-avoidance}}
    \end{subfigure}
    \hfill
    \begin{subfigure}[b]{0.3\textwidth}
     \centering
     \includegraphics[width=\textwidth]{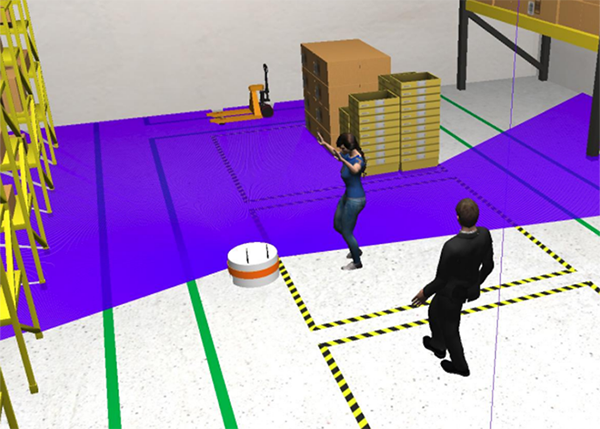}
     {\small \caption{HuNavSim}}
    \end{subfigure}
    \hfill
    \begin{subfigure}[b]{0.3\textwidth}
     \centering
     \includegraphics[width=\textwidth]{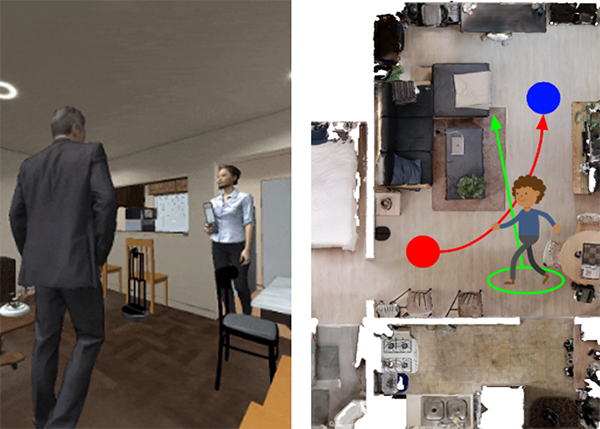}
     {\small \caption{iGibson}}
    \end{subfigure}
    \bigskip

    \begin{subfigure}[b]{0.3\textwidth}
     \centering
     \includegraphics[width=\textwidth]{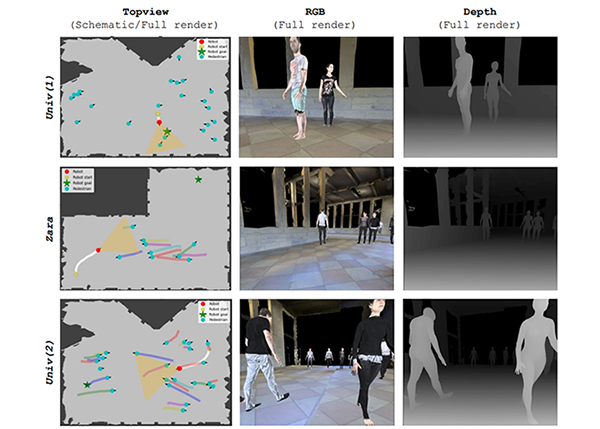}
     {\small \caption{SocNavBench}}
    \end{subfigure}
    \hfill
    \begin{subfigure}[b]{0.3\textwidth}
     \centering
     \includegraphics[width=\textwidth]{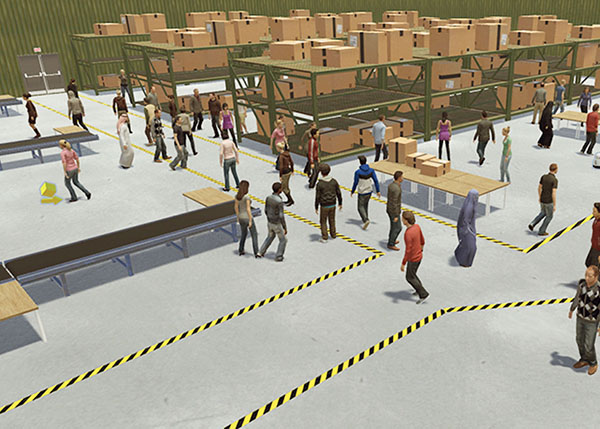}
     {\small \caption{SEANavBench in SEAN 2.0}}
    \end{subfigure}
    \hfill
    \begin{subfigure}[b]{0.3\textwidth}
     \centering
     \includegraphics[width=\textwidth]{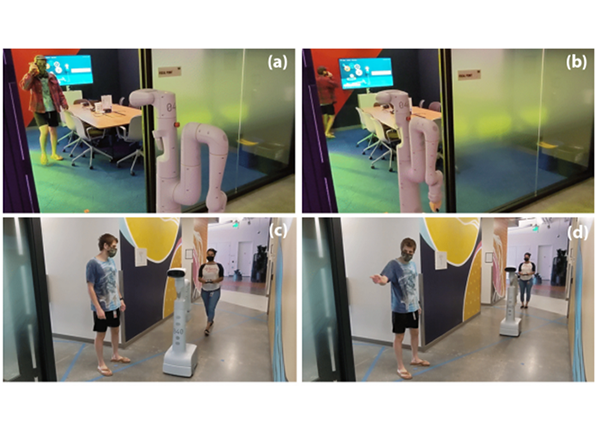}
     {\small \caption{Social Navigation Protocol}}
    \end{subfigure}
    \bigskip
    
    {\small \caption{Commonly used social benchmarks. Benchmarks range from abstract tests of dynamic obstacle avoidance to simulated interactions with moving humans of varying degrees of fidelity to protocols for setting up physical experiments in well-specified scenarios.}}
    \label{fig:benchmarks}
\end{figure*}

\subsubsection{Social Navigation Benchmarks}
\textsc{ArenaBench}~\cite{kastner2022arena} 
is a downloadable, simulated social navigation benchmark designed to test how navigation algorithms perform under different tasks. Building on the 2D Flatland\footnote{\url{https://flatland-simulator.readthedocs.io/en/latest/}} and 3D Gazebo~\cite{koenig2004design}
simulators and the Pedsim~\cite{gloor2016pedsim}
implementation of the Social Forces Model (SFM)~\cite{helbing1995social}, \textsc{ArenaBench} provides the ability to evaluate both classical and learning-based approaches in the ROS~\cite{quigley2009ros}
framework. In addition to providing tools for automatically and manually creating scenarios, \textsc{ArenaBench} supplies both the non-learned baselines MPC~\cite{rosmann2017time}, DWA~\cite{fox1997dynamic}, TEB~\cite{rosmann2012trajectory} and the learned baselines NAVREP~\cite{dugas2021navrep}, Gring~\cite{guldenring2020learning} as well as \textsc{ArenaBench}'s own trained ROSNAV approach.
Supported robots include the Robois Turtlebot3,
ClearPath Jackal,
and Festo Robotino 4.
ArenaBench provides a variety of navigation metrics including success rate, collision, time to goal, path length, velocity, acceleration, jerk, curvature, angle over length, roughness, and clearing distance. However, \textsc{ArenaBench} does not at this time support human evaluation of robot behavior.

\textsc{DynaBARN}~\cite{nair7dynabarn} is a downloadable, simulated social navigation benchmark designed to test how algorithms respond to a variety of different pedestrian models. Building on the \textsc{barn} navigation benchmark~\cite{perille2020benchmarking}, \textsc{DynaBARN} provides 60 environments in the Gazebo simulator. \textsc{DynaBARN} evaluates algorithms against social behavior through crowds of cylindrical pedestrians controlled by motion trajectories specified by polynomials of different orders and different numbers of pedestrians. It is customizable to different robot platforms, with a Jackal provided. Only success rate (collision-free navigation reaching the goal) is provided as a metric, though the platform is extensible. \textsc{DynaBARN} provides several baselines including DWA~\cite{fox1997dynamic}, TEB~\cite{rosmann2012trajectory}, a behavior cloned (BC)~\cite{pomerleau1989alvinn} policy, and a TD3~\cite{fujimoto2018addressing} RL policy. While \textsc{DynaBARN} does not support human evaluation of robot behavior, it includes a demonstration pipeline to collect human teleoperation baselines of navigation in dynamic environments. 

\textsc{gym-collision-avoidance}~\cite{Everett18_IROS} is a downloadable, simulated benchmark used to evaluate multi-agent collision avoidance. Created to evaluate the GA3C-CADRL algorithm~\cite{Everett18_IROS,everett2019collision} against the baselines ORCA~\cite{van2011reciprocalORCA}, SA-CADRL~\cite{chen2017sociallyawareDRLmit_iros17}, and DRLMACA~\cite{long2018towards}, this benchmark provides a variety of multi-agent scenarios involving cylinders in simplified synthetic environments and measures Success Rate, Collisions, Stuck and Time-to-Goal metrics. However, it focuses on policy-controlled agents interacting with each other and does not support human evaluation of robot behavior.

\textsc{HuNavSim}~\cite{perez2023hunavsim} is a downloadable, simulated benchmark focused on improving the development of social navigation systems around a variety of human behaviors. \textsc{HuNavSim} combines Behavior Trees (BT)~\cite{colledanchise2018behavior} and the Social Forces Model (SFM)~\cite{helbing1995social} to provide a variety of human behaviors ranging from indifferent, surprised, curious, fearful and aggressive. \textsc{HuNavSim} is implemented as a framework that can work with various simulators, and provides a plugin to work with ROS2 and Gazebo. \textsc{HuNavSim} provides a variety of metrics comparable those used in the \textsc{SEAN} simulator~\cite{tsoi2022sean} and other benchmarks, but does not provide baseline policies or a way to evaluate robot behavior with human ratings.

\textsc{SocNavBench}~\cite{biswas2022socnavbench} is a downloadable, simulated benchmark used to evaluate social navigation algorithms against prerecorded episodes of human pedestrian behavior drawn from the \textsc{ucy}~\cite{lerner2007crowds} and \textsc{eth}~\cite{pellegrini2009you} datasets. \textsc{SocNavBench} provides visually realistic pedestrians and environments, as well as baselines based on the Social Forces Model (SFM)~\cite{helbing1995social}, ORCA~\cite{van2011reciprocalORCA}, and SA-CADRL~\cite{chen2017sociallyawareDRLmit_iros17} as well as a pedestrian-unaware policy. \textsc{SocNavBench} provides a wide variety of metrics in areas such as path quality, motion quality, robot-pedestrian interaction, and episode statistics. However, \textsc{SocNavBench}'s purpose is to automatically generate scores, so it makes the design decision to focus on automatically generated metrics which approximate human ratings instead.

\subsubsection{Social Navigation Challenges} 
The \textsc{CrowdBot}~\cite{inria2020safe} Challenge is an effort to develop a benchmarking platform for social robot navigation in dense crowds. \textsc{CrowdBot} supports four different robot morphologies interacting with simulated crowds of walking humans controlled by a flexible framework called  UMANS~\cite{van2011rvo2}, with several crowd setups provided in the initial benchmark. \textsc{CrowdBot} is a downloadable, simulated challenge\footnote{\url{https://gitlab.inria.fr/CrowdBot/CrowdBotUnity/-/tree/master}}; initial phases were held in 2020 and  2021 but a full public challenge has not yet been held.

The \textsc{iGibson} Challenge at the CVPR 2021 Embodied AI Workshop\footnote{\url{https://svl.stanford.edu/igibson/challenge2021.html}} is a social navigation benchmark based on the eponymous \textsc{iGibson}~\cite{li2021igibson, shen2021igibson} simulation environment for navigation and manipulation tasks in household scenes.
In this benchmark challenge, robots must navigate to targets without collision among pedestrians \cite{DArpino_ICRA2021}, which are simulated via the ORCA model \cite{van2011rvo2} in fifteen interactive indoor household scenes. Evaluation metrics include STL (Success weighted by Time Length) for reaching the goal quickly, and PSC (Personal Space Compliance) for maintaining a comfortable distance from all pedestrians. This benchmark enabled quantitative comparison of approaches from over a dozen teams, including methods based on techniques like DD-PPO~\cite{wijmans2019dd}, PPO~\cite{schulman2017proximal}, SAC~\cite{haarnoja2018soft}, and so on, providing a clear picture of which algorithms were superior for the task. \textsc{iGibson} is a downloadable, simulated challenge, but it does not include human ratings, and in 2021 did not include on-robot tests.

The \textsc{SEANavBench} Challenge is a social navigation benchmark created for the SEANavBench workshop\footnote{\url{https://seanavbench.interactive-machines.com/}} held at ICRA'22. SEANavBench combines SocNavBench~\cite{biswas2022socnavbench} within the SEAN 2.0~\cite{tsoi2022sean} simulator which enables social navigation algorithms to run on simulated robots via ROS in environments rendered in the Unity game engine. Social navigation algorithms can be evaluated in simulated environments across a variety of environment sizes, crowd densities, and pedestrian behavior, including simulated pedestrians and replay of pedestrian datasets. This enables the analysis of how algorithms can succeed or fail as environmental conditions change and the measurement of performance using a variety of metrics. \textsc{SEANavBench} is a simulated benchmark to which users can upload their code and compare performance against other submissions and baselines. While the public version of the challenge did not use human ratings, \textsc{SEANavBench} uses SEAN-EP~\cite{tsoi2021approach} to run the SEAN 2.0 simulation environment on the web, which could be used to collect human feedback.

\subsection{Strengths and Limitations of Existing Benchmarks}
As we can see from Table~\ref{tab:benchmark-characteristics}, social navigation benchmarks support a variety of scopes, from dynamic obstacle avoidance to human-robot interactions to navigation through crowds. All attempt to address features of social behavior, and many of them are downloadable, simulated benchmarks that can be efficiently deployed and which provide metrics for evaluation and sometimes baselines for comparison.

Broadly speaking, however, different types of benchmarks have characteristic limitations: (a) scalable benchmarks tend not to ground their evaluations in human data (b) benchmarks that use human data tend to need manual setup or additional components, (c) protocols for designing experiments focus only on human evaluations, and (d) few benchmarks have meaningful coverage of edge cases of navigation behavior. 

We believe these limitations are resolvable, and next present our recommendations for how good benchmarks should be designed and outline steps the community could take to improve existing benchmarks.

\subsection{Properties of a Good Social Navigation Benchmark}
Existing social navigation benchmarks have many purposes, from testing in large crowds, smaller social scenarios, algorithm improvements and even tests of benchmark fidelity themselves. However, for results of one benchmark to be useful to the rest of the community, it is important to have a common language for benchmarking, and to have a shared understanding of what it is that a benchmark tests.

To ensure that social navigation benchmarks evaluate approaches for social navigation in a way that communicates their results broadly in the social navigation community, we argue that benchmarks themselves should be evaluated against a set of commonly agreed-upon criteria. 

Based on how benchmarks are used in the field and what results they need to communicate, we recommend that benchmarks (1) evaluate social behavior, (2) include quantitative metrics, (3) provide baselines for comparison, (4) be efficient, repeatable and scalable, (5) ground human evaluations in human data, and (6) use well-validated evaluation instruments. Next, we unpack these criteria and explain how they should guide the development and usage of benchmarks.

\begin{enumerate}
\item \textbf{Guideline B1: Evaluate Social Behavior:}
A good social benchmark should evaluate the properties of algorithms in social scenarios which involve humans and robots interacting. Therefore, a social benchmark should have metrics related to social behavior and not just contain pure navigation metrics such as Success weighted by Path Length \cite{anderson2018evaluation} or pure task metrics such as success rates.

\item \textbf{Guideline B2: Include Quantitative Metrics:}
The benchmark should provide quantitative metrics on a variety of dimensions of interest, enabling researchers with different goals to use the benchmark to evaluate their algorithms with respect to their task and context and to compare to other approaches in the literature. See Section~\ref{s:metrics} for example metrics that can be used for various social navigation scenarios. Quantitative metrics are ideal to enable comparisons between approaches; these include both metrics which can be measured objectively (e.g., Personal Space Compliance \cite{igibson_challenge}) or which have validated measurement instruments (such as Likert scale evaluation with validated questions).
Ideally, these should include metrics important to the social navigation community and include both traditional navigation metrics along with socially relevant metrics, such as task success, speed of performance, safety, and proximity to humans.

\item \textbf{Guideline B3: Provide Baselines for Comparison:} At a minimum, it is recommended to have baseline policies that show worst case performance (e.g., a straight line planner that stops at obstacles) to serve as a lower bound for the benchmark. An upper bound oracle performance (e.g., demonstrations from a human, or an appropriate state-of-the-art algorithm) can also be provided if feasible. Ideally, if a state of the art approach exists, it should be compared, but it is not always feasible to include these in a given benchmark due to availability or cost.

\item \textbf{Guideline B4: Be efficient, repeatable and scalable:} 
To democratize benchmarks and promote productive competition and collaboration amongst different scientists, efficient, repeatable, and scalable benchmarks are preferable. For example, the cost to run the benchmark should not be prohibitively expensive. While some benchmarks explicitly seek to reveal unique in-the-wild variations, the benchmark should nevertheless be repeatable such that it can be repeated multiple times with comparable results when scaled to a large number of trials. A good rule of thumb is at least 30 samples for real robot trials, but this number can be determined in a more principled statistical way from data if means and variances are available.

\item \textbf{Guideline B5: Ground Human Evaluations in Human Data:}
At this point, many researchers agree that we do not have a good enough model of how humans react to robots to predict how they will react from other observables. Therefore, many researchers propose  benchmarks should measure socialness based on human evaluations. An alternative approach is to use a learned model to predict human perception of socialness of robot behaviors using a dataset of labeled examples; some researchers argue this provides a more validated metric than an ad-hoc social score; other researchers argue the context that makes these learned metrics can be lost if used in other scenarios. Nonetheless, learned metrics could offer repeatable and scalable approximations of human responses, which could be evaluated via user studies. 

\item \textbf{Guideline B6: Use well-validated evaluation instruments:}  Ideally, human questionnaires should be standardized or empirically validated and should be ecologically valid for the task at hand; validating metrics is an iterative process which involves proposing metrics, conducting studies, statistically analyzing responses, and exposing metrics to peer review in the community. Objective metrics should also be empirically validated to ensure they measure what they purport to measure. 

\end{enumerate}

To address the shortcomings of existing benchmarks against these criteria, we recommend the following:

\begin{enumerate}
    \item \textbf{Promote more human evaluation}: Many benchmarks use proxies of human ratings; while this is reasonable to enable fast evaluations, the community should encourage benchmark developers to collect human ratings, and should push for broader adoption of rating pipelines such as SEAN-EP~\cite{tsoi2021approach} to facilitate this collection.
    \item \textbf{Standardize social questionnaires}: While it is useful to have well defined scenarios as in the \textsc{Social Navigation Protocol}, the improvements to the questionnaires made by subsequent work in this area should be standardized and made available to inform labeling pipelines.
    \item \textbf{Standardize quantitative metrics}: While some existing benchmarks and protocols specify minimum quantitative metrics, \textsc{SocNavBench}, \textsc{SEANavBench} and \textsc{HuNavSim} are converging on metrics similar to \textsc{CrowdBot}'s metrics; the community should encourage adopting a minimum set of these metrics.
    \item \textbf{Test corner cases on standard benchmarks}: While social metrics are important, ensuring safe, reliable navigation performance is also important. Navigation benchmarks such as \textsc{barn}~\cite{perille2020benchmarking} or \textsc{Bench-MR}~\cite{heiden2021bench} should be used to validate traditional navigation behaviors.
\end{enumerate}

Finally, it is worth noting that there are additional multiagent benchmarks focused on gridworlds such as \textsc{Asprilo}\footnote{\url{https://asprilo.github.io/}} for logistics and \textsc{mapf}\footnote{\url{https://movingai.com/benchmarks/mapf.html}} for multiagent pathfinding which we did not discuss as they do not focus on aspects of social behavior; however, as social navigation approaches become integrated into multiagent or logistically complex domains, features from these benchmarks may also be useful for testing corner cases.

\begin{figure*}
    \centering
    \begin{subfigure}[b]{0.18\textwidth}
     \centering
     \includegraphics[width=\textwidth]{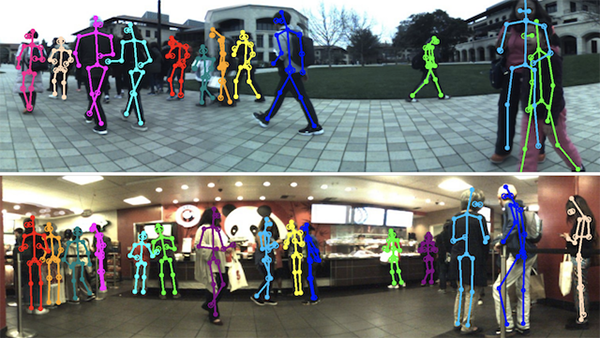}
     {\small \caption{\textsc{jrdb}}}
    \end{subfigure}
    \hfill
    \centering
    \begin{subfigure}[b]{0.18\textwidth}
     \centering
     \includegraphics[width=\textwidth]{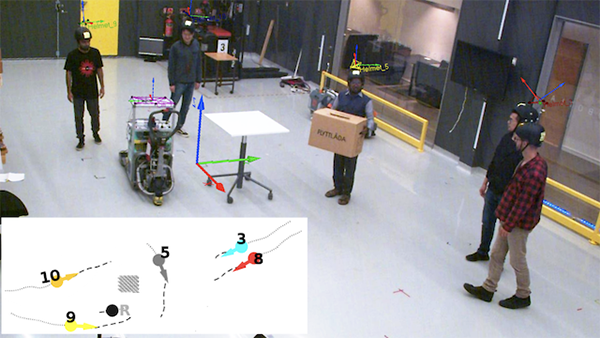}
     {\small \caption{\textsc{thor}}}
    \end{subfigure}
    \hfill
    \begin{subfigure}[b]{0.18\textwidth}
     \centering
     \includegraphics[width=\textwidth]{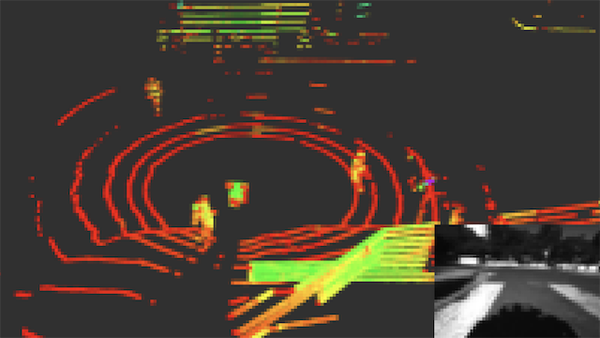}
     {\small \caption{\textsc{scand}}}
    \end{subfigure}
    \hfill
    \begin{subfigure}[b]{0.18\textwidth}
     \centering
     \includegraphics[width=\textwidth]{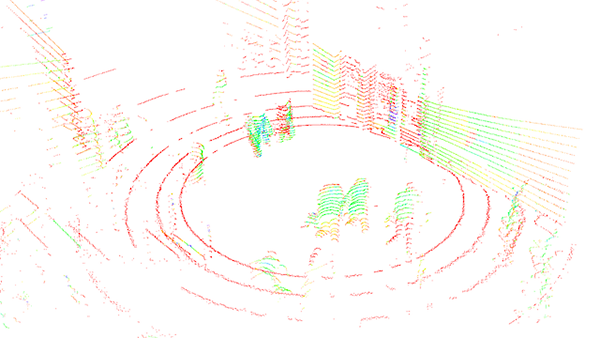}
     {\small \caption{\textsc{lcas}}}
    \end{subfigure}
    \hfill
    %
    \centering
    \begin{subfigure}[b]{0.18\textwidth}
     \centering
     \includegraphics[width=\textwidth]{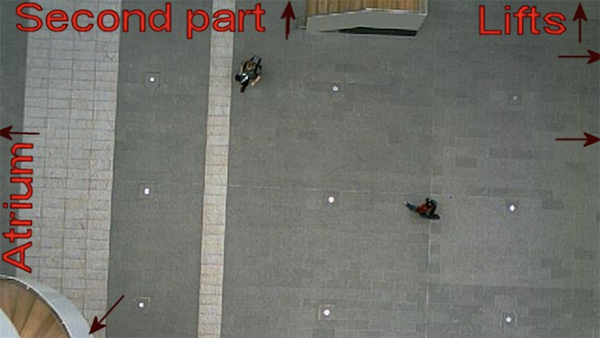}
     {\small \caption{\textsc{eifpd}}}
    \end{subfigure}
    \bigskip
    \begin{subfigure}[b]{0.18\textwidth}
     \centering
     \includegraphics[width=\textwidth]{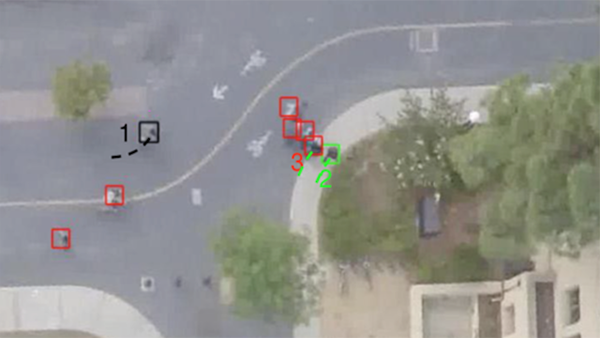}
     {\small \caption{\textsc{sdd}}}
    \end{subfigure}
    \hfill
    \begin{subfigure}[b]{0.18\textwidth}
     \centering
     \includegraphics[width=\textwidth]{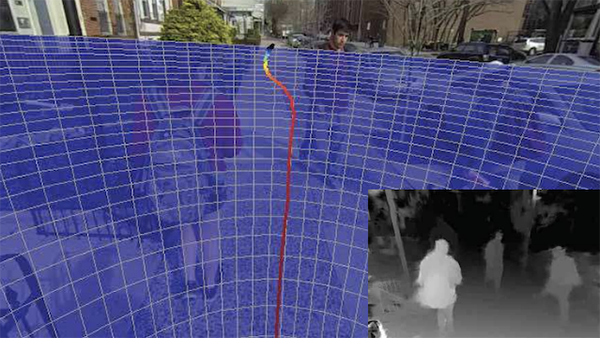}
     {\small \caption{\textsc{efl}}}
    \end{subfigure}
    \hfill
    \begin{subfigure}[b]{0.18\textwidth}
     \centering
     \includegraphics[width=\textwidth]{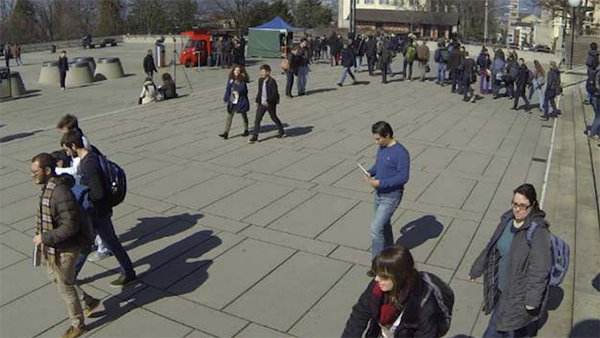}
     {\small \caption{\textsc{wildtrack}}}
    \end{subfigure}
    \hfill
    \begin{subfigure}[b]{0.18\textwidth}
     \centering
     \includegraphics[width=\textwidth]{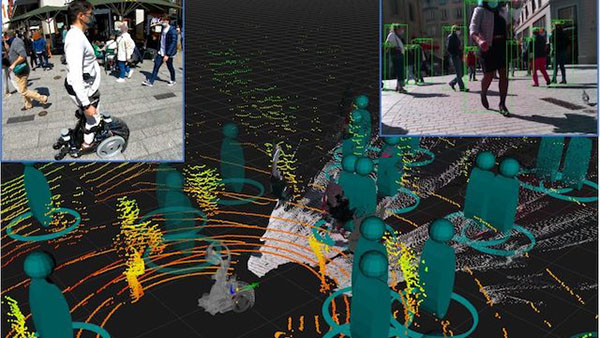}
     {\small \caption{\textsc{crowdbot}}}
    \end{subfigure}
    \hfill
    \centering
    \begin{subfigure}[b]{0.18\textwidth}
     \centering
     \includegraphics[width=\textwidth]{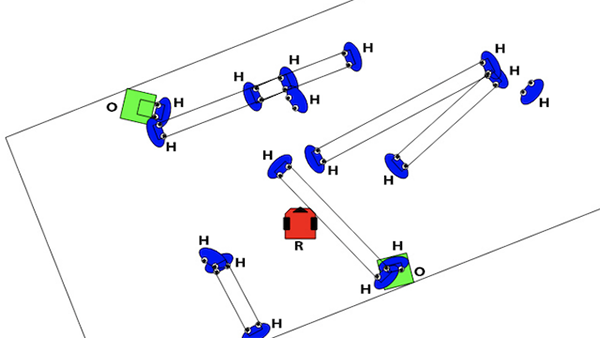}
     {\small \caption{\textsc{SocNav1}}}
    \end{subfigure}
    \bigskip
    
    \begin{subfigure}[b]{0.18\textwidth}
     \centering
     \includegraphics[width=\textwidth]{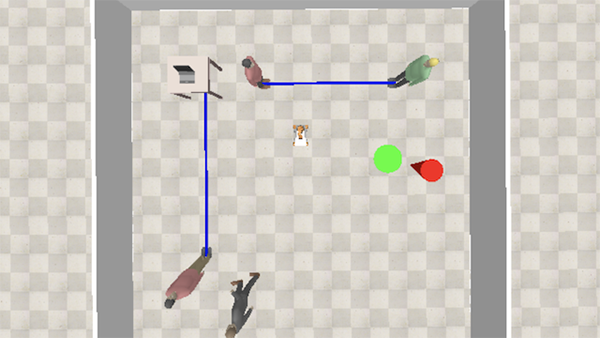}
     {\small \caption{\textsc{SocNav2}}}
    \end{subfigure}
    \hfill
    \begin{subfigure}[b]{0.18\textwidth}
     \centering
     \includegraphics[width=\textwidth]{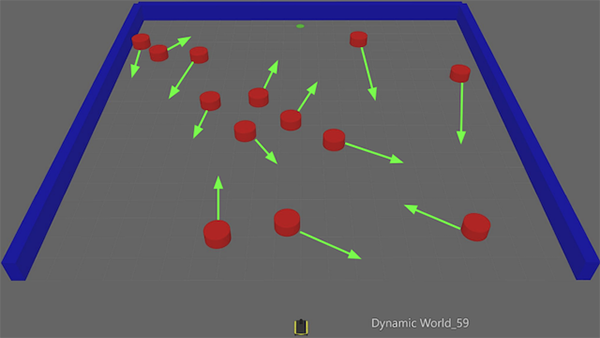}
     {\small \caption{\textsc{DynaBARN}}}
    \end{subfigure}
    \hfill
    \begin{subfigure}[b]{0.18\textwidth}
     \centering
     \includegraphics[width=\textwidth]{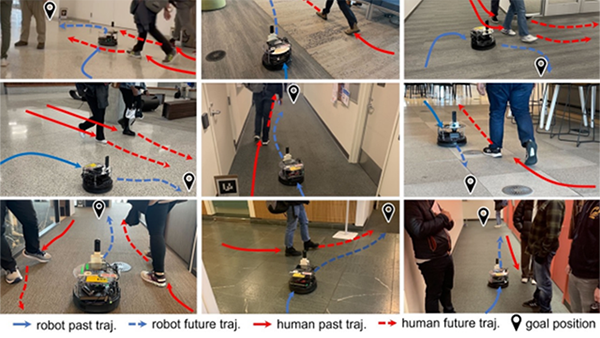}
     {\small \caption{\textsc{SACSoN}}}
    \end{subfigure}
    \hfill
    \begin{subfigure}[b]{0.18\textwidth}
     \centering
     \includegraphics[width=\textwidth]{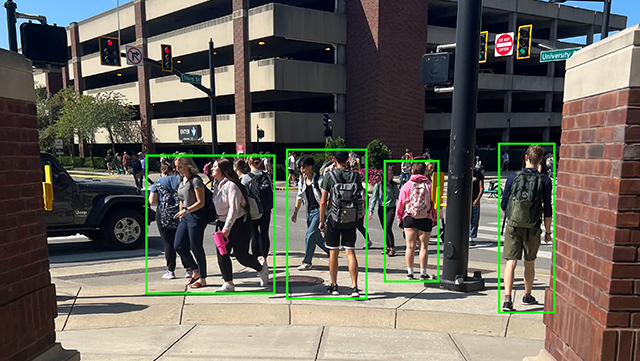}
     {\small \caption{\textsc{SG-LSTM}}}
    \end{subfigure}
    \bigskip

\caption{Illustration of various social navigation datasets. See the text and Table~\ref{tab:dataset-characteristics} for details.}
\label{fig:dataset}
\vspace{-0.1cm}
\end{figure*}

\section{Social Navigation Datasets}
\label{s:datasets}

In this section, we provide a deeper look at datasets with regards to the factors listed in Sec.~\ref{s:taxonomy}. First we review desired dataset characteristics, noting that analyzing datasets requires drilling in deeper on factors such as robot hardware, sensors, and behavior authoring methods, as well as additional factors for analysis such as data collected, dataset coverage, sampling distribution, annotations, and privacy and fairness handling. Then we use these factors to analyze several datasets, including 
\textsc{jrdb} \cite{martin2019jrdb}, \textsc{thor}~\cite{thorDataset2019}, \textsc{Trajnet++}~\cite{kothari2021its}, \textsc{eth/ucy} \cite{lerner2007crowds, pellegrini2009you}, \textsc{eifpd} \cite{majecka2009statistical}, \textsc{sdd} \cite{robicquet2016learning}, \textsc{efl} \cite{park2016egocentric}, \textsc{wildtrack} \cite{chavdarova2018wildtrack}, \textsc{scand} \cite{scanddataset,scandpaper}, \textsc{MuSoHu}~\cite{nguyen2023toward}, \textsc{CrowdBot}~\cite{paez20213d}, \textsc{DynaBARN}~\cite{nair7dynabarn}, \textsc{SocNav1}~\cite{manso2020socnav1}, \textsc{SocNav2}~\cite{bachiller2022graph}, \textsc{SACSon}~\cite{hirose2023sacson}, and \textsc{sg-lstm}~\cite{bhaskara2023sglstm}, reviewing them with respect to the criteria (Fig.~\ref{fig:dataset}).

\subsection{Expanding the Factors for Dataset Analysis}
In addition to the factors listed in Section \ref{s:factors}, additional aspects must be considered for datasets:

\paragraph{Robot Hardware Platform} As different robot morphologies might elicit different human responses, it might be of importance to consider a \textit{larger set of robots to collect data with}. Further, it might be useful to utilize props, e.g. engaging face, human like head and eye appearance and movement, to elicit stronger engagement with humans.

\paragraph{Sensors} In addition to robot sensors, a good practice is to \textit{record teleoperation commands}, e.g. joystick controls, together with the data.

\paragraph{Robot Behavior Authoring Methods} The core of a Social Navigation Dataset are demonstrations of desired socially-aware robot behaviors. How are these demonstrations defined (see Sec.~\ref{sec:social-scenarios} for a deeper discussion of this topic). Should the robot behave as a human or as a different social agent (see Sec.~\ref{sec:social-robot} for a deeper discussion on this topic). In the case of a dataset, some of the options are:
\begin{enumerate}
    \item Pedestrians/humans: If the definition of a social robot is to behave as a human, recordings of moving humans/pedestrians might suffice.
    \item Teleoperators: If a behavior is desired that might be different from human behaviors, then data can be collected via robot teleoperation. Hence, an important principle in creating a dataset is to have \textit{explicit and clear instructions to teleoperators of how to control the robot.} These instructions should cover following topics:
    \begin{itemize}
        \item Is the teleoperator visible to humans?
        \item Where is the teleoperator positioned w.r.t. the robot?
        \item Instructions should ideally guarantee that the teleoperator does not affect the human-robot interaction. 
        \item Utilize multiple teleoperators, especially for the same scenarios, to encourage diversity in the data.
    \end{itemize}
\end{enumerate}

\paragraph{Data Collected} When it comes to dataset creation one of the major questions is for what social scenarios does one collect data for. Therefore, the guidelines in Sec.~\ref{sec:social-scenarios} apply here. Note that for datasets in the wild there is limited ability to control the scenarios. On one side, one can opt for a completely unconstrained collection in a given environment, e.g. building, city, area. On the other side, one can target specific events / activities, e.g. busy areas around campus, campus cafeteria, boardwalk crowds, etc. 

An important guideline is to \textit{define the scope of the dataset such that the available dataset resources (hours of collection) are sufficient to collect data that thoroughly explores this scope.} The scope should be broad enough to present interesting challenges for the community to study. Therefore, it is desirable to \textit{make the dataset scope as broad as possible.} 

At the same time, one needs sufficient data for the dataset to be useful. More concretely, each \textit{scenario within the dataset scope should be well sampled in the dataset.} This can help ensure that methods developed on the dataset can be deployed in the real world within the scope of the dataset, as they are less likely to encounter out-of-distribution scenarios.

\paragraph{Annotations} A question specific to a dataset are the annotations generated after the data has been collected. When it comes to social navigation, there aren't existing taxonomies of human-robot or human-human interactions. Existing computer vision datasets and benchmarks for activity recognition can provide a good starting point, e.g. ActivityNet~\cite{caba2015activitynet}.

Another consideration is the granularity of annotation. When it comes to activities, one can annotate whole navigation episodes with global labels, or segments within these episodes. Similarly, for human tracks, one can annotate tracks only, tracks with bounding boxes, skeletal tracking and gaze, etc. 

\paragraph{Privacy and Fairness} As social navigation datasets contain humans, privacy is an important concern. Decisions must be made whether to anonymize humans and how to comply with privacy protection regulations.

\subsection{Existing Social Navigation Datasets}
\begin{table*}[t]
    \centering\tiny
    \begin{tabular}{|R{0.1\linewidth} | L{0.1\linewidth}| L{0.1\linewidth}| L{0.1\linewidth}| L{0.1\linewidth}| L{0.1\linewidth}| L{0.1\linewidth}| L{0.1\linewidth}|}
    \hline
    \scriptsize\textbf{Benchmark Dataset} & \scriptsize\textbf{THOR} & \scriptsize\textbf{SCAND} & \scriptsize\textbf{ETH/UCY} & \scriptsize\textbf{Trajnet++} & \scriptsize\textbf{EIFPD} & \scriptsize\textbf{SocNav1} & \scriptsize\textbf{SocNav2} \\
    \hline\hline
    \scriptsize Dataset Context and Scope & Three predefined human-robot social role-play activities. & Socially compliant navigation in-the-wid  & Human trajectories recorded from a bird's eye view vantage point & Human trajectories recorded from a bird's eye view vantage point & Human trajectories recorded from a bird's eye view vantage point & Scenarios with interactions labeled with a social score & Short sequences with interactions labeled with social and holistic scores \\
    \hline
    \scriptsize Environment & Curated indoor environment: two rooms with arranged furniture and motion caption system & Indoor and Outdoor campus-scale environment & Outdoor, fixed environment & Indoor and Outdoor, fixed environment & Outdoor, fixed environment & Indoor, abstract & Indoor, abstract \\
    \hline
    \scriptsize Data Collected & 60 minutes of motion tracking across 600 human trajectories & 522 minutes, consisting of 138 trajectories of teleoperation data from 4 demonstrators & Bird's eye view frames, with annotated human trajectories across time on 5 scenes & $>$200K human trajectories across dozens type scenes with high density crowds & Bird's eye view frames, with annotated human trajectories across time & 9280 static scenario descriptions with social scores & 53600 dynamic scenario descriptions with social and all-encompassing scores \\
    \hline
    \scriptsize Scenarios & Visiting areas, carrying boxes, inspecting targets & Goal-oriented social navigation & Pedestrian navigation & Pedestrian navigation & Pedestrian navigation & Evaluating robot disturbance & Evaluating robot trajectories \\
    \hline
    \scriptsize Robot Platform & Linde CitiTruck robot (W 1.56m x L 0.55x x H 1.17m) & Boston Dynamic Spot, ClearPath Jackal & N/A & N/A & N/A & Turtlebot-sized robot & Turtlebot-sized robot \\
    \hline
    \scriptsize Robot Behavior & Programmed to follow a predefined path in a socially unaware manner & Human teleoperation in a socially compliant manner & N/A & N/A & N/A & Static placement & Teleoperation and policy \\
    \hline 
    \scriptsize Human behavior & Follow a pre-defined path, and solving tasks in presence of other humans & Demonstrators teleoperate the robot in an open environment with other humans & Open world navigation & Open world navigation & Open world navigation & Static placement & Randomized simulated trajectories \\
    \hline
    \scriptsize Sensors & Stationary Velodyne 3D LiDAR, Qualisys Oqus 7+ motion tracking system, Tobii Pro Glasses for gaze tracking & Velodyne 3D LiDAR, Azure RGB, Odometry, Joystick & Stationary RGB camera overlooking pedestrians & Stationary RGB camera overlooking pedestrians & Camera fixed overhead 23 meters from the floor & Overhead view of robots and humans & Robot and human poses, with 53600 short videos \\
    \hline
    \scriptsize Tasks and Metrics & N/A & N/A & Human trajectory prediction & Human trajectory prediction & Human trajectory prediction & Acceptability of robot disturbance of humans & Acceptability of robot movement around humans \\
    \hline
    \end{tabular}

\vspace{0.5cm}

    \begin{tabular}{|R{0.1\linewidth} | L{0.09\linewidth}| L{0.075\linewidth}| L{0.065\linewidth}| L{0.08\linewidth}| L{0.09\linewidth}| L{0.06\linewidth}| L{0.06\linewidth}|L{0.06\linewidth}|L{0.07\linewidth}|}
    \hline
    \scriptsize\textbf{Benchmark Dataset} & \scriptsize\textbf{SDD} & \scriptsize\textbf{EFL} & \scriptsize\textbf{LCAS} & \scriptsize\textbf{WILDTRACK} & \scriptsize\textbf{JRDB} & \scriptsize\textbf{Crowd-Bot} & \scriptsize\textbf{Dyna-BARN} & \scriptsize\textbf{SACSoN} & \scriptsize\textbf{SG-LSTM} \\
    \hline\hline
    \scriptsize Dataset Context and Scope & Human trajectories recorded from a bird's eye view vantage point using a drone & Human trajectories recorded from the perspective of a human & Online human detection from 3D lidar scans & Multi-camera detection and tracking of moving humans & Dataset of social interactions in indoor and outdoor environments for solving perceptual tasks & Outdoor pedestrian tracking around a personal mobility robot & Diverse set of moving agent scenarios & Autonomous policy interacting with humans & Curated interaction, movements and formation of pedestrian groups \\
    \hline
    \scriptsize Environment & Outdoor environment, focusing on diverse social navigation scenarios & Outdoor scenes such as parks, malls and a university campus & Outdoor environment & Outdoor environment & Indoor and outdoor environment & Crowded outdoor scenes & Indoor, abstract & Indoor environment & Outdoor university campus environment \\
    \hline
    \scriptsize Data Collected & Bird's eye view frames with annotations of pedestrians, bikes, cars, etc in 100 different scenes with annotations of social interactions & RGBD frames recorded from an egocentric perspective & 49 minutes of 3D lidar scans & Multi-camera synchronized frames at 10fps & 60,000 annotated frames of humans, recorded from an egocentric robot view & 250K frames / 200 minutes from an egocentric POV & Moving polynomial agent scenarios & 75 hours of visual navigation with 4000 human-robot interactions collected from robot POV &  Color and depth frames, pedestrian and group bounding boxes \\
    \hline
    \scriptsize Scenarios & Real-world navigation with social interactions & Egocentric real-world navigation & Real-world navigation in crowded environments & Third-person view open-world pedestrian navigation in crowded environments & Navigation in a campus-scale crowded environment & Real-world navigation in a crowded outdoor environment & Multiple moving agents & Real-world navigation in a crowded indoor environment & Paths, green spaces, study spaces, cafes, gatherings, weather events \\
    \hline
    \scriptsize Robot Platform & 3DR Solo drone & N/A & Pioneer 3-AT mobile robot & N/A & JackRabbot mobile manipulator & Qolo personal mobility robot & N/A & iRobot Roomba & GO1 Edu robot \\
    \hline
    \scriptsize Robot Behavior & N/A & N/A & Human teleoperation & N/A & Teleoperated & Both shared-control and reactive control & N/A & Policy-controlled & Unobtrusive data collection \\
    \hline
    \scriptsize Human Behavior & Performing navigation activities such as walking, driving, and biking in a socially compliant manner & Socially compliant navigation in the open world & Open-world navigation & Open-world navigation & Open-world navigation & Open-world navigation & Navigation among moving obstacles & Open-world navigation & Many types of pedestrian groups \\
    \hline
    \scriptsize Sensors & RGB camera & GoPro Hero 3 stereo cameras with 100mm baseline & Velodyne VLP-16 3D LiDAR & three GoPro Hero 4 and four GoPro Hero 3 & 360 RGB cameras, Velodyne LiDARs, Sick LiDARs, microphone, RGBD camera, fisheye camera, wheel encoders & Point clouds, RGBD, people trackers, pose, contact forces & N/A & Spherical RGBD, fisheye RGB, 2D LIDAR, odometry, bumper & RGB-D cameras \\
    \hline
    \scriptsize Task and Metrics & Social activity recognition, planning and trajectory prediction & Trajectory prediction & Online human detection and tracking & Trajectory prediction, person tracking and re-identification & Benchmark and metrics for 2D and 3D Person detection and tracking & Benchmark for crowd navigation & Dataset of scenarios for benchmarking dynamic navigation & Learning dataset of autonomous policy interacting with humans. & Group size, walking speeds, proximity, cohesiveness, interactions. \\
    \hline

    \end{tabular}
    \caption{Characteristics of existing social navigation datasets}
    \label{tab:dataset-characteristics}
\end{table*}

In this section we review some of the existing dataset in the context of our social navigation characteristics. These are presented in Table~\ref{tab:dataset-characteristics}. We review the following datasets.

\textsc{jrdb} \cite{martin2019jrdb} is a multi-modal dataset containing stereo 360 RGB video, 3D lidar scans, audio, and wheel encoder measurements from both indoor and outdoor environments. It provides annotations for human tracking and detection along with a benchmark and metrics to compare different algorithms.

\textsc{thor}~\cite{thorDataset2019} is a public dataset providing motion trajectories of robots and humans in a range of curated scenarios of humans visiting and inspecting areas or  carrying objects. 

\textsc{scand} \cite{scandpaper, scanddataset} is a public dataset providing socially compliant navigation demonstrations recorded via teleoperating two different mobile robots in a socially compliant manner by human demonstrators. The objective behind the \textsc{scand} dataset is to study the social navigation behavior of robots in the presence of human crowds. Similar to \textsc{scand}, \textsc{MuSohu} \cite{nguyen2023toward} includes 3D lidar scans, RGBD camera images, 360\textdegree~ camera images, IMU data, and ambient sound collected from a sensor suite mounted on a helmet worn by humans walking around public spaces (instead of on a teleoperated robot), from which social robot navigation can be learned. This allows social human navigation data to be collected in the wild with a low setup cost, making \textsc{MuSoHu} easily extendable. 

Also like \textsc{scand}, \textsc{lcas} \cite{yan2017online} is a public dataset containing 3D lidar scans collected using a mobile robot teleoperated in heavily crowded environments. Unlike \textsc{scand}, the robot is not necessarily teleoperated in a socially compliant manner. The focus of \textsc{lcas} is to solve perception-related challenges in social navigation, such as online human detection.

The \textsc{eth/ucy} \cite{lerner2007crowds,pellegrini2009you} dataset consists of human trajectories recorded in public spaces from a bird’s eye view vantage point using an RGB camera. The trajectories are extracted by tracking humans from the bird's eye view images. The motivation behind the \textsc{eth/ucy}  dataset is to provide real-world trajectories of humans navigating among other humans in the scene so one can replicate-by-copying such trajectories in a simulator. Trajectories from the \textsc{eth/ucy} dataset can be used to simulate a diverse set of realistic social scenarios.  \cite{pellegrini2009you} propose conditioning the predicted future trajectory also on scene knowledge and social interactions among agents.

The \textsc{Trajnet++} \cite{kothari2021its} dataset is composed of several existing datasets such as \textsc{eth/ucy} \cite{lerner2007crowds,pellegrini2009you},  CFF crowd dataset \cite{alahi2014socially} with other synthetic data generated with ORCA\cite{van2011reciprocalORCA}. Kothari et al. \cite{kothari2021its} have shared a benchmark and challenge focusing on agent-agent scenarios. They provide proper sampling of trajectories and a unified extensive evaluation system to test the gathered methods for a fair comparison.

The Ediburgh Informatics Forum Pedestrian Database (\textsc{eifpd}) \cite{majecka2009statistical} is again similar to \textsc{eth/ucy}, while providing a much higher number of humans captured in the dataset, the camera is fixed overhead roughly about 23 meters from the floor. Humans are detected by processing this bird’s eye view image from the scene and tracking them in the scene. 

Stanford Drone Dataset (\textsc{sdd}) \cite{robicquet2016learning} is similar to \textsc{eth/ucy} since it also provides a bird’s eye view frame, recorded using a drone (unlike \textsc{eth/ucy} that uses a statically mounted camera). Compared to \textsc{eth/ucy}, the unique selling point of this dataset is large-scale images and videos of diverse scenarios including bicyclists, skateboarders, cars, buses, and golf carts navigating in the real world. 

Egocentric Future Localization (\textsc{efl}) \cite{park2016egocentric} provides RGBD sequences of frames from the perspective of a human in  various indoor and outdoor scenes such as Parks, Malls, and a Campus, with various activities such as walking, shopping, and social interaction. \textsc{efl}'s focus is human trajectory prediction in novel scenes. 

\textsc{wildtrack} \cite{chavdarova2018wildtrack} is similar to \textsc{eth/ucy}. A GoPro camera is mounted in an outdoor environment scene consisting of crowds of people walking around. This dataset focuses on person detection in the presence of severe obstacles such as other humans and static obstacles in the scene. 

The \textsc{CrowdBot}~\cite{paez20213d} consists of egocentric RGBD and point-cloud data from a Qolo robot~\cite{paez2022qolo},\cite{granados2018qolo} captured in autonomous and teleoperated modes in outdoor scenes.  

Several datasets present synthetic trajectories for benchmark comparison. \textsc{SocNav1}~\cite{manso2020socnav1} and \textsc{SocNav2}~\cite{bachiller2022graph} are datasets of human-labeled simulated human-robot interactions used for both benchmarking algorithms and as training datasets for learning algorithms. \textsc{DynaBARN}~\cite{nair7dynabarn} includes 300 synthetic environments with agents with different motion profiles.

The \textsc{SACSoN}~\cite{hirose2023sacson} dataset is a collection of egocentric RGB, RGBD, LIDAR, odometry and bumper data from a policy-controlled iRobot Roomba navigating autonomously under policy control in indoor human environments. The dataset was created by a scalable system wrapping the policy control with a help-and-rescue module enabling long-term data collection, resulting in 75 hours of data and 58 kilometers of interaction with over 4000 individual human-robot interactions. The dataset supported a continual-learning architecture which showed the ability to learn from collected data. Interestingly, the experimenters collected an ``interaction-rich'' subset of data in which the robot was encouraged to drive closer from humans - then negated this objective and used this data to train a socially-compliant policy.

The \textsc{sg-lstm}~\cite{bhaskara2023sglstm} dataset is a curated dataset designed to provide insight into pedestrian behavior collected on Purdue University's West Lafayette, Indiana campus using a GO1 Edu robot by Unitree Robotics navigating unobtrusively among pedestrians. \textsc{SG-LSTM} focuses on the interactions, movements, and group formations of pedestrians in a variety of scenarios including campus thoroughfares, gatherings, dining and study areas, green spaces, and inclement weather events.

\subsection{Guidelines for Datasets}

\textbf{Guideline D1: Make datasets as broad as possible.} This will ensure the dataset is useful to the community and will ensure investment in the data collection is well spent.

\textbf{Guideline D2: Scope datasets based on resources.} Ensure the available dataset resources are sufficient to collect data that thoroughly explores the dataset scope.

\textbf{Guideline D3: Ensure each scenario is well-sampled.} This ensures that methods trained on the dataset do not encounter out-of-distribution scenarios and the dataset is representative.

\textbf{Guideline D4: Use robots if robot behavior is desired.} While datasets of pedestrians are useful, if robots are expected to behave differently than people, recording actual robot behavior rather than just pedestrians is desirable.

\textbf{Guideline D5: Use diverse robot platforms:} Different robot morphologies may elicit different human responses, so if feasible datasets should use more than one robot morphology. 

\textbf{Guideline D6: Record behavior generation commands.} In addition to normal robot sensors, teleoperation commands should be recorded if robots are human-driven, or policy actions should be recorded if the behavior is authored.

\textbf{Guideline D7: Collect annotations systematically.} While standards for social navigation annotation are still being developed, formalizing data collection and modeling it on existing benchmarks in other fields can help. Data should be well labeled: methods used for generating human and robot behavior and collecting labels should be specified.

\textbf{Guideline D8: Consider privacy issues early.} The collection of data involving humans involves privacy, policy, legal and moral issues. Considering these issues early can ensure that the dataset does not face legal challenges.

\begin{table*}[t]
    \centering\scriptsize
    \begin{tabular}{|R{0.05\linewidth} |L{0.045\linewidth} |L{0.045\linewidth} |L{0.045\linewidth} | L{0.05\linewidth}| L{0.045\linewidth}| L{0.05\linewidth}| L{0.045\linewidth}| L{0.045\linewidth}| L{0.045\linewidth}| L{0.045\linewidth}|  L{0.045\linewidth}| L{0.045\linewidth}| 
    L{0.05\linewidth}|
    }
    \hline
    \mbox{\textbf{Simulator}} & \textbf{Crowd-Bot} & \textbf{CrowdNav} & \textbf{Dyna-BARN} & \textbf{gym-collision-avoidance} & \textbf{Hu-Nav-Sim} & \textbf{iGibson} & \textbf{InHuS} & \textbf{IMHuS} & \textbf{Menge-ROS} & \textbf{PedSimROS}  & \textbf{SEAN 2.0} & \textbf{SocialGym 2.0} & \textbf{Soc-Nav-Bench} \\
    \hline
    \hline
    Sim Focus& Crowd Sim & Crowd Sim & Dyn. \mbox{Obstacles} & Collision Avoid. & Human Sim & Social Nav & Human Sim & Human Sim & Crowd Sim & Crowd Sim& Social Nav  & Social Nav & Social Nav  \\
    \hline
    Sim Platform & Gazebo & Crowd-Nav & Gazebo & gym-collision-avoidance & Gazebo & iGibson & MORSE \& Stage & Gazebo & Menge-ROS & Gazebo & SEAN 2.0 & SocialGym 2.0 & Soc-Nav-Bench \\
    \hline
    Agent Repr. & Ped. \& Robot & Ped. \& Robot & Ped. \& Robot & Robots & Ped. \& Robot & Ped. \& Robot & Ped. \& Robot& Ped. \& Robot&  Ped. \& Robot & Ped. \& Robot & Ped. \& Robot & Ped. \& Robot & Ped. \& Robot\\
    \hline
    Scene Repr. & Model 3D & Geom. 2D & Geom. 2D & Geom. 2D & Model 3D & Scanned Mesh & Geom. 2D & Geom. 2D & Geom. 2D & Model 3D & Model 3D & Geom. 2D & Scanned Mesh \\
    \hline
    Scene Fidelity & Realist. 3D & 2D & Abstr. 3D & 2D & Abstr. 3D & Abstr. 3D & Realist. 3D& Abstr. 3D & 2D & Abstr. 3D & Realist. 3D & 2D & Realist. 3D \\
    \hline
    Physical Fidelity & Physics Model & Kinematic & Kinematic & Kinematic & Kinematic & Force \&~Mass & Kinematic & Kinematic & Kinematic & Physics Model & Physics Model & Kino-dynamic & Kinematic \\
    \hline
    Robot Fidelity & Robot Dyn. & Disc & Robot Shape & Disc & Robot Dyn. & Robot Shape & Robot Shape & Robot Shape & Disc & Robot Shape & Robot Dyn. & Kino-dynamic & Robot Shape \\
    \hline
    Ped. Sim Fidelity & UMANS & Move w/o~Gait & Poly-nomials & Policy-Based & Gait~\& Attitude & Move w/o~Gait & Gait~\& Attitude & Gait~\& Attitude & Move w/o~Gait & Gait~\& Activity & Move w/o~Gait & Move w/o~Gait & Gait \&~Pose\\
    \hline
    Ped. Viz Fidelity & Detail'd & Disc & Cylind. & Disc & Detail'd & Detail'd & Detail'd & Detail'd & Disc & Sensors & Detail'd & Polyg. & Detail'd \\
    \hline
    Ped. Reaction & UMANS & ORCA & Policy-Based & Policy-Based & SFM~\& Attitude & ORCA & React~\& Attitude & ORCA, Attitude & SFM \& ORCA & SFM & SFM & SFM & Replay Only\\
    \hline
    Sim Interop & Unity Interf. & Gym & ROS bag & Gym & ROS Interf. & Gym & ROS Interf. & ROS Interf. &  ROS Interf. & ROS Interf. & ROS \&~Unity & Gym, ROS & ROS Interface\\
    \hline
    \end{tabular}
    \caption{Characteristics of existing social navigation simulations. See Section \ref{s:simulator-factors} for details.}
    \label{tab:simulation-characteristics}
    
\end{table*}

\section{Simulation-based Evaluation}
\label{s:simulation}

The fundamental requirement for a social navigation simulator is the ability to simulate two agents at one time in a social encounter - without that, it's just traditional navigation. Beyond this core requirement, social navigation simulators span the gamut from supporting crowds of simplified agents that test dynamic navigation algorithms to simulators that recreate human appearances, footsteps, behavioral diversity, and environmental interactivity. Most benchmarks discussed in Section~\ref{s:benchmarks} rely on a simulator to make benchmarking efficient, repeatable and scalable.

In this section, we expand the social navigation factors particular to simulators, review existing simulators including
\textsc{CrowdBot}, \textsc{CrowdNav}, \textsc{DynaBarn}, \textsc{gym-collision-avoidance}, \textsc{HuNavSim}, \textsc{iGibson}, \textsc{InHuS}, \textsc{IMHuS}, \textsc{MengeROS}, \textsc{PedSimROS}, \textsc{SEAN 2.0}, \textsc{SocialGym 2.0}, and \textsc{SocNavBench}, analyze the properties of these simulators and how they may be improved. We then we attempt to find a common ground between simulators and benchmarks for social navigation by proposing a unified API in order to compute metrics along a single code path, including discussions of its high level requirements, implementation of the high-level API, and implementation efforts in representative simulation environments. We conclude with guidelines for simulator usage and development.

\subsection{Expanding the Factors for Simulator Analysis}
\label{s:simulator-factors}
In addition to the factors listed in Section \ref{s:factors}, additional aspects must be considered for simulators, including:

\paragraph{Abstraction Level} Some social simulations model large-scale crowds and do not attempt to model humans or robots in detail. For our purposes here and in Table~\ref{tab:simulation-characteristics}, we discuss only simulations that are at least capable of modeling individual human-robot interactions.
\paragraph{Simulation Focus} Similar to the notion of context, social simulations can be targeted at large-scale crowd simulation, social navigation interaction between humans and robots, or  more narrowly on dynamic obstacle avoidance.
\paragraph{Simulation Platform} Some social simulations are standalone codebases; others are built atop of existing simulators such as Gazebo or MORSE.
\paragraph{Agent Representation} Some simulations represent only one kind of interacting agent (generally, presumed to be all humans or all robots); others represent robots and pedestrians separately.
\paragraph{Scene Representation} Environmental assets for simulators include 2D geometry, modeled 3D geomoetry, and scanned meshes of real scenes; these scenes can represent abstract, indoor, or outdoor environments.
\paragraph{Scene Visual Fidelity} Some simulations are purely 2D; others use abstracted 3D representations; others attempt to render realistic 3D scenes with rich shaders.
\paragraph{Physics Simulation Fidelity} Some simulations only model the kinematics of moving agents in static environments; others model forces and object mass or kinodynamic constraints; others incorporate full physics models.
\paragraph{Robot Simulation Fidelity} Some simulations model robots as points or cylinders; others support detailed robot morphologies or even full robot simulation.
\paragraph{Pedestrian Simulation Fidelity} Some simulations model humans movement as point movement controlled by a crowd algorithm; others model humans as three-dimensional objects, and some model the human walking gait. Some add variability based on the human's personality or attitude.
\paragraph{Pedestrian Visual Fidelity} Pedestrians can be represented by 2D points, discs or polygons, 3D cylinders, basic human meshes which don't change shape, animated meshes with basic walking movements, or photorealistic agents. As photorealistic is subjective, we lump all human meshes into "detailed" for the purpose of Table~\ref{tab:simulation-characteristics}.
\paragraph{Pedestrian Reactivity} Pedestrians can move on pre-recorded trajectories without reacting to other agents, or may react using a model such as the Social Forces Model (SFM)~\cite{helbing1995social} or ORCA~\cite{van2011reciprocalORCA}. Pedestrian behavior may be also modulated with individual attitudes, behavioral styles or social activities specified by higher-level modules or using techniques such as behavior trees~\cite{colledanchise2018behavior}.
\paragraph{Simulation Interoperability} Some simulators are standalone; others support the OpenAI Gym API~\cite{brockman2016openai} or have interfaces to integrate with environments such as ROS~\cite{quigley2009ros}.

\subsection{Existing Social Navigation Simulators}
A variety of social navigation simulators have been used in the literature, from simple simulators designed to test individual algorithms to complex standalone simulators used in multiple contexts. These include:

The \textsc{CrowdBot}~\cite{inria2020safe} simulator supports four different robot morphologies interacting with simulated crowds of walking humans controlled by a flexible framework called  UMANS~\cite{van2011rvo2} built on the Gazebo simulator~\cite{koenig2004design}.

\textsc{CrowdNav}~\cite{chen2019crowd} is a 2D simulator for multi-agent scenarios using ORCA to orchestrate pedestrian discs around policy-controlled discs in simplified environments. 

\textsc{DynaBARN}~\cite{nair7dynabarn} is the simulator used in the \textsc{DynaBARN} benchmark. \textsc{DynaBARN} models crowds of pedestrians controlled by polynomial motion trajectories moving through simulated environments. Humans are reprsented by cylinders but robots are represented with full morphologies.

\textsc{gym-collision-avoidance}\footnote{\url{https://github.com/mit-acl/gym-collision-avoidance}} is a 2D simulator for multi-agent scenarios using policy-controlled cylinders in simplified environments. Humans and robots are not distinguished.

The \textsc{HuNavSim}~\cite{perez2023hunavsim} benchmark contains a simulator using SFM and behavior trees to provide a variety of human behaviors ranging from indifferent, surprised, curious, fearful and aggressive. It can work with various simulators, and represents both human gait and robot morphologies.

The \textsc{iGibson}~\cite{li2021igibson, shen2021igibson} simulation environment supports navigation and manipulation tasks in household scenes. Pedestrians are represented with moving mannequins controlled via ORCA~\cite{liu2021igibson,van2011rvo2} but robots are represented with full morphologies and objects in the environment can be moved.

\textsc{InHuS}~\cite{favier2022intelligent} is a simulator for testing social navigation algorithms against a variety of human behaviors called attitudes. It provides a general interface to ROS simulators and is currently integrated with the MORSE and Stage simulators.

The InHuS system is extended to simulate multiple human agents with modulated behaviors. This new system, called IMHuS~\cite{hauterville2022interactive}, uses ORCA for motion planning of agents and is built atop of Gazebo. The behaviors are modeled and controlled using a supervisor module.

\textsc{MengeROS}~\cite{aroor2017mengeros} is a 2D simulation designed to support very large crowds. Robots are discs, but several pedestrian reactivity models are supported including SFM and ORCA. A ROS interface allows this to be used with a variety of systems.

\textsc{PedSimROS}
\footnote{\url{https://github.com/srl-freiburg/pedsim_ros}}
 is a ROS package for pedestrian simulation based on SFM augmented with group behaviors and social activities. PedSimROS simulates behaviors in 2D, but can integrate with 3D simulators like Gazebo to incorporate physics models. Robot and pedestrian models are realistic enough for point-cloud sensors but pedestrians are visually simplified.

The \textsc{sean 2.0}~\cite{tsoi2020sean, tsoi2022sean} simulator enables social navigation algorithms to run on simulated robots via ROS in environments rendered in the Unity game engine; pedestrians are represented with full gaits and environments can be detailed.

\textsc{SocialGym 2.0}~\cite{holtz2022socialgym,socialgym2} is a simulation supporting diverse robot types and human behaviors in a 2D simulation that respects kinodynamic constraints, built atop the PettingZoo~\cite{terry2021pettingzoo} multi-agent reinforcement learning environment. 

The \textsc{SocNavBench}~\cite{biswas2022socnavbench} benchmark contains a simulator to replay  prerecorded episodes of human pedestrian behavior drawn from the \textsc{ucy}~\cite{lerner2007crowds} and \textsc{eth}~\cite{pellegrini2009you} datasets. \textsc{SocNavBench} provides visually realistic pedestrians and environments as well as robot morphologies.

\begin{figure*}[tb!p]
    \centering
    \begin{subfigure}[b]{0.24\textwidth}
     \centering
     \includegraphics[width=\textwidth]{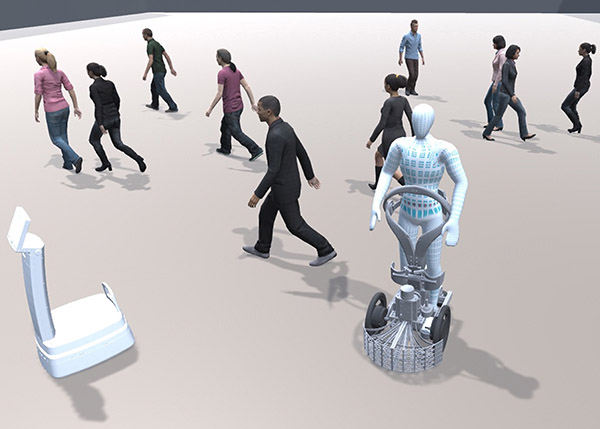}
     {\small \caption{CrowdBot}}
    \end{subfigure}
    \hfill
    \centering
    \begin{subfigure}[b]{0.24\textwidth}
     \centering
     \includegraphics[width=\textwidth]{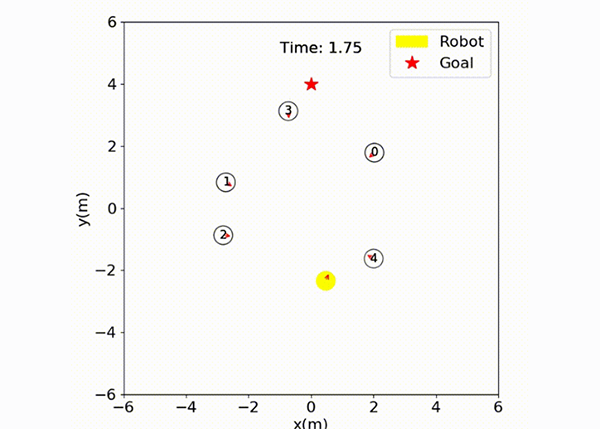}
     {\small \caption{CrowdNav}}
    \end{subfigure}
    \hfill
    \begin{subfigure}[b]{0.24\textwidth}
     \centering
     \includegraphics[width=\textwidth]{im/benchmark/dynabarn.jpg}
     {\small \caption{DynaBarn}}
    \end{subfigure}
    \hfill
    \begin{subfigure}[b]{0.24\textwidth}
     \centering
     \includegraphics[width=\textwidth]{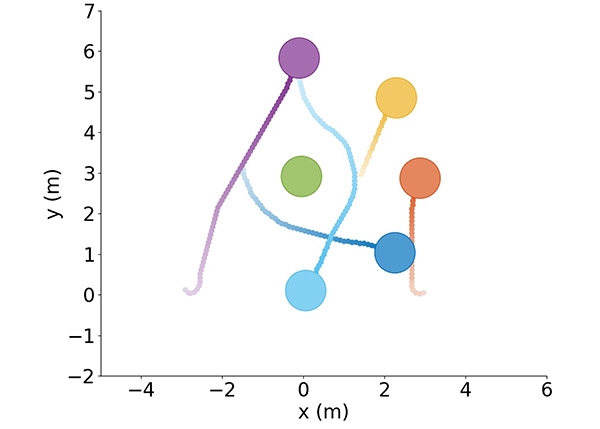}
     {\small \caption{gym-collision-avoidance}}
    \end{subfigure}
    \bigskip

    \begin{subfigure}[b]{0.24\textwidth}
     \centering
     \includegraphics[width=\textwidth]{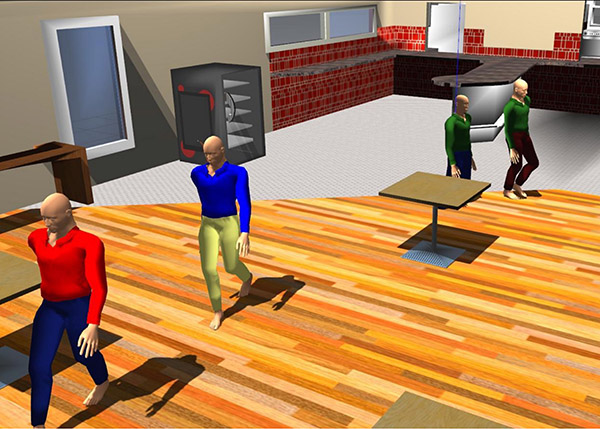}
     {\small \caption{HuNavSim}}
    \end{subfigure}
    \hfill
    \begin{subfigure}[b]{0.24\textwidth}
     \centering
     \includegraphics[width=\textwidth]{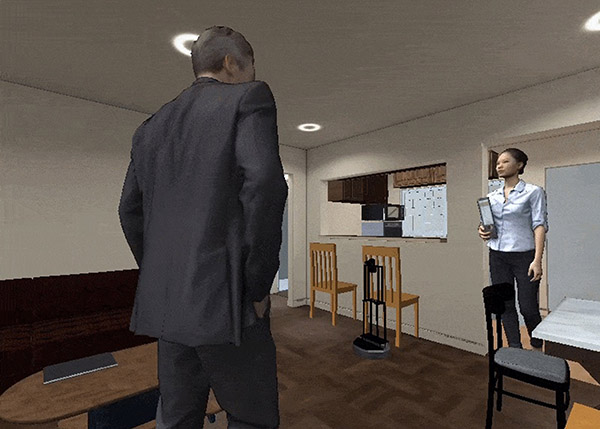}
     {\small \caption{iGibson}}
    \end{subfigure}
    \hfill
    \begin{subfigure}[b]{0.24\textwidth}
     \centering
     \includegraphics[width=\textwidth]{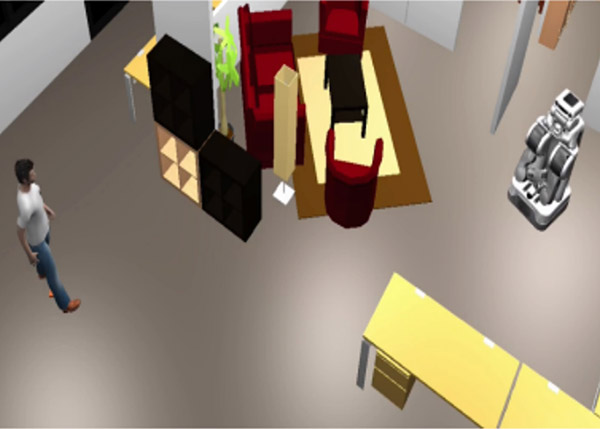}
     {\small \caption{InHuS}}
    \end{subfigure}
    \hfill
    \begin{subfigure}[b]{0.24\textwidth}
     \centering
     \includegraphics[width=\textwidth]{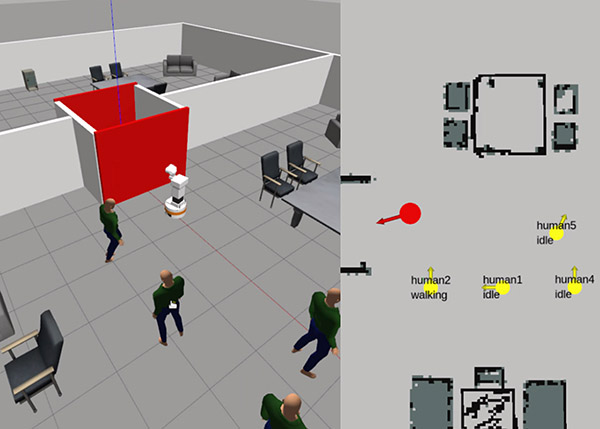}
     {\small \caption{IMHuS}}
    \end{subfigure}
    \bigskip
    
    \begin{subfigure}[b]{0.19\textwidth}
     \centering
     \includegraphics[width=\textwidth]{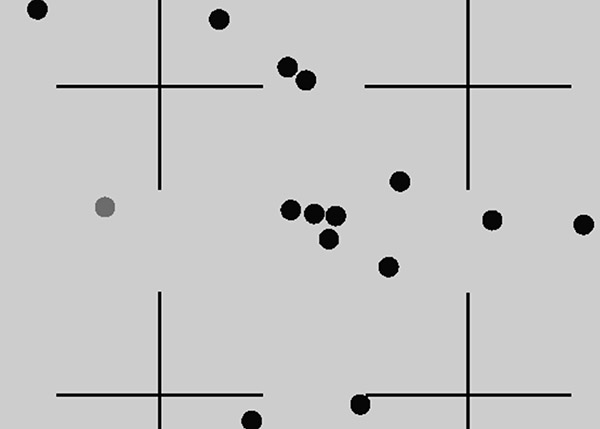}
     {\small \caption{Menge-ROS}}
    \end{subfigure}
    \hfill
    \begin{subfigure}[b]{0.19\textwidth}
     \centering
     \includegraphics[width=\textwidth]{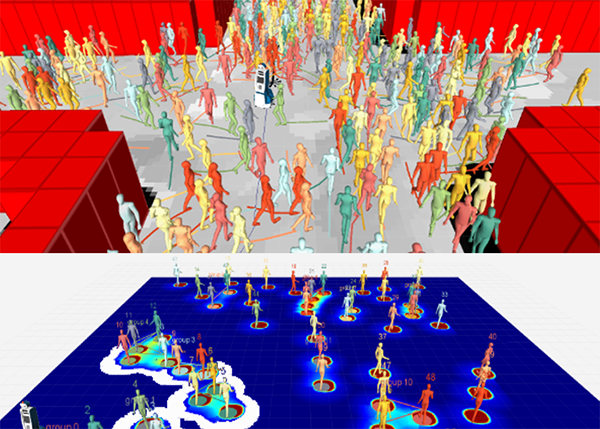}
     {\small \caption{PedSim-ROS}}
    \end{subfigure}
    \hfill
    \begin{subfigure}[b]{0.19\textwidth}
     \centering
     \includegraphics[width=\textwidth]{im/benchmark/sean2.jpg}
     {\small \caption{SEAN 2.0}}
    \end{subfigure}
    \hfill
    \begin{subfigure}[b]{0.19\textwidth}
     \centering
     \includegraphics[width=\textwidth]{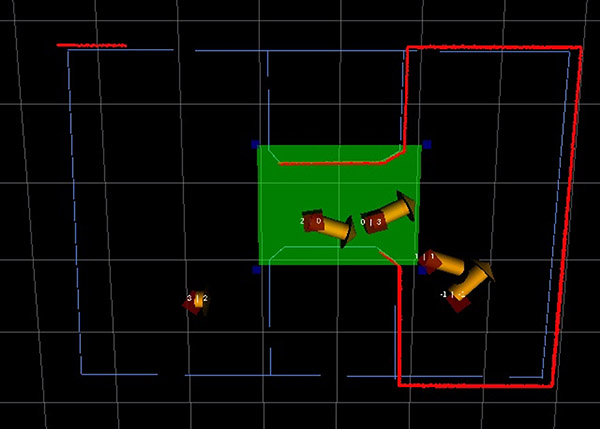}
     {\small \caption{Social-Gym 2.0}}
    \end{subfigure}
    \hfill
    \begin{subfigure}[b]{0.19\textwidth}
     \centering
     \includegraphics[width=\textwidth]{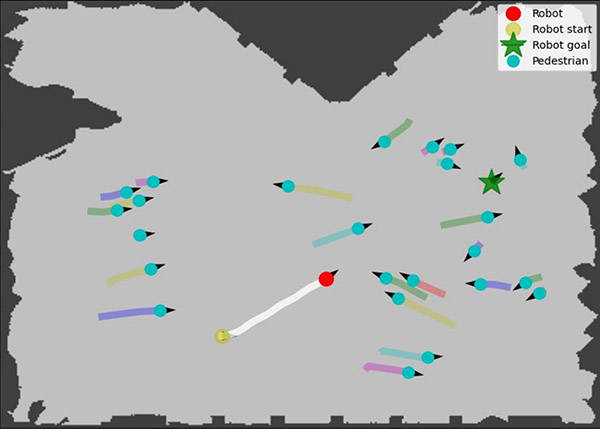}
     {\small \caption{SocNavBench}}
    \end{subfigure}
    
    {\small \caption{Visual description of select simulators. See Section \ref{s:simulator-factors} for details.}}
    \label{fig:simulators}
\end{figure*}

\subsection{Analysis of Simulation Platforms}

\subsubsection{Simulation Focus}
Each simulation platform has been designed with a focus on a particular problem area.
Example areas of focus include crowd simulation, how a robot should deal with dynamic obstacles, or specific tasks such as social navigation or collision avoidance.
Algorithms developed in different simulators may have a unique focus area as well, which implies we should be mindful when comparing algorithms across different simulators.
For example, results from an algorithm trained in a simulator that uses a cylindrical representation of pedestrians may not be directly comparable to an algorithm that incorporates pedestrian gait.

We acknowledge the need for specialized simulators focusing on different problem areas.
At the same time, we believe the community could benefit from a common social navigation simulator or a common API for multiple simulators.
This common interface would provide access to a shared set of features that span focus areas.
A common simulator or common API would enable training and evaluation across different approaches and promote the reuse of features from simulators that are focused on different areas.

\subsubsection{Common Platforms}
Many of the simulators listed in Table \ref{tab:simulation-characteristics} have shared properties. For example, several simulators use Unity, ROS, or Gazebo as an underlying technology; simulators that use the same type of scene representations could share these representations; and methods of pedestrian reactivity could also be shared across simulators.

\subsubsection{Pedestrian Reactivity}
How pedestrians react to other nearby agents is an important factor to consider because the actions of simulated pedestrians directly influence both the training and evaluation of social navigation policies.
Ideally, each simulated pedestrian would act in a manner identical to how a real-world pedestrian would act.
Real-world pedestrian behavior can be observed by recording real-world pedestrian trajectories and playing them back in a simulator. However, there are two downsides to this approach. First, some fidelity of human motion is lost in the recording and playback process, for example, it is typical to capture motion only along the ground plane and not incorporate the body pose~\cite{lerner2007crowds, pellegrini2009you}. Second, because the position of each pedestrian is determined by a pre-recorded trajectory, pedestrians cannot react to changes in the simulation. As soon as some element of the simulation deviates from the original data, such as the robot changing course, pedestrian motion is no longer realistic.

Motion models for pedestrians such as SFM~\cite{helbing1995social} and ORCA~\cite{van2011reciprocalORCA} enable them to move in reaction to changes in the environment.
While no model of human motion is perfect, the modeling of reactive agents in simulation allows researchers to explore how changes made to the environment by different robot policies affect task performance.

Pedestrian motion and reactivity play a critical role in the study of social navigation~\cite{mavrogiannis2019effectsnavstrategiesHRI, chen2019crowd}.
The two imperfect solutions we have discussed indicate an opportunity for collaboration with the community to develop better alternatives.  

\subsubsection{Multi-Agent Policies}
In the previous section, we explored various approaches to model pedestrian motion, including the use of SFM or ORCA, as well as playback of recorded trajectory data. However, in many real-world scenarios, the policies of agents are unknown and must be learned simultaneously. The field of robotics literature extensively covers navigation among dynamic obstacles, and there has been significant progress in multi-agent reinforcement learning~\cite{chen2017sociallyawareDRLmit_iros17}, which has enabled the development of socially aware behavior in robots operating in constrained environments.

\subsubsection{Environments}
The scenario plays a crucial role in social navigation. Social navigation is not commonly observed in open environments; rather, it predominantly occurs in geometrically constrained or highly dense scenarios. Indoor spaces such as corridors, hallways, and dense areas like malls or airports are typical examples of such environments. These locations share similarities in terms of their physical characteristics. Thus, the simulators discussed thus far incorporate models that capture various aspects of such environments.

\subsubsection{Metrics}
Simulation can be a cost-effective alternative to the real world when training and evaluating robot control policies, which can in turn promotes scalability and reproducibility.
The ability to compute metrics in a fair and comparable way, across robot control algorithms and simulators, is crucial to understanding the state of the field and making progress.
Running trials and computing metrics under the same initial conditions in the real world is challenging.
Simulation, however, allows the calculation of analytical metrics using ground-truth data, which is provided by the simulator, under common initial conditions when evaluating different algorithms.
Moreover, learned metrics can be easily computed in a similar fashion and subjective metrics, which are based on human feedback, can be collected as well
\cite{manso2020socnav1, Tsoi_2021_Sean_EP, nair7dynabarn}.

\begin{figure*}
\centering
\includegraphics[width=0.95\textwidth]{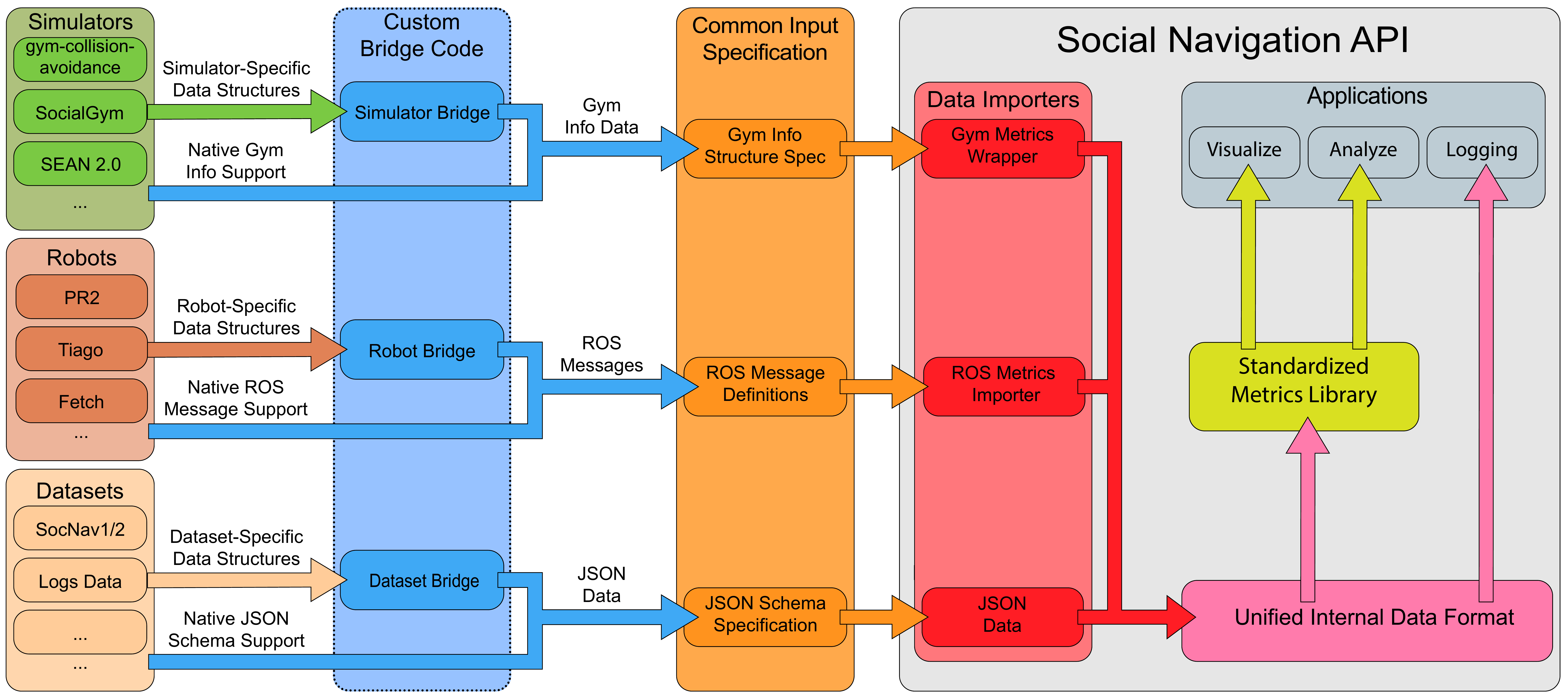}
\caption{Proposed social navigation simulation metrics API. A wide diversity of simulators and robot platforms exist, many of them supporting one or more platforms APIs such as OpenAPI Gym or ROS. We propose to define a unified API that specifies the inputs needed to generate our recommended social navigation metrics, specified as either Gym observations or ROS messages. A unified metrics API with implementations for Gym and ROS will output a single output representation, enabling post-processing tools to generate visualizations, analytics and logging with uniform code. To take advantage of these tools, simulator and robot developers only need to contribute bridge code to output the required Gym or ROS data; dataset developers only need output the single output representation.}
\label{fig:social_nav_api}
\end{figure*}

\subsection{Towards a Unified API for Social Navigation Simulation}
\label{s:social_nav_api}
As discussed in Section~\ref{s:benchmarks}, many benchmarks have been created using a variety of simulators to evaluate different aspects of social navigation. However, these benchmarks lack a unified standard for collecting metrics, making comparisons between benchmarks difficult and fragmenting the community. While different benchmarks and simulators often have divergent emphases, nevertheless, we argue many common factors could be captured by a single high-level API, which would reduce fragmentation by easing comparisons. 

Therefore, we propose a high-level API to calculate social navigation metrics that could be shared between simulators, enabling easier comparison of data collected from benchmarks built from those simulators. For broad adoption, we argue this simulation metrics API should be easy to use with a variety of simulators, real robots, and datasets, either natively or with easy-to-develop bridge code. To facilitate this interoperation, the API will specify both the data that it needs to compute metrics as well as implementations in the Robotic Operating System (ROS) and the OpenAI Gym API. These implementations will compute the metrics in Table~\ref{t:metrics} in a consistent way using common library code. While simulators often have very different code structures and philosophies, the proposed high-level API aims to help unify disparate efforts by defining a common set of data required to compute typical social navigation metrics, common library code, and a common data output format. This will make data from all API-compatible simulators and datasets available to use in shared analysis and visualization tools, which can be implemented in the future.

Figure~\ref{fig:social_nav_api} illustrates the flow of data through the proposed API. Next, we outline the API's design and our preliminary work on implementing it for common simulators.

\subsubsection{Design of the high-level simulator metric API}
To enable the calculation of the metrics recommended in Table~\ref{t:metrics} from a variety of real robots, simulators, and datasets, the simulation metric API must specify its expected input data, including the robots under test, human pedestrians and other agents, and static and dynamic obstacles in the environment. The metrics API will enable the development of common downstream tools, but to make it broadly useful to the community it should also clearly its output format, as well as provide mechanisms for extensibility to support novel use cases as they develop.

\begin{itemize}
    \item \textbf{Input Specification} To compute the desired metrics, the API requires specific data from robots, simulators, or datasets. Specifying this data requires both the format needed for specific implementations, such as ROS message or OpenAI Gym info structures, as well as the content needed for metrics, including human pedestrians and other agents, the trajectories of robots under test, and static and dynamic obstacles in the environment.
    \begin{itemize}
        \item \textbf{Pedestrian data} Simulators represent agents such as pedestrians or other robots in different ways. Typical data points include trajectories, teleoperation commands, current goals, and collective data about multiple agents, such as crowd flow. To compute many desired metrics, the proposed API requires at a minimum a pose for each agent over each timestep.
        \item \textbf{Robot data} For the robot (or other agent, such as simulated pedstrian) under test the API needs not just pose but what the robots observed, what actions they performed, and what trajectories resulted - regardless of whether robots are guided by recorded trajectories, teleoperation by humans, or control policies.
        \item \textbf{Obstacle Data} The API needs information about the geometry of the physical environment to calculate certain metrics, such as collisions or the safety of an agent's behavior; obstacle data includes static (wall geometry) and dynamic (doors, chairs) components.
    \end{itemize}
    \item \textbf{Metric Computation} The API will compute a variety of metrics listed in Table~\ref{t:metrics}, including step-wise and task/episode level metrics as discussed in Section~\ref{s:metrics}. Ideally, this metric computation should be done by standardized libraries so metrics are computed according to common definitions. This library for computing metrics should be extensible by the community, as different metrics are important to different researchers.
    \item \textbf{Output specification} The API should have a well-defined output specification so downstream tools can parse data from any system with a compatible format, facilitating the integration of datasets like those in Section~\ref{s:datasets}, even if they cannot readily be replayed in simulators.
    \item \textbf{Downstream Tools} This common output data format output will allow downstream tools to generate analyses and visualizations in a consistent way, as well as enable other data-driven applications to use data from API-compatible robots, simulators, and datasets. This could enable researchers to not only evaluate their systems in a common way but also analyze, visualize, and train data-driven systems on a variety of data from different sources with minimal feature engineering effort.
\end{itemize}

\subsubsection{Implementation of Social Navigation API}To enable the broad usage of this API, we are developing an open-source implementation at \url{https://github.com/SocialNav/SocialNavAPI}. This reference implementation will include:

\begin{enumerate}
    \item A JSON Schema specification for the data input format, along with implementations that generate this data for ROS and for GymCollisionAvoidance simulator.
    \item Reference implementations of the metrics in C++ and Python, packaged as libraries so different groups can reuse the same implementation to get comparable results.
    \item A JSON schema for the output format, with examples generating output data for ROS and OpenAI Gym.
\end{enumerate}

To use the proposed API, researchers must implement bridge code that translates data from their robots, simulators, or datasets into a format the API can consume. To make implementing bridge code easier, we will provide implementations for GymCollisionAvoidance and ROS which can be adapted for other systems, as shown in Figure \ref{fig:social_nav_api}. We are also working with the developers of SEAN 2.0, SocialGym, and DynaBarn to develop bridge code for these systems as well.

\subsection{Guidelines for Simulators}

Each simulator has its own purpose and scope, but, based on our analysis, we feel that a number of guidelines can be made for social navigation simulators which are intended to have broad use. First among these are guidelines which make it easier for simulators to interoperate:

\textbf{Guideline S1: Use Standardized APIs.} When possible, simulators should use standard APIs that enable approaches to be tested across different simulators.

\textbf{Guideline S2: Support Standard Metrics.} Simulators should provide quantitative metrics on a variety of dimensions of interest to enable different researchers to compare results - ideally, leveraging standard APIs so that metrics are computed in consistent ways, as suggested in Section~\ref{s:social_nav_api}.

\textbf{Guideline S3: Support Extensibility.} Regardless of the features a simulator supports, it is impossible to satisfy every use case. Novel research may require specific features that cannot be anticipated. Therefore, simulators should be designed with extensibility in mind, specifically enabling expert users to incorporate new functionality within the existing framework.

Next, we suggest guidelines to make simulators participate in the lifecycle of social navigation research: 

\textbf{Guideline S4: Support Dataset Generation.} Simulators should make it easy to create datasets by systematically recording data from large-scale simulated runs.

\textbf{Guideline S5: Support Benchmark Creation.} Simulators should provide an API to create tasks and scenarios and to combine them with metrics and baselines to create a social navigation benchmark.

\textbf{Guideline S6: Support Human Labeling.} Simulators should make it easy to collect human labels of the acceptability or socialness of simulated episodes.

In addition, to support the increasing sophistication of social navigation scenarios and policies, we suggest guidelines for supporting increased visual and behavioral fidelity: 

\textbf{Guideline S7: Support Common Robot Morphologies.} Simulators should provide instantiations of common robot morphologies to enable easy comparisons.

\textbf{Guideline S8: Support Detailed Pedestrians.} Where possible, simulators should support detailed pedestrian simulations to enable visual policies to react to walking pedestrian gaits. Ideally, this would extend to full visual realism of backgrounds as well, as well as replay of realistic pedestrians.

\textbf{Guideline S9: Provide Options for Behavior Authoring.} Simulators should provide ways to support behavior authoring, including playback of pedestrian recording, standard simulated models such as ORCA, and controls by custom policies. Supporting behavioral diversity in the generated policies is also important to capture the range of pedestrian behavior.

Finally, it is important to validate the simulation setup against its intended usage. Simulators should be periodically validated and refined to improve the realism and scope of the social navigation behaviors that they support.

\section{Conclusions}
\label{s:conclusions}

Social robot navigation is critical to the success of mobile robots in human environments, but challenging because it combines all the problems of traditional robot navigation with the twin challenges of understanding how a robot can and should participate with moving humans and understanding how humans react to this participation.
In this paper, we have outlined principles for social robot navigation and discussed guidelines for how these principles can be properly evaluated in scenarios, benchmarks, datasets, and simulators.

We defined a socially navigating robot as a robot that acts and interacts with humans or other robots, achieving its navigation goals while modifying its behavior to enable the other agents to better achieve theirs, and identified the key aspects needed to achieve this as safety, comfort, legibility, politeness, social competency, understanding other agents, proactivity, and responding appropriately to context.

Building on this foundation, we reviewed the methodology of social navigation research and defined a taxonomy of factors used to describe social navigation metrics, scenarios, benchmarks, datasets, and simulators. Based on a review of existing work, we proposed a list of criteria for good benchmarking, including evaluate social behavior, include quantitative metrics, provide baselines for comparison, be efficient, repeatable and scalable, round human evaluations in human data, and use well-validated evaluation instruments.

Figure \ref{fig:guidelines} summarizes these guidelines to help researchers analyze their own research efforts and make good choices for benchmarking social robot navigation. We hope this framework for understanding social robot navigation will promote clearer benchmarking and faster progress in this field, and to promote this, we also proposed a common API for social navigation metrics to improve the ease of comparison.

\section*{ACKNOWLEDGMENT}

An excerpt of this paper, focusing on the human-robot interaction (HRI) aspects of the social navigation problem, was published \cite{francis2023socialhri} in the AAAI 2023 Spring Symposium on HRI in Academia and Industry: Bridging the Gap.\footnote{https://sites.google.com/view/aaai-hri-bridge/home}

\section*{Contributions}
\label{s:contributions}

\begin{small}
\textbf{Alexandre Alahi} was a presenter at the symposium and contributed to the Definition working group.
\vspace{2mm}

\textbf{Rachid Alami} was a presenter at the symposium, and contributed to the Metrics, Evaluation and  Scenarios working groups.
\vspace{2mm}

\textbf{Aniket Bera} contributed to the Definition working group.
\vspace{2mm}

\textbf{Abhijat Biswas} was a presenter at the symposium  and contributed to the Benchmarks working group.
\vspace{2mm}

\textbf{Joydeep Biswas} contributed to the Datasets and Simulation working groups.
\vspace{2mm}

\textbf{Rohan Chandra} contributed to the Simulations working group and the Definition section.
\vspace{2mm}

\textbf{Lewis Chiang} contributed to the Metrics and Benchmarks working groups.
\vspace{2mm}

\textbf{Claudia P\'{e}rez-D’Arpino} co-led the direction and writing of the paper as co-DRI (directly responsible individual), and was an organizer, presenter and moderator of the symposium, 
interviewed symposium participants and co-authors, 
collected their position papers and organized them into a draft, edited the manuscript, 
led the Evaluation working group, and assisted with the Simulator subgroup.
\vspace{2mm}

\textbf{Michael Everett} was a presenter at the symposium, contributed a position paper, and contributed to the Simulators working group.
\vspace{2mm}

\textbf{Anthony Francis} co-led the writing and direction of the paper as co-DRI (directly responsible individual), was an organizer and moderator of the symposium, interviewed symposium participants and co-authors, collected their position papers and organized them into a draft, helped draft many sections of the manuscript, edited the manuscript, led the Benchmarks working group and its subgroups, assisted with the Evaluation, Metrics and Simulation working groups, helped write and organize the Principles and Guidelines, helped organize, create, edit and format the tables and figures, and created Figures \ref{fig:principles}, \ref{fig:context}, \ref{fig:lifecycle}, and \ref{fig:scenarios}.
\vspace{2mm}

\textbf{Sehoon Ha} was a presenter at the symposium, contributed a position paper, and contributed to the Benchmarks working group.
\vspace{2mm}

\textbf{Justin Hart} was a presenter at the symposium, contributed a position paper, and contributed to the Definition and Datasets working groups.
\vspace{2mm}

\textbf{Jonathan How} was a presenter at the symposium, contributed a position paper, and helped edit the document.
\vspace{2mm}

\textbf{Haresh Karnan} contributed to the Datasets working group.
\vspace{2mm}

\textbf{Edward Lee} contributed to the Metrics and Evaluation working groups.
\vspace{2mm}

\textbf{Chengshu Li} was an organizer, presenter and moderator for the symposium, helped define the process for the paper, helped collate the participant’s comments, helped draft many of the sections of the document, led the Metrics working group, and assisted the Simulator working group.
\vspace{2mm}

\textbf{Luis J. Manso} was a presenter at the symposium, contributed a position paper, and contributed to the Metrics working group.
\vspace{2mm}

\textbf{Roberto Martin-Martin} was an organizer and presenter for the symposium, helped guide the direction for the paper, helped collate the participant position papers, helped draft many of the sections of the document, led the Definition working group, and assisted with the Datasets working group.
\vspace{2mm}

\textbf{Reuth Mirksy} contributed a position paper and contributed to the Definition and Metrics working groups.
\vspace{2mm}

\textbf{S\"oren Pirk} was a presenter at the symposium and contributed to the Metrics, Benchmarks and Evaluation working groups.
\vspace{2mm}

\textbf{Phani Teja Singamaneni} was a presenter at the symposium and contributed to the Figures, Metrics, Simulation and Evaluation working groups.
\vspace{2mm}

\textbf{Peter Stone} was a presenter at the symposium, contributed a position paper, and contributed to the Definition and Datasets working groups.
\vspace{2mm}

\textbf{Ada V. Taylor} was a presenter at the symposium, contributed to the Metrics working group, Related Work, and editing across sections.
\vspace{2mm}

\textbf{Alexander Toshev} was an initiator, organizer and presenter for the symposium, proposed and helped set the direction for the paper, helped collate the participant position papers, helped draft many of the sections of the document, led the Datasets working group, and assisted with the Metrics working group.
\vspace{2mm}

\textbf{Peter Trautman} was a presenter at the symposium, advised on the paper, contributed a position paper, and contributed to the Datasets and Simulators working groups.
\vspace{2mm}

\textbf{Nathan Tsoi} was a presenter at the symposium, contributed a position paper, and contributed to the Metrics, Simulation, and Benchmarks working groups.
\vspace{2mm}

\textbf{Marynel Vázquez} was a presenter at the symposium, contributed a position paper, and contributed to the Metrics working group.
\vspace{2mm}

\textbf{Fei Xia} was an organizer and moderator for the symposium, helped define the process for the paper, helped collate the participant position papers, helped draft many of the sections of the document, and led the Simulator working group.
\vspace{2mm}

\textbf{Xuesu Xiao} was a presenter at the symposium, contributed a position paper, and contributed to the Benchmarks and Datasets working groups.
\vspace{2mm}

\textbf{Peng Xu} contributed to the Benchmarks working group.
\vspace{2mm}

\textbf{Naoki Yokoyama} was a presenter at the symposium, contributed a position paper, and contributed to the Benchmarks working group.

\end{small}


\bibliographystyle{IEEEtranN}
\bibliography{root}

\end{document}